\documentclass[10pt,twocolumn,letterpaper]{article}

\usepackage{iccv}
\usepackage{times}
\usepackage{epsfig}
\usepackage{graphicx}
\usepackage{amsmath}
\usepackage{amssymb}

\usepackage{mathtools}
\usepackage[ruled,vlined]{algorithm2e}
\newcommand{\floor}[1]{\ensuremath\left\lfloor#1\right\rfloor}
\newcommand{\round}[1]{\ensuremath\left\lfloor#1\right\rceil}
\usepackage{amsthm}

\theoremstyle{remark}

\usepackage{caption}
\usepackage{subcaption}
\usepackage{stix}
\usepackage[dvipsnames]{xcolor}
\usepackage{rotating}
\usepackage{comment}
\usepackage{lipsum}
\usepackage{dblfloatfix}    
\usepackage[inline]{enumitem}

\usepackage[pagebackref=true,breaklinks=true,letterpaper=true,colorlinks,bookmarks=false]{hyperref}

\let\oldFootnote\footnote
\newcommand\nextToken\relax

\renewcommand\footnote[1]{%
    \oldFootnote{#1}\futurelet\nextToken\isFootnote}

\newcommand\isFootnote{%
    \ifx\footnote\nextToken\textsuperscript{,}\fi}

\iccvfinalcopy 

\def\iccvPaperID{7566} 

\ificcvfinal\fi

\begin{document}

\title{Evaluating Post-Training Compression in GANs using Locality-Sensitive Hashing}

\author{Gon\c{c}alo Mordido\\
Hasso Plattner Institute\\
Potsdam, Germany\\
{\tt\small goncalo.mordido@hpi.de}
\and
Haojin Yang\\
Hasso Plattner Institute\\
Potsdam, Germany\\
{\tt\small haojin.yang@hpi.de}
\and
Christoph Meinel\\
Hasso Plattner Institute\\
Potsdam, Germany\\
{\tt\small christoph.meinel@hpi.de}
}

\maketitle
\ificcvfinal\thispagestyle{empty}\fi

\begin{abstract}
  The analysis of the compression effects in generative adversarial networks (GANs) after training, \textit{i.e.} without any fine-tuning, remains an unstudied, albeit important, topic with the increasing trend of their computation and memory requirements.
  While existing works discuss the difficulty of compressing GANs during training, requiring novel methods designed with the instability of GANs training in mind, 
  we show that existing compression methods (namely clipping and quantization) may be directly applied to compress GANs post-training, without any additional changes.
  High compression levels may distort the generated set, likely leading to an increase of outliers that may negatively affect the overall assessment of existing k-nearest neighbor (KNN) based metrics.
  We propose two new precision and recall metrics based on locality-sensitive hashing (LSH), which, on top of increasing the outlier robustness, decrease the complexity of assessing an evaluation sample against $n$ reference samples from $O(n)$ to $O(\log(n))$, if using LSH and KNN, and to $O(1)$, if only applying LSH.
  We show that low-bit compression of several pre-trained GANs on multiple datasets induces a trade-off between precision and recall, retaining sample quality while sacrificing sample diversity.
\end{abstract}

\section{Introduction}\label{sec:intro}

Existing compression techniques have been shown to negatively affect Generative Adversarial Networks (GANs~\cite{gans}) during training~\cite{shu2019co,chen2020distilling,qgan,yu2020self,li2020gan}, often leading to low sample quality and/or low sample diversity (commonly referred to as mode collapse).
Yu and Pool~\cite{yu2020self} suggested the high-entropy in the generator ($G$) input and output as well as the inherent training instability of GANs~\cite{mordido2018dropout,mordido2020microbatchgan,sauder-etal-2020-best} as the main causes for this ineffectiveness.

In this work, we study the effects of post-training compression in pre-trained GANs, without the need for any retraining fine-tuning. This is important when there is a lack of computational power or time to fine-tune a compressed model, as well as inaccessibility to the training data or training details used to train the original model. Our main goals are two-fold:  
\begin{enumerate*}[label={(\arabic*)}]
  \item analyze if current compression techniques, such as linear quantization, pruning~\cite{mcq}, splitting~\cite{ocs} and clipping~\cite{aciq}, may be successful applied to compress GANs in a post-training regime, and
  \item separately assess the detriment in terms of quality and diversity of the generated set.
\end{enumerate*}

Objective evaluation metrics that separately assess the sample quality and sample diversity in terms of precision and recall, respectively, have been recently proposed~\cite{prd,impar,mark-evaluate,fti}. Common methods use k-nearest neighbors (KNN) to obtain a topological representation, \textit{i.e. } approximated manifolds, of both a reference and evaluation set, consisting of real and generated samples. Although simple to implement, KNN-based metrics have linear complexity and are sensitive to outliers in both the reference and evaluation sets, which may compromise the overall assessment. 

\begin{figure*}[h]
     \centering
     \begin{subfigure}[b]{0.23\textwidth}
         \centering
         \includegraphics[width=\textwidth]{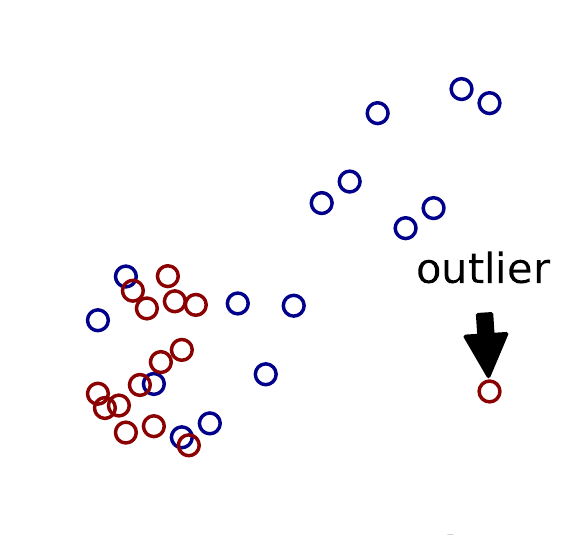}
         \vspace*{-8.5px}
         \caption{{\color{Blue}{Real}} and {\color{Red}{generated}} points.}
         \vspace*{4px}
         \label{fig:points}
     \end{subfigure}
     \hfill
     \begin{subfigure}[b]{0.23\textwidth}
         \centering
         \includegraphics[width=\textwidth]{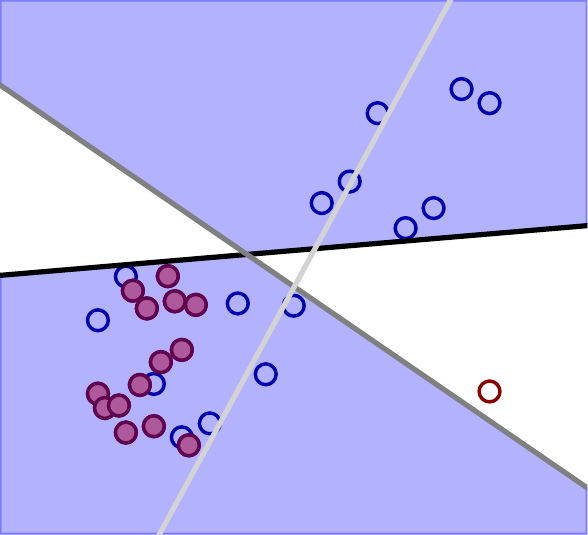}
         \caption{Precision (\textbf{LSH}): $\dfrac{\color{Red}\mdlgblkcircle}{\color{Red}\circlebottomhalfblack}$.}
         \label{fig:lsh_precision}
     \end{subfigure}
     \hfill
     \begin{subfigure}[b]{0.23\textwidth}
         \centering
         \includegraphics[width=\textwidth]{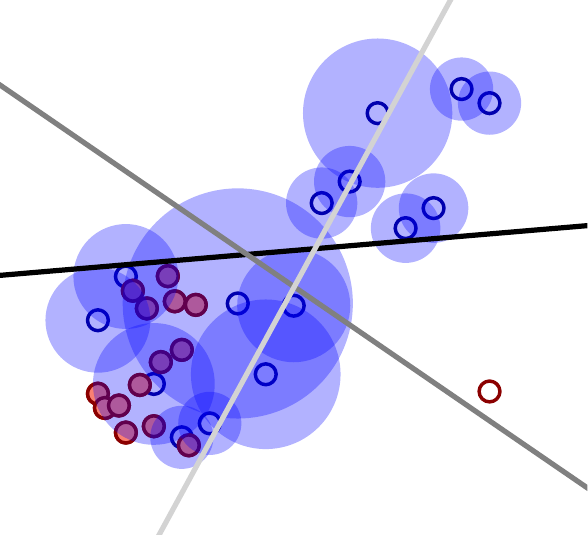}
         \caption{Precision (\textbf{LSH+KNN}): $\dfrac{\color{Red}\mdlgblkcircle}{\color{Red}\circlebottomhalfblack}$.}
         \label{fig:lsh_knn_precision}
     \end{subfigure}
     \hfill
     \begin{subfigure}[b]{0.23\textwidth}
         \centering
         \includegraphics[trim={0 2px 0 0},clip,width=\textwidth]{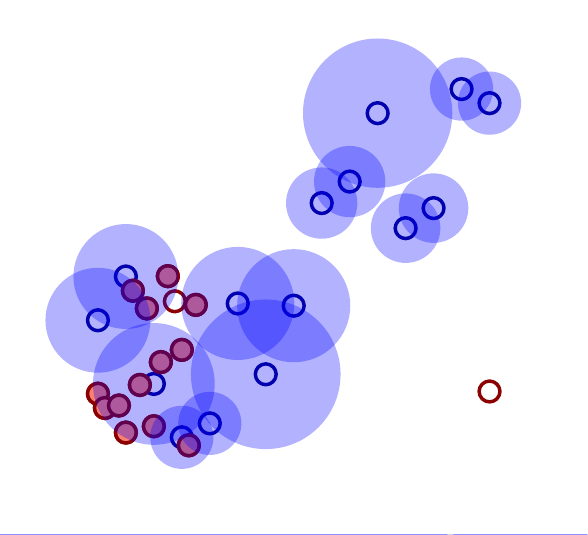}
         \vspace*{-10.5px}
         \caption{Precision (KNN~\cite{impar}): $\dfrac{\color{Red}\mdlgblkcircle}{\color{Red}\circlebottomhalfblack}$.}
         \label{fig:knn_precision}
     \end{subfigure}
     \hfill
     \begin{subfigure}[b]{0.23\textwidth}
         \centering
         \includegraphics[width=\textwidth]{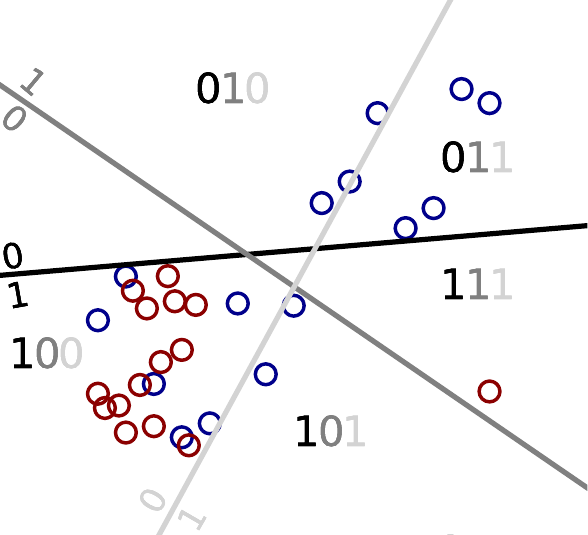}
         \vspace*{-8.5px}
         \caption{Random hyperplanes.}
         \vspace*{4px}
         \label{fig:hyperplanes}
     \end{subfigure}
     \hfill
      \begin{subfigure}[b]{0.23\textwidth}
         \centering
         \includegraphics[trim={0 0 0 2px},clip,width=\textwidth]{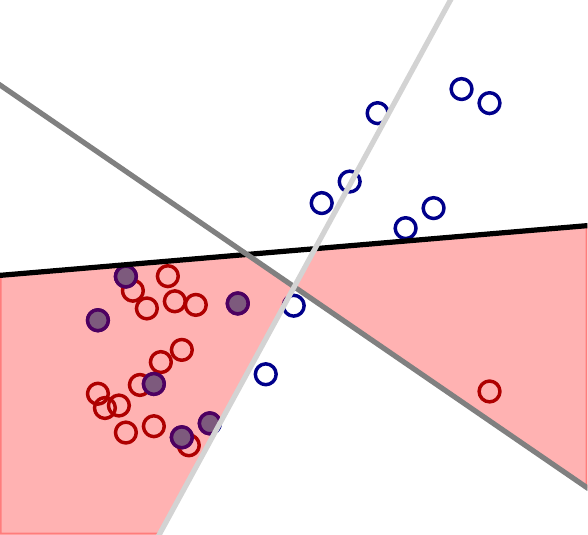}
         \caption{Recall (\textbf{LSH}): $\dfrac{\color{Blue}\mdlgblkcircle}{\color{Blue}\circlebottomhalfblack}$.}
         \label{fig:lsh_recall}
     \end{subfigure}
     \hfill
     \begin{subfigure}[b]{0.23\textwidth}
         \centering
         \includegraphics[width=\textwidth]{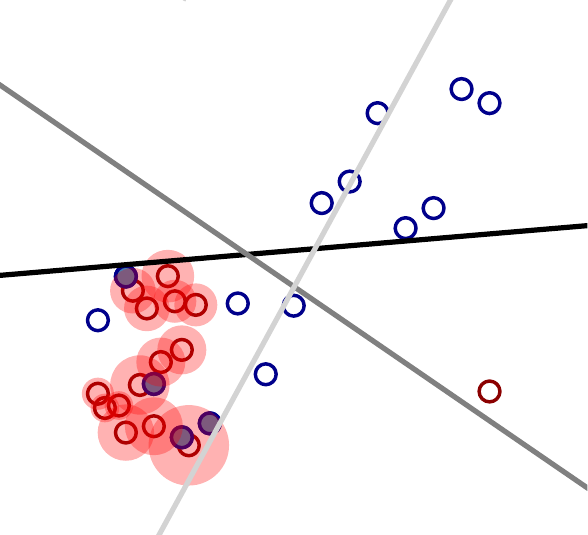}
         \caption{Recall (\textbf{LSH+KNN}): $\dfrac{\color{Blue}\mdlgblkcircle}{\color{Blue}\circlebottomhalfblack}$.}
         \label{fig:lsh_knn_recall}
     \end{subfigure}
     \hfill
     \begin{subfigure}[b]{0.23\textwidth}
         \centering
         \includegraphics[width=\textwidth]{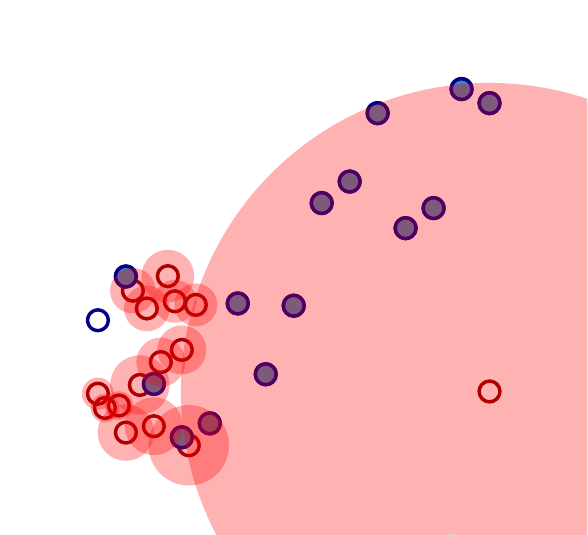}
         \caption{Recall (KNN~\cite{impar}): $\dfrac{\color{Blue}\mdlgblkcircle}{\color{Blue}\circlebottomhalfblack}$.}
         \label{fig:knn_recall}
     \end{subfigure}
     \hfill
     \caption{Evaluation methods. Considering a set of real and generated points~(a), we split the data space into several regions by generating random hyperplanes (e). Each region is characterized by which side it lies regarding each hyperplane. In our LSH metric, we consider points inside the same region, \textit{i.e.} manifold, to calculate precision and recall (b, f). In our LSH + KNN metric, we consider the hyperspheres of nearest points (K=1 in this example) within each region as the approximated manifold (c, g). Existing KNN approaches~\cite{impar}, consider all points to generate hyperspheres (d, h), which may have undesirable effects in dealing with outliers (h). Our methods help mitigate this by likely inserting outliers in isolated regions (f, g).
     }
     \label{fig:methods}
\end{figure*}

In our use-case, it is especially important to deal with outliers since high compression rates may distort the generated set. To this end, we propose to approximate the KNN manifold by using locality-sensitive hashing (LSH). In particular, we use random hyperplanes to divide the data space into different regions. Then, we only calculate KNN on points that are in the same region. The computation may be improved at the cost of a poorer approximation by not applying KNN at all and considering entire regions as manifolds. Since outliers are likely to be isolated in a certain region, they are less likely to corrupt the evaluation process (see Figure~\ref{fig:methods}).

On another note, we improve the computational complexity of assessing an evaluation sample against $n$ reference samples of current KNN-based metrics from $O(n)$ to $O(\log(n))$, when applying KNN on top of LSH, and to $O(1)$, if only using LSH. Improving the complexity in order of $n$ is of extreme importance since increasing $n$ is likely to lead to a more accurate assessment in all evaluation methods.

We find that existing compression techniques, such as clipping and quantization, may be successfully applied to pre-trained GANs, providing a trade-off between precision and recall: while sample quality is mostly retained, sample diversity is majorly affected. 

\section{Related work}

Several recent methods propose to alleviate the issues of compressing GANs \textit{during training}.
Shu \etal~\cite{shu2019co} used evolutionary algorithms to perform channel pruning and construct a compressed $G$.
Aguinaldo \etal~\cite{aguinaldo2019compressing} used knowledge distillation~\cite{hinton2015distilling} to train a compressed student $G$ based on a larger teacher $G$ trained with the discriminator ($D$) feedback.
Chen \etal~\cite{chen2020distilling} added a student $D$ to the distilling framework allowing an increase of student $G$'s compression levels.
Koratana \etal~\cite{lit} used learned intermediate representation training (LIT) to learn only a student $G$, without updating a pre-trained teacher $G$.
Wang \etal~\cite{qgan} proposed a new compression scheme based on Expectation-Maximization (EM) algorithms to compress both $G$ and $D$ during training.
Liu \etal~\cite{liu2020binarizing} suggested quantizing more layers in $D$ than $G$ to stabilize training.
Yu \etal~\cite{yu2020self} proposed a self-supervised compression mechanism that combines the feedback from a pre-trained $D$ as well as the original $G$ to train a pruned $G$.
Wang \etal~\cite{wang2020ganslimming} combined model distillation, channel pruning, and quantization on top of the original minimax objective to compress GANs in an end-to-end manner.
Li \etal~\cite{li2020gan} proposed a general framework to compress generators in conditional GANs by combining knowledge distillation and neural architecture search~\cite{nas}.

One big shortcoming of previous works is in the evaluation of the generated samples using a single-valued metric, namely FID~\cite{fid} and Inception Score~\cite{is}. This is sub-optimal since no separate assessment of the sample quality and sample diversity was performed. Several precision and recall metrics have been recently proposed~\cite{prd,impar,mark-evaluate,fti}. In this work, we show that common single-valued metrics, such as FID and KID~\cite{kid} tend to correlate more with recall than precision, biasing against models with high sample quality but low sample diversity. This reinforces the importance of using double-valued metrics that separately assess the quality and diversity of the generated set. Hence, we propose two new evaluation metrics that evaluate precision and recall with LSH and KNN. Our metrics are outlier-aware and more efficient than current KNN-based approaches.

To the best of our knowledge, we are the first work to study the effects of \textit{post-training} compression in GANs, which enable the compression of pre-trained GANs without the need of additional training. Since we do not have to deal with the training instability of GANs, we show that current quantization methods may be directly applied to compress a pre-trained G to low bit-width.




\section{Locality-sensitive hashing (LSH)}

As briefly mentioned, using solely KNN~\cite{impar} to evaluate a generated set has several shortcomings. The more meaningful of which is the inability to deal with outliers in the reference and evaluation sets, which may greatly distort the approximation of manifolds and lead to an inaccurate assessment of precision and/or recall. The aforementioned distortion in the approximated manifolds may also occur with sparse data, where points are sparsely distributed and manifolds may cover most of the data space. Lastly, KNN-based approaches have linear complexity on the number of reference samples, which is sub-optimal and expensive at evaluation time.

One common way of improving the efficiency of KNN is to either reduce the data dimensionality or only compare each sample to a subset of samples. In this work, we focus on the latter. Several algorithms may be used to achieve this. Specifically, K-D trees~\cite{kdtrees} are a viable alternative for low-dimensional, continuous data, with the downside of possibly missing nearest neighbors. Inverted lists are commonly used for high-dimensional, discrete data, providing an efficient alternative without the risk of not detecting nearest neighbors. 

However, neither K-D trees nor inverted lists are good alternatives in our use-case due to the nature of our data. As initially proposed by Kynkäänniemi \etal~\cite{impar}, each image is passed as input to a pre-trained VGG-16 model~\cite{vgg}, being represented as the embeddings of the second fully connected layer. Hence, our data is high-dimensional and continuous, discarding both alternatives so far. Finally, LSH is designed to handle high-dimensional, continuous (or discrete) data, being the most suited alternative. Similar to K-D trees, LSH may miss nearest neighbors, but this may act as a prevention mechanism against outliers, as previously mentioned.

In particular, LSH considers only the subset of points that are within the same region as an evaluation point. The goal of LSH is then to generate identical hashes for points that are nearby, storing that information efficiently in a hash table. A simple, yet effective way of generating each key is by using random projections~\cite{slaney2008locality}, as described next.

\subsection{Hash generation with random projections}

To simulate the random projections, we generate $H$ random hyperplanes, with each hyperplane consisting of a random vector $h \sim \mathcal{N}_d(0,1)$, where $d$ is the data dimension, and a random variable $b \sim \mathcal{U}(0,1)$ in the form of:

\begin{equation}
    h_0x_0 + \ldots + h_{d-1}x_{d-1} + b = 0.
\end{equation}

Each bit of a $d$-dimensional point $\phi$'s key is the sign of the random projection with each random hyperplane, obtained by a dot product:

\begin{equation}
    \text{hash}_{h,b}(\phi) = 
    \begin{cases}
        1,& \text{if } \underbrace{\text{sign}(\phi \cdot h + b)}_{\in \{-1, 0, 1\}} \geq 0\\
        0,              & \text{otherwise.}
    \end{cases}
\end{equation}

In other words, we are checking on which "side" of a random hyperplane the point $\phi$ lies on. The final $\phi$'s key consists of $H$ bits, obtained by comparing $\phi$ with all $H$ random hyperplanes:

\begin{equation}
    \text{key}_H(\phi) = \text{hash}_{h_0,b_0}(\phi)  \ldots \text{hash}_{h_{H-1},b_{H-1}}(\phi),
\end{equation}

where $h_{0,\ldots,H}$ and $b_{0,\ldots,H}$, are independently generated and each bit/hash is computed in parallel. 


We store this information in an initialized hash table $\text{HT}_\Phi$, which is populated by iterating through a set $\Phi$:

\begin{equation}
    \sum_{\phi \in \Phi} \text{HT}_\Phi[\text{key}_H(\phi)]\text{.append}(\phi).
\end{equation}

\subsection{Assessing the quality and diversity of a set}

We propose two novel metrics to evaluate the quality and diversity of a generated set in terms of precision (P) and recall (R), respectively. The simplest of which solely uses LSH (Section~\ref{sec:lsh}), calculating the ratio of evaluation samples with at least one reference sample in their region. Moreover, we propose an additional metric that uses LSH and KNN (Section~\ref{sec:lsh_knn}), calculating the ratio of evaluation samples inside at least one reference sample's hypersphere in their region.

\subsubsection{LSH metric}
\label{sec:lsh}

Consider the binary function $f_\text{LSH}(\text{key}, \text{HT}_\Phi)$, which returns if a key is in the list of keys of a populated hash table $\text{HT}_\Phi$:

\begin{equation}
    f_\text{LSH}(\text{key}, \text{HT}_\Phi) = 
    \begin{cases}
        1,& \text{if } \text{key} \in \text{HT}_\Phi\\
        0,              & \text{otherwise.}
    \end{cases}
\end{equation}

Given two sets $\Phi_a$ and $\Phi_b$, we assess quality in terms of precision:

\begin{equation}
    \text{P}_\text{LSH}(\Phi_a, \Phi_b) = \dfrac{1}{\vert\Phi_b\vert} \sum_{\phi_b \in \Phi_b} f_\text{LSH}(\text{key}_H(\phi_b), \text{HT}_{\Phi_{a}}),
\end{equation}

and diversity in terms of recall:

\begin{equation}
    \text{R}_\text{LSH}(\Phi_a, \Phi_b) = \dfrac{1}{\vert\Phi_a\vert} \sum_{\phi_a \in \Phi_a} f_\text{LSH}(\text{key}_H(\phi_a), \text{HT}_{\Phi_{b}}).
\end{equation}

\subsubsection{LSH + KNN metric}
\label{sec:lsh_knn}

To ease method comparisons, we follow Kynkäänniemi \etal~\cite{impar}'s formulations and describe a binary function that indicates if a given point $\phi$ is located within the volume formed by all the hyperspheres of each point $\phi^\prime \in \Phi$ and its $k$-nearest point in the set $\Phi$, returned by NN$_k(\phi^\prime,\Phi)$:

\begin{equation}
    \begin{split}
    f_\text{LSH+KNN}(\phi, \Phi) =\\=
    \begin{cases}
        1,& \text{if } \lVert \phi - \phi^\prime \rVert_2 \leq \lVert \phi - \text{NN}_k(\phi^\prime, \Phi) \rVert_2\\ & \text{for at least one }\phi^\prime \in \Phi\\
        0,              & \text{otherwise.}
    \end{cases}
    \end{split}
\end{equation}

Note that, if the number of neighbors is equal or higher than the number of samples in a given region, \textit{i.e.} $k \geq \vert\Phi\vert$, then the farthest point from $\phi^\prime$ in its region is retrieved by NN$_k(\phi^\prime,\Phi)$.

Given two sets $\Phi_a$ and $\Phi_b$, we assess quality in terms of precision:

\begin{equation}
    \begin{split}
    \text{P}_\text{LSH+KNN}(\Phi_a, \Phi_b)  =\\ \dfrac{1}{\vert\Phi_b\vert} \sum_{\phi_b \in \Phi_b} f_\text{LSH+KNN}(\phi_b, \text{HT}_{\Phi_{a}}(\text{key}_H(\phi_b))),
    \end{split}
\end{equation}

and diversity in terms of recall:

\begin{equation}
    \begin{split}
    \text{R}_\text{LSH+KNN}(\Phi_a, \Phi_b) =\\ \dfrac{1}{\vert\Phi_a\vert} \sum_{\phi_a \in \Phi_a} f_\text{LSH+KNN}(\phi_a, \text{HT}_{\Phi_{b}}(\text{key}_H(\phi_a))).
    \end{split}
\end{equation}

\subsection{Assessing the realism of individual samples}

Kynkäänniemi \etal~\cite{impar} further proposed a \textit{realism score (RS)} to calculate the quality of a given sample. Such metric can be modified and be applied to our LSH + KNN metric by:

\begin{equation}
    \begin{split}
    RS(\phi_a, \Phi_b)=\\\max_{\phi_b}\bigg\{\dfrac{\lVert \phi_b - \text{NN}_k(\phi_b,\text{HT}_{\Phi_b}(\text{key}_H(\phi_b))) \rVert_2}{\lVert \phi_a - \phi_b \rVert_2}\bigg\}.
    \end{split}
\end{equation}

Once again, the KNN is performed only with samples in $\Phi_b$ that are in the same region as $\phi_b$. More concretely, $RS \geq 1$ means that $\phi_a$ is inside the hypersphere of at least one $\phi_b$ in its region.

Notably, Kynkäänniemi \etal~\cite{impar} tried to mitigate the aforementioned outlier problem and computed this individual score by discarding half the samples in $\Phi_b$ with the largest radii. Even though this was a step in the right direction, no samples were discarded on their calculation of precision and recall when evaluating a sample set. 

\subsection{Complexity analysis}
\label{sec:complexity}

\begin{figure}
    \centering
    \includegraphics[width=0.47\textwidth]{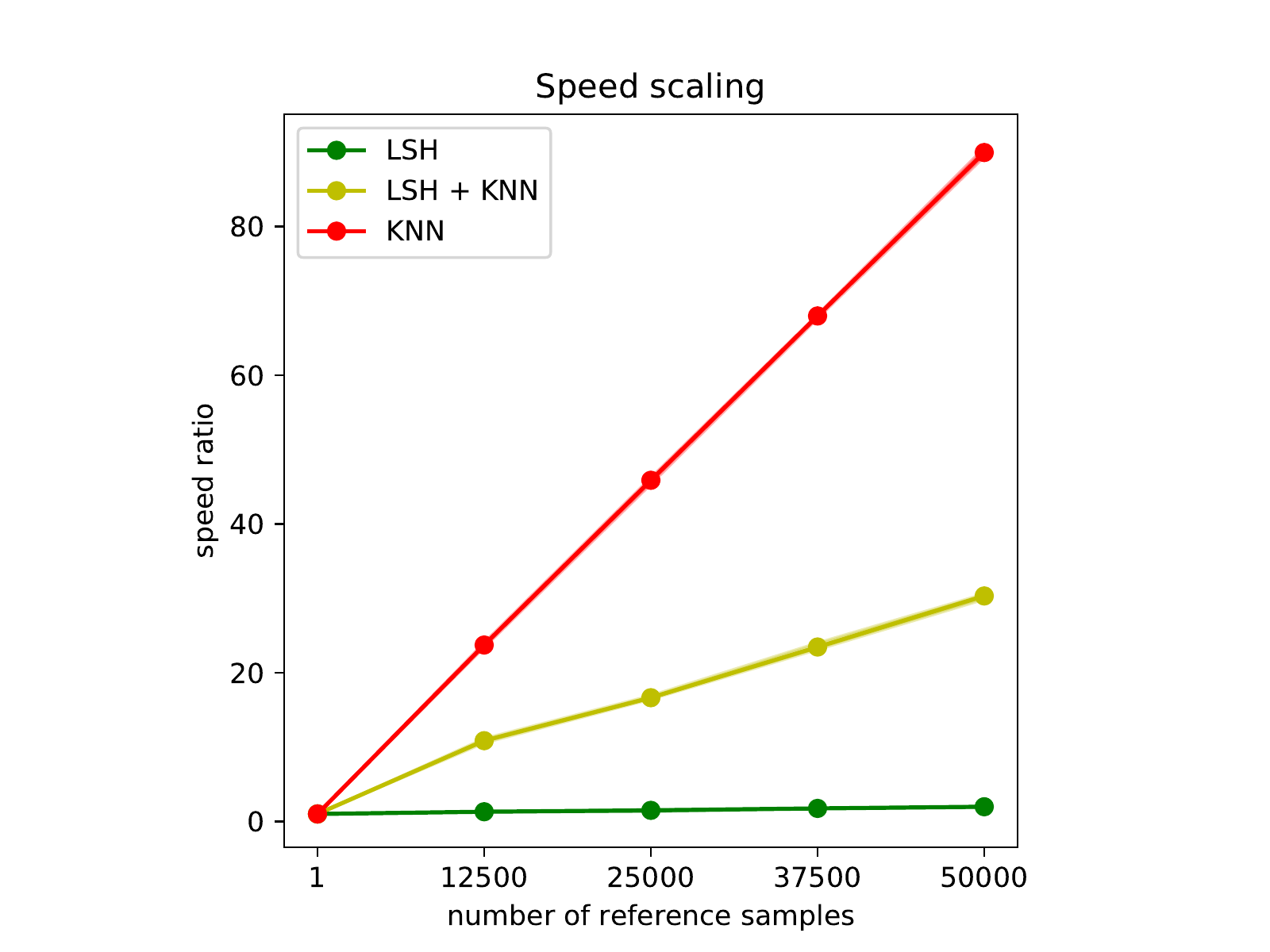}
    \caption{Scaling analysis on one NVIDIA GeForce GTX 1080 Ti GPU using 50k evaluation samples. Average, min and max, over 3 evaluation runs.}
    \label{fig:speed_scaling}
\end{figure}

Calculating the of a $d$-dimensional point costs $d \times H$, since $H$ dot products are required, \textit{i.e.} one for each random hyperplane. By generating $H$ random hyperplanes, we are splitting the data space into $\approx 2^H$ regions. Hence, on average, we will have $n/2^H$ points inside each region. Moreover, the comparison cost inside each region is, on average, $d \times n/2^H$. Note that, if we set $H \sim \log(n)$, we reduce the comparison cost to $d$, becoming constant in terms of $n$. Table~\ref{tab:complexity} summarizes the complexities in order of $n$ of the different evaluation methods. Figure~\ref{fig:speed_scaling} shows how the methods scale on practical hardware.

\begin{table}
\begin{center}
\resizebox{0.47\textwidth}{!}{\begin{tabular}{|l|c|c|}
\hline
Evaluation method & Computational cost & Complexity in $n$\\
\hline\hline
LSH (ours) & $d \times H$ & $O(1)$\\
LSH + KNN (ours) & $d \times H + d \times n / 2^H$ & $O(\log(n)), H \sim \log(n)$\\
KNN~\cite{impar} & $d \times n$ & $O(n)$\\
\hline
\end{tabular}}
\end{center}
\caption{Comparison of the different methods in assessing a $d$-dimensional evaluation sample against $n$ reference samples. $H$ denotes the number of random hyperplanes.}
\label{tab:complexity}
\end{table}

\section{Compression techniques}

Wang \etal \cite{qgan} studied the effects of linear quantization schemes while training GANs. In our work, we study different compression techniques other than quantization, such as pruning, splitting, and clipping while compressing GANs post-training and without any retraining or fine-tuning. Zhao \etal~\cite{ocs} performed similar studies for image classification networks\footnote{\url{https://github.com/cornell-zhang/dnn-quant-ocs}}. Here, we extrapolate such studies to the GANs domain, compressing the weights of several pre-trained generators on multiple datasets.

\subsection{Quantization (Q)}

Our linear quantization scheme (LQ) simply maps the floating-point weights to their closest discrete value covering their symmetric dynamic range:

\begin{equation}
    LQ(x) = \round{\dfrac{x (2^{b-1}-1)}{\text{max}(\vert x \vert)}}\times\dfrac{\text{max}(\vert x \vert)}{2^{b-1}-1},
\end{equation}
where $b$ represents the number of bits.
This is considered one of the most straightforward quantization schemes. Since it uses the full dynamic range of the original weights, LQ may be sensitive to outlier weights, which may hurt the final weight approximation.

\subsection{Pruning and quantization (P + Q)}

Mordido \etal~\cite{mcq,mcgq} leveraged Monte Carlo techniques to perform pruning and quantization in one algorithm called Monte Carlo Quantization (MCQ). In sum, they use the probability density function (PDF) of the floating-point weight distribution to generate a cumulative density function (CDF). Then, they perform importance sampling on the CDF using $N$ uniformly distributed samples with the assigned discrete value for each weight being the number of times it got hit during the sampling process. Weights that are not hit, may then be pruned at the end. 

The bit-width and pruning levels are controlled by $N$, \textit{i.e.} the amount of sampling performed. Hence, more sampling likely leads to a better approximation of the original weights at the cost of higher bit-width and fewer pruned weights.

\subsection{Splitting and quantization (S + Q)}

Outlier Channel Splitting (OCS)~\cite{ocs} proposed to improve the approximation of the original weights by duplicating outlier weights and halving their values. Concretely, outlier weights are moved closer to the center of the distribution, reducing the dynamic range and improving the approximation of the original weights when applying linear quantization after the splitting process. Hence, this method introduces a new trade-off between quantization loss and final network size. Based on the reported results in their original paper, we used an \textit{expand ratio} of 0.01 in our experiments.

\subsection{Clipping and quantization (C + Q)}

Analytical Clipping for Integer Quantization (ACIQ)~\cite{aciq} clips the original weight distribution by approximating the optimal clip threshold that minimizes the mean-squared-error between the continuous and discrete weight distributions. To achieve this, it uses the statistics of a Gaussian or a Laplacian distribution, depending on which one is closer to the original weight distribution. 
Due to its clipping nature, ACIQ improves the final approximation by increasing outlier distortion.
Inspired by Zhao \etal~\cite{ocs}, we applied linear quantization on top of clipping in our experiments.

\section{Experiments}
\label{sec:experiments}

We experimented compressing GANs pre-trained on several datasets: Flickr-Faces-HQ (FFHQ~\cite{stylegan}), a high-quality image dataset of human faces, Animal Faces-HQ (AFHQ~\cite{choi2020starganv2}), a high-quality image dataset of cats, dogs, and wild life faces, and ImageNet~\cite{imagenet}, the large-scale dataset organized according to the WordNet hierarchy. We used different resolutions for each dataset: 1024x1024 and 256x256 for FFHQ, 512x512 for AFHQ, and 128x128 for ImageNet.

Moreover, we tested compressing the weights of different generators of popular GANs. Namely,
PA-GAN~\cite{pagan} applies progressive augmentation in $D$'s input during training to reduce overfitting. 
zCR-GAN~\cite{zcr} uses latent consistency regularization by augmenting $G$'s input with additional small-magnitude noise, stimulating diversity in the generated set. 
SN-GAN~\cite{sngan} applies spectral normalization to $D$'s weights to stabilize training.
SS-GAN~\cite{ssgan} adds an auxiliary rotation loss for the $G$ and $D$, combining adversarial training and self-supervision.
StyleGAN2~\cite{stylegan2} improves its predecessor StyleGAN~\cite{stylegan} by revamping $G$'s normalization, progressive growing and regularization. 
ADA~\cite{sg2ada} introduces adaptive discriminator augmentation to promote learning with limited data.
BigGAN~\cite{biggan} applies orthogonal regularization to $G$ to train class-conditional GANs with large-scale data while controlling the trade-off between sample quality and diversity.
All of the used pre-trained generators are publicly available \footnote{\url{https://tfhub.dev/deepmind/biggan-128/2}}\footnote{\url{https://nvlabs-fi-cdn.nvidia.com/stylegan2-ada/pretrained/}}\footnote{\url{https://nvlabs-fi-cdn.nvidia.com/stylegan2/networks/}}.

To alleviate possible negative compression effects, we do not compress the parameters in the last residual block of each network. However, these layers only hold a small percentage ($\approx$ 4 to 6\%) of the total number of parameters, depending on the network architecture and dataset resolution. We use $k=3$, as originally proposed by Kynkäänniemi \etal~\cite{impar}, and $H = \floor{\log(n)}$, as discussed in Section~\ref{sec:complexity}, in all of our experiments.

\subsection{Mean images}

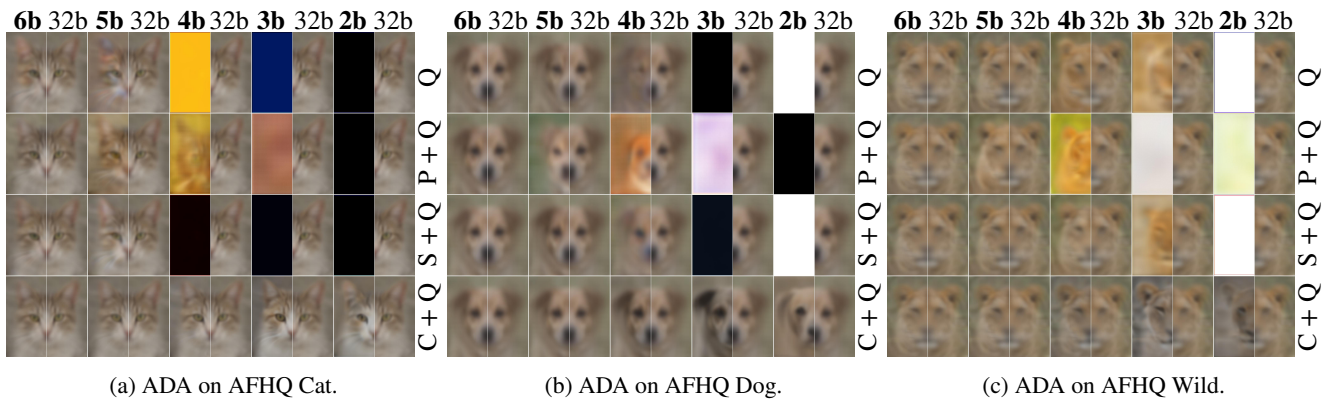
\begin{figure*}[!b]
\vspace{10px}
\centering
    \begin{subfigure}[b]{0.33\textwidth}
        \includegraphics[width=0.94\textwidth]{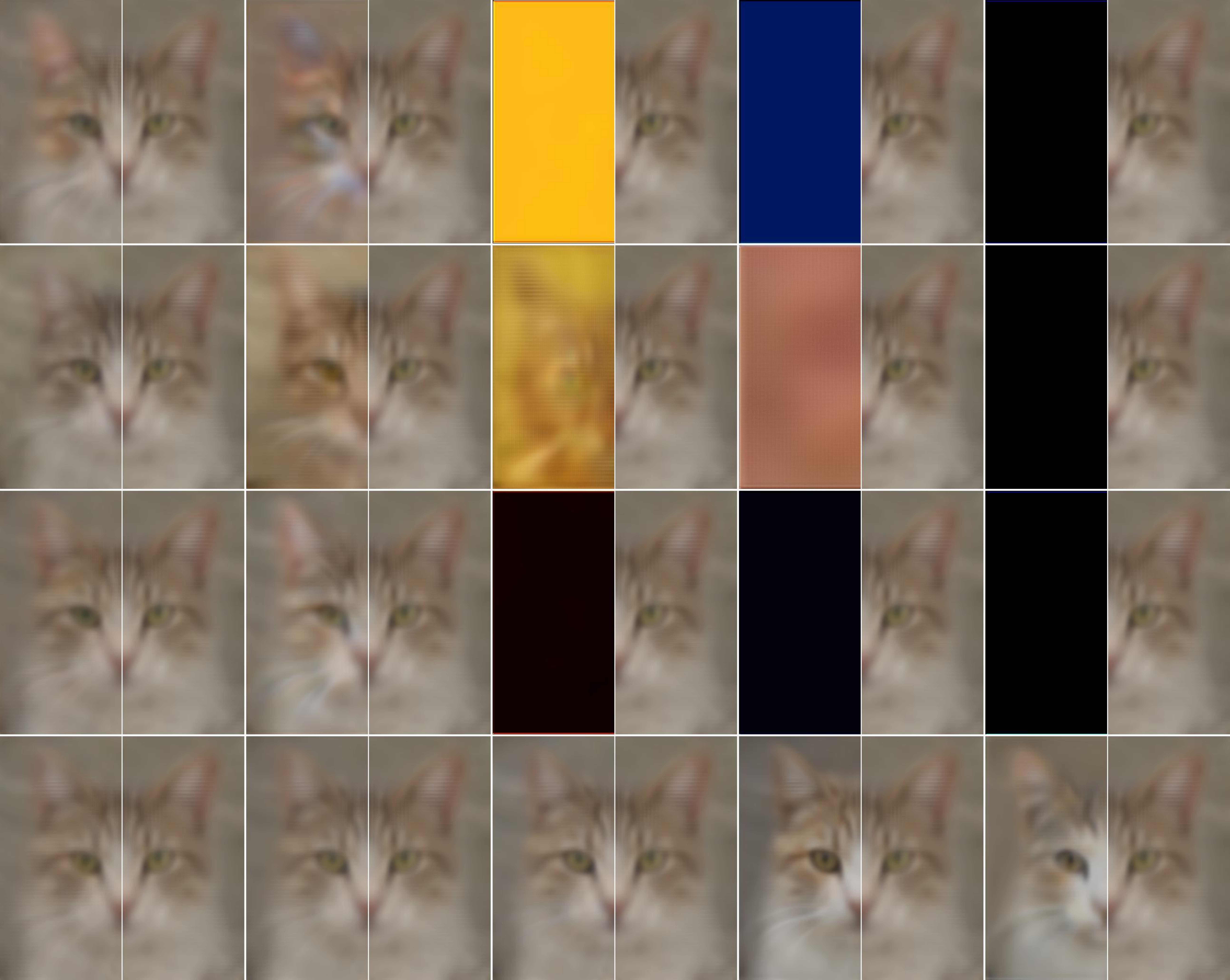}
        \begin{picture}(0,0)
            \put(-154,124){\bf6b}
            \put(-141,124){32b}
            \put(-123,124){\bf5b}
            \put(-110,124){32b}
            \put(-92,124){\bf4b}
            \put(-80,124){32b}
            \put(-62,124){\bf3b}
            \put(-48,124){32b}
            \put(-31,124){\bf2b}
            \put(-18,124){32b}
            \put(-1,102){\rotatebox{90}{Q}}
            \put(-1,65){\rotatebox{90}{P + Q}}
            \put(-1,34){\rotatebox{90}{S + Q}}
            \put(-1,2){\rotatebox{90}{C + Q}}
        \end{picture}     
        \caption{ADA on AFHQ Cat.}
        \label{fig:mean_images_2_to_6_bits_afhqcat_stylegan2_ada}
    \end{subfigure}
    \hfill
    \begin{subfigure}[b]{0.33\textwidth}
        \includegraphics[width=0.94\textwidth]{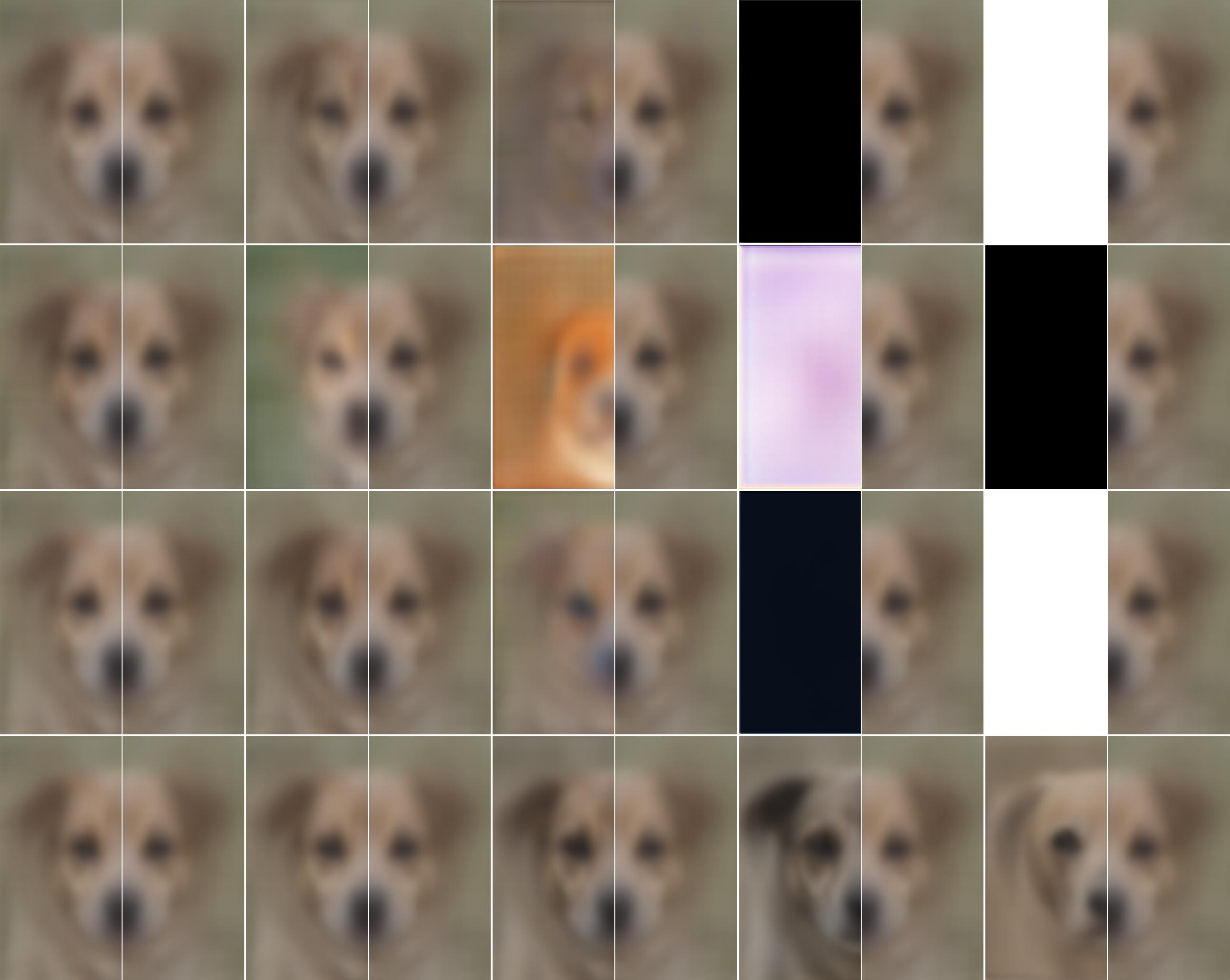}
        \begin{picture}(0,0)
            \put(-154,124){\bf6b}
            \put(-141,124){32b}
            \put(-123,124){\bf5b}
            \put(-110,124){32b}
            \put(-92,124){\bf4b}
            \put(-80,124){32b}
            \put(-62,124){\bf3b}
            \put(-48,124){32b}
            \put(-31,124){\bf2b}
            \put(-18,124){32b}
            \put(-1,102){\rotatebox{90}{Q}}
            \put(-1,65){\rotatebox{90}{P + Q}}
            \put(-1,34){\rotatebox{90}{S + Q}}
            \put(-1,2){\rotatebox{90}{C + Q}}
        \end{picture}         
        \caption{ADA on AFHQ Dog.}
        \label{fig:mean_images_2_to_6_bits_afhqdog_stylegan2_ada}
    \end{subfigure}
    \hfill
    \begin{subfigure}[b]{0.33\textwidth}
        \includegraphics[width=0.94\textwidth]{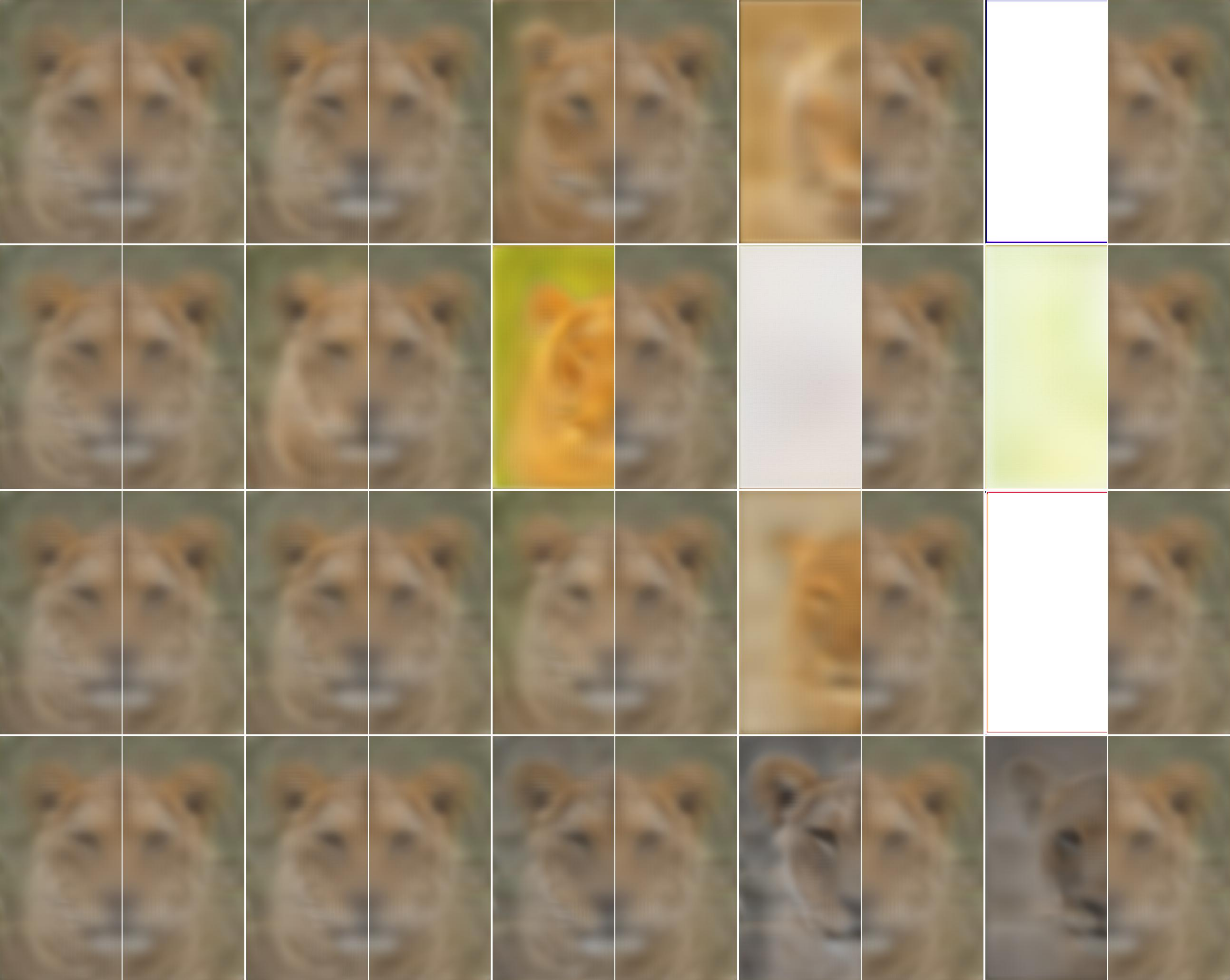}
        \begin{picture}(0,0)
            \put(-154,124){\bf6b}
            \put(-141,124){32b}
            \put(-123,124){\bf5b}
            \put(-110,124){32b}
            \put(-92,124){\bf4b}
            \put(-80,124){32b}
            \put(-62,124){\bf3b}
            \put(-48,124){32b}
            \put(-31,124){\bf2b}
            \put(-18,124){32b}
            \put(-1,102){\rotatebox{90}{Q}}
            \put(-1,65){\rotatebox{90}{P + Q}}
            \put(-1,34){\rotatebox{90}{S + Q}}
            \put(-1,2){\rotatebox{90}{C + Q}}
        \end{picture}     
        \caption{ADA on AFHQ Wild.}
        \label{fig:mean_images_2_to_6_bits_afhqwild_stylegan2_ada}
    \end{subfigure}
    \vspace{-15px}
    \caption{ADA mean 10k images on AFHQ Cat, AFHQ Dog, and AFHQ Wild (512x512). Compression by quantizing (Q), pruning (P), splitting (S), and clipping (C).}
    \label{fig:mean_images_2_to_6_bits_afhq}
\end{figure*}

\begin{figure}
    \vspace{10px}
    \includegraphics[width=0.47\textwidth]{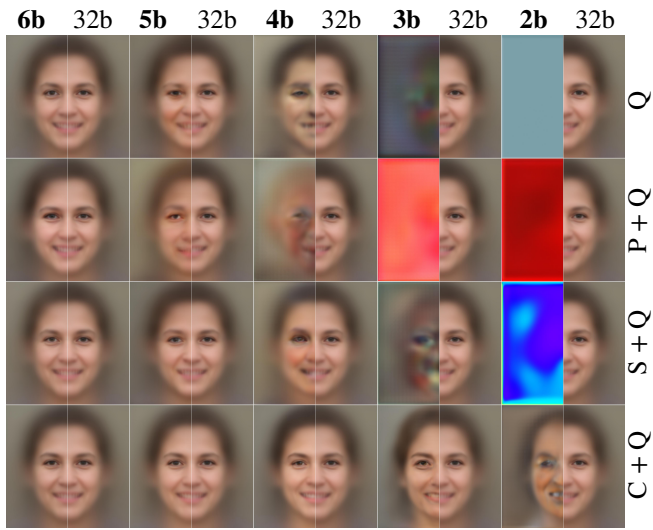}
    \begin{picture}(0,0)
        \put(-232,189){\bf6b}
        \put(-211,189){32b}
        \put(-186,189){\bf5b}
        \put(-164,189){32b}
        \put(-138,189){\bf4b}
        \put(-117,189){32b}
        \put(-90,189){\bf3b}
        \put(-69,189){32b}
        \put(-42,189){\bf2b}
        \put(-21,189){32b}
        \put(-1,158){\rotatebox{90}{Q}}
        \put(-1,104){\rotatebox{90}{P + Q}}
        \put(-1,58){\rotatebox{90}{S + Q}}
        \put(-1,11){\rotatebox{90}{C + Q}}
    \end{picture}         
    \caption{StyleGAN2 mean 10k images on FFHQ 1024x1024. Compression by quantizing (Q), pruning (P), splitting (S), and clipping (C).}
    \label{fig:mean_images_2_to_6_bits_ffhq1024_stylegan2}
\end{figure}

We will first study the effects that the different compression schemes, \textit{i.e.} quantization, pruning, splitting, and clipping, have on the different models and datasets.
Overall, in low-bit settings (2 to 4 bits), combining clipping and quantization is better than the alternatives, enabling the highly compressed models to still generate realistic samples (see Figures~\ref{fig:mean_images_2_to_6_bits_afhq} and~\ref{fig:mean_images_2_to_6_bits_ffhq1024_stylegan2}). This suggests a high tolerance to outlier weight distortion in $G$ induced by clipping.
In mid-bit settings (5 to 6 bits), all of the studied compression techniques are able to successfully compress the different models, with none of them showing no apparent loss of performance at 6 bits. Mean images for the different models on FFHQ 256x256 share similar conclusions and are provided in the appendix. 

One interesting observation in the mean generated images when applying clipping and quantization (C+Q) at low bit-widths, is that they still resemble faces both on the FFHQ and AFHQ datasets. This suggests that the generated set may be losing diversity, due to the average face being different than the 32-bit baselines, but still maintaining some of its original quality. We will now take a deeper diving into analyzing the effects of clipping and quantization. (Results for the rest compression techniques are in the appendix.)

\begin{figure}
\vspace{10px}
\centering
    \begin{subfigure}[b]{0.15\textwidth}
        \centering
        \includegraphics[trim={1050px 0 0 780px},clip,width=\textwidth]{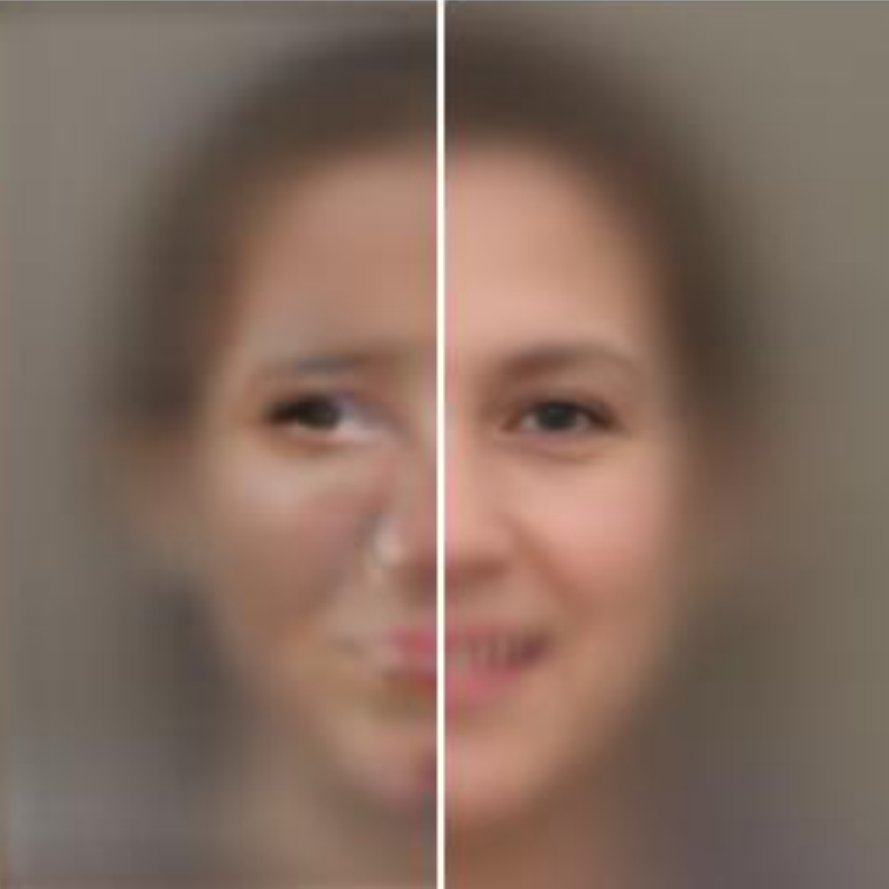}
        \begin{picture}(0,0)
            \put(5,88){32 bits}
            \put(-28,88){2 bits}
        \end{picture}     
        \vspace{-13px}    
        \caption{ADA.}
        \label{fig:mean_images_32_to_2_bits_ffhq_ada}
    \end{subfigure}
    \hfill
    \begin{subfigure}[b]{0.15\textwidth}
        \centering
        \includegraphics[trim={1050px 0 0 780px},clip,width=\textwidth]{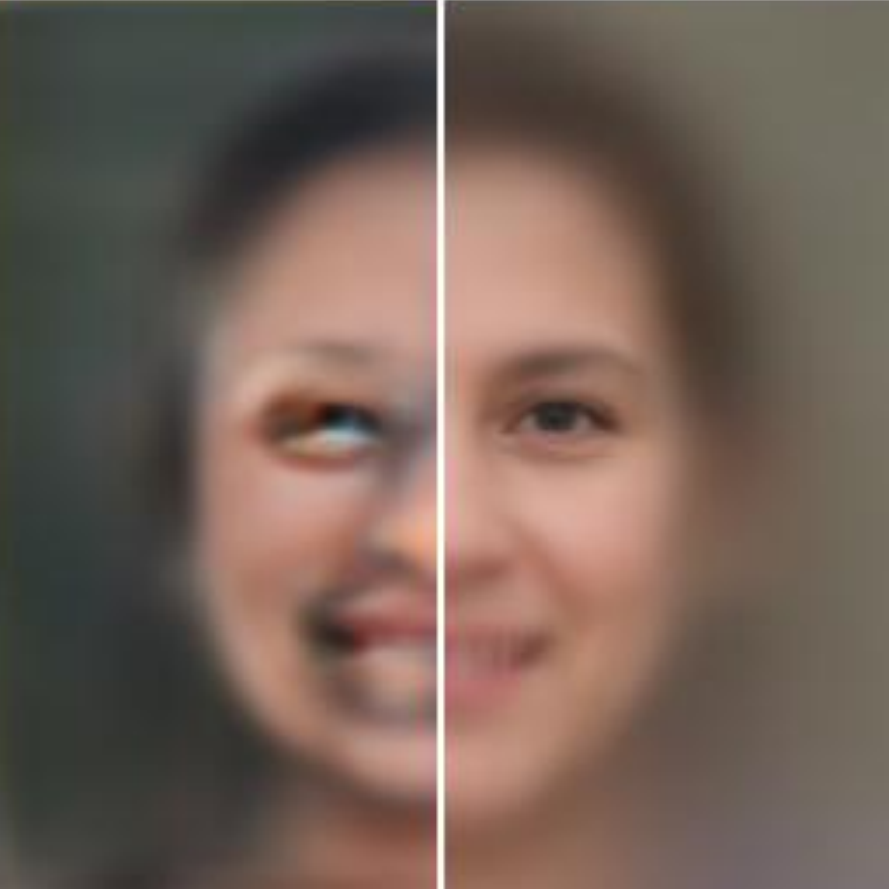}
        \begin{picture}(0,0)
            \put(5,88){32 bits}
            \put(-28,88){2 bits}
        \end{picture}     
        \vspace{-13px}    
        \caption{StyleGAN2.}
        \label{fig:mean_images_32_to_2_bits_ffhq_stylegan2}
    \end{subfigure}
    \hfill
    \begin{subfigure}[b]{0.15\textwidth}
        \centering
        \includegraphics[trim={1050px 0 0 780px},clip,width=\textwidth]{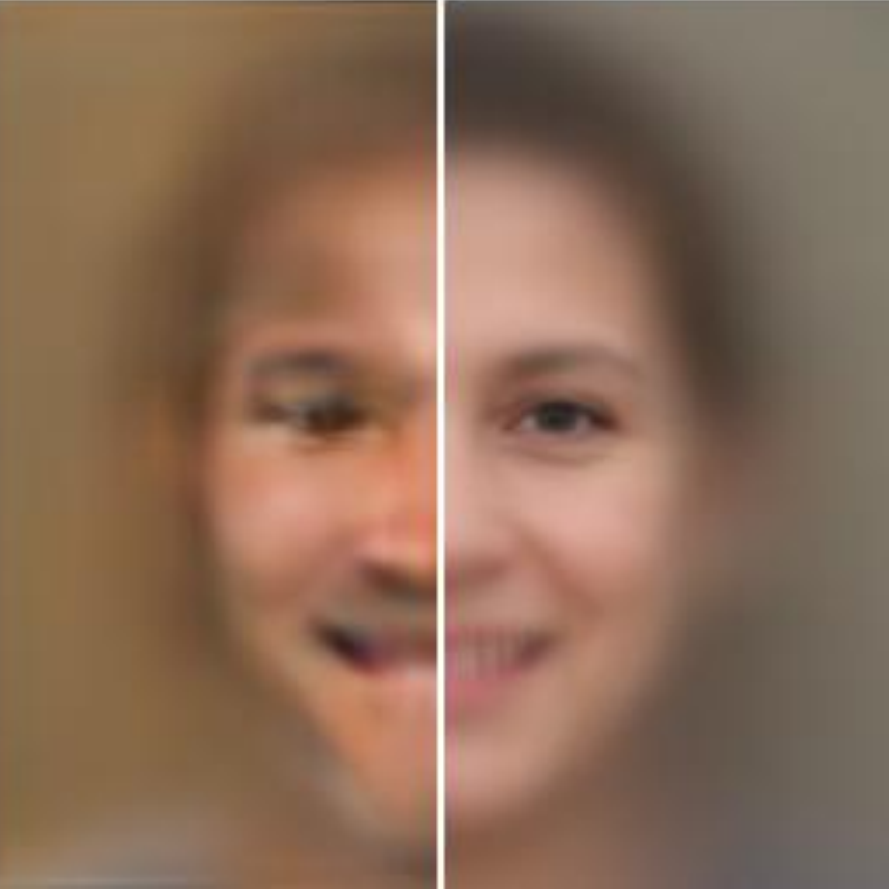}
        \begin{picture}(0,0)
            \put(5,88){32 bits}
            \put(-28,88){2 bits}
        \end{picture}     
        \vspace{-13px}    
        \caption{zCR-GAN.}
        \label{fig:mean_images_32_to_2_bits_ffhq_zcr}
    \end{subfigure}
    \hfill
    \vspace{10px}
    \begin{subfigure}[b]{0.15\textwidth}
        \centering
        \includegraphics[trim={1050px 0 0 780px},clip,width=\textwidth]{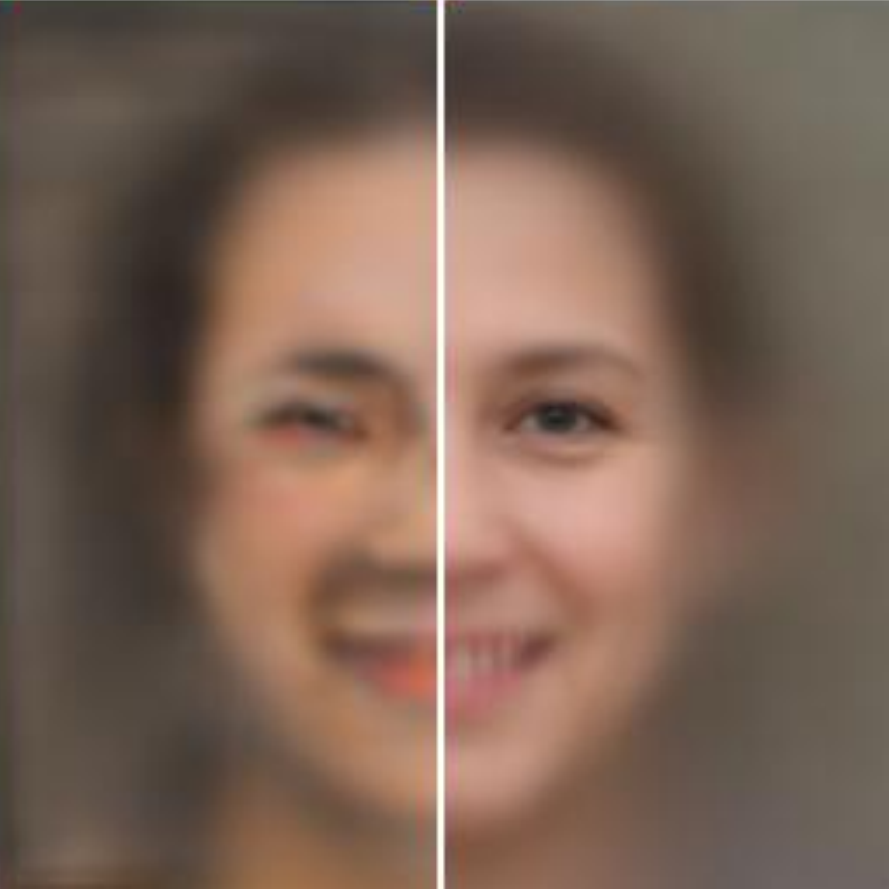}
        \begin{picture}(0,0)
            \put(5,88){32 bits}
            \put(-28,88){2 bits}
        \end{picture}     
        \vspace{-13px}    
        \caption{SN-GAN.}
        \label{fig:mean_images_32_to_2_bits_ffhq_sngan}
    \end{subfigure}
    \hfill
    \begin{subfigure}[b]{0.15\textwidth}
        \centering
        \includegraphics[trim={1050px 0 0 780px},clip,width=\textwidth]{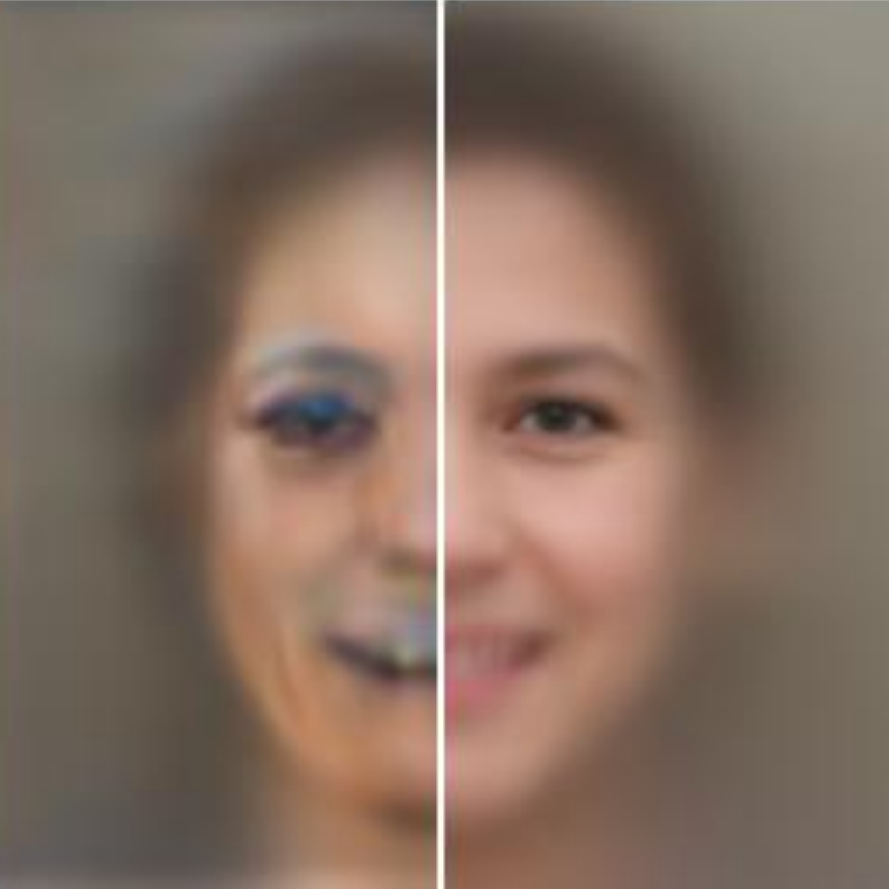}
        \begin{picture}(0,0)
            \put(5,88){32 bits}
            \put(-28,88){2 bits}
        \end{picture}     
        \vspace{-13px}    
        \caption{PA-GAN.}
        \label{fig:mean_images_32_to_2_bits_ffhq_pagan}
    \end{subfigure}
    \hfill
    \begin{subfigure}[b]{0.15\textwidth}
        \centering
        \includegraphics[trim={1050px 0 0 780px},clip,width=\textwidth]{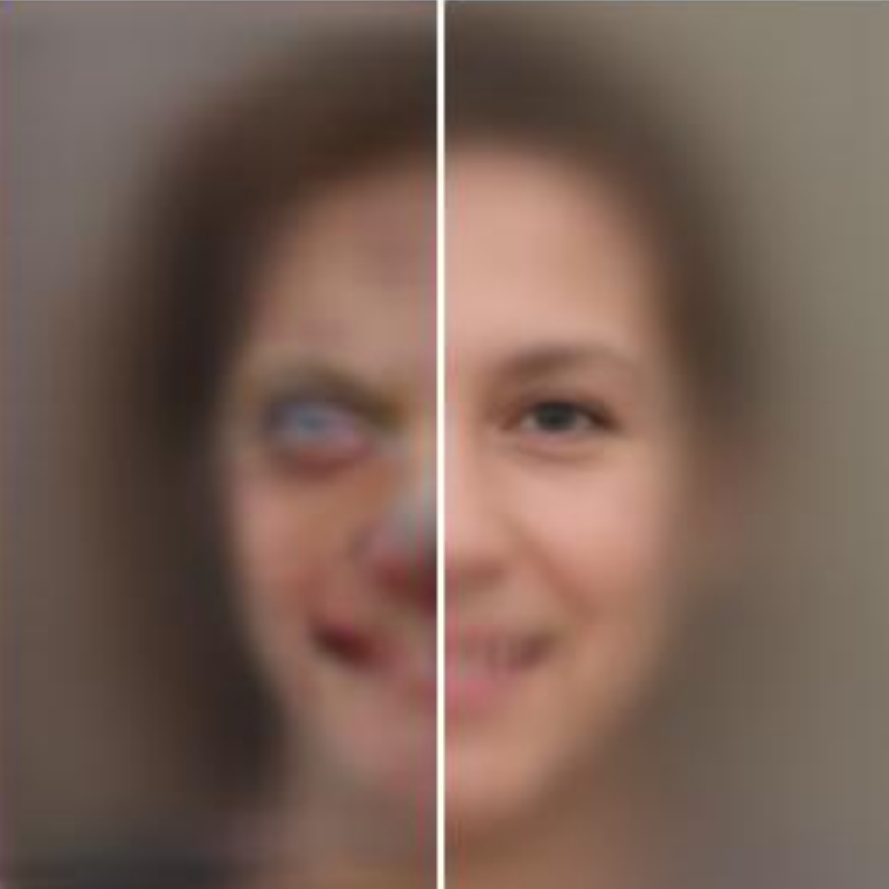}
        \begin{picture}(0,0)
            \put(5,88){32 bits}
            \put(-28,88){2 bits}
        \end{picture}     
        \vspace{-13px}    
        \caption{SS-GAN.}
        \label{fig:mean_images_32_to_2_bits_ffhq_ssgan}
    \end{subfigure}
    \hfill
    \caption{Mean 10k images FFHQ 256x256 (C + Q).}
    \label{fig:mean_images_32_to_2_bits_ffhq}
\end{figure}
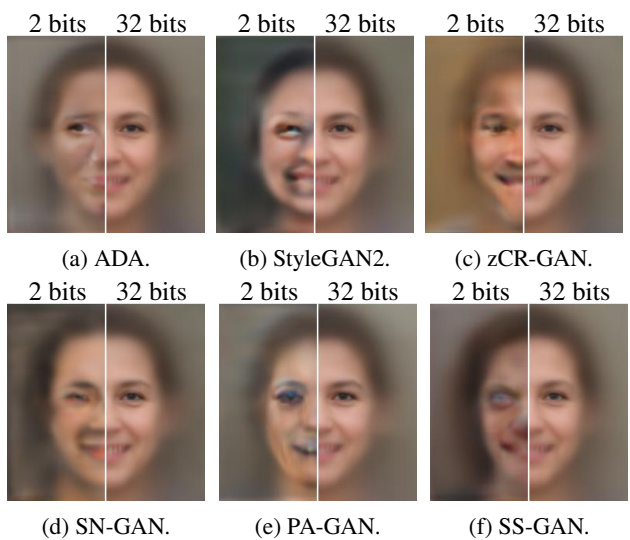

We may also analyze how the pre-trained GANs trained on FFHQ 256x256 differ from one another by visualizing their mean images (Figure~\ref{fig:mean_images_32_to_2_bits_ffhq}). Looking at the mean 32-bit image of the different models, we observe that all models successfully learn how to generated faces with similar attributes. However, at 2 bits, we start seeing differences between the mean images, which different models generating different types of face attributes, such as elder (Figure~\ref{fig:mean_images_32_to_2_bits_ffhq_pagan}), longer hair (Figures~\ref{fig:mean_images_32_to_2_bits_ffhq_ada} and~\ref{fig:mean_images_32_to_2_bits_ffhq_ssgan}), darker hair (Figure~\ref{fig:mean_images_32_to_2_bits_ffhq_stylegan2} and~\ref{fig:mean_images_32_to_2_bits_ffhq_sngan}), and male (Figure~\ref{fig:mean_images_32_to_2_bits_ffhq_zcr}).

\subsection{Generated images}

\begin{figure}
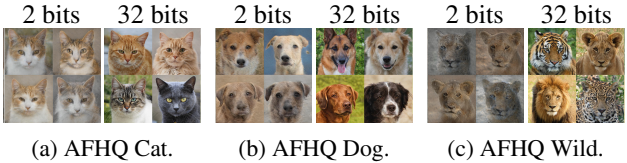

\vspace{10px}
\centering
    \begin{subfigure}[b]{0.15\textwidth}
        \includegraphics[trim={5140px 0 0 3100px},clip,width=0.488\textwidth]{images/fake_images_2_to_6_bits/afhqcat-mirror-paper512-ada-lq-mcq-ocs-aciq.pdf}
        \hfill
        \includegraphics[trim={0 0 5140px 3100px},clip,width=0.488\textwidth]{images/fake_images_2_to_6_bits/afhqcat-mirror-paper512-ada-lq-mcq-ocs-aciq.pdf}
        \begin{picture}(0,0)
            \put(7,50){2 bits}
            \put(43,50){32 bits}
        \end{picture}     
        \vspace{-13px}    
        \caption{AFHQ Cat.}
        \label{fig:fake_images_32_to_2_bits_afhqcat_ada}
    \end{subfigure}
    \hfill
        \begin{subfigure}[b]{0.15\textwidth}
        \includegraphics[trim={5140px 0 0 3100px},clip,width=0.488\textwidth]{images/fake_images_2_to_6_bits/afhqdog-mirror-paper512-ada-lq-mcq-ocs-aciq.pdf}
        \hfill
        \includegraphics[trim={0 0 5140px 3100px},clip,width=0.488\textwidth]{images/fake_images_2_to_6_bits/afhqdog-mirror-paper512-ada-lq-mcq-ocs-aciq.pdf}
        \begin{picture}(0,0)
            \put(7,50){2 bits}
            \put(43,50){32 bits}
        \end{picture}     
        \vspace{-13px}    
        \caption{AFHQ Dog.}
        \label{fig:fake_images_32_to_2_bits_afhqdog_ada}
    \end{subfigure}
    \hfill
        \begin{subfigure}[b]{0.15\textwidth}
        \includegraphics[trim={5140px 0 0 3100px},clip,width=0.488\textwidth]{images/fake_images_2_to_6_bits/afhqwild-mirror-paper512-ada-lq-mcq-ocs-aciq.pdf}
        \hfill
        \includegraphics[trim={0 0 5140px 3100px},clip,width=0.488\textwidth]{images/fake_images_2_to_6_bits/afhqwild-mirror-paper512-ada-lq-mcq-ocs-aciq.pdf}
        \begin{picture}(0,0)
            \put(7,50){2 bits}
            \put(43,50){32 bits}
        \end{picture}
        \vspace{-13px}    
        \caption{AFHQ Wild.}
        \label{fig:fake_images_32_to_2_bits_afhqwild_ada}
    \end{subfigure}
    \hfill
    \caption{ADA on AFHQ 512x512 (C + Q).}
    \label{fig:fake_images_32_to_2_bits_affhq}
\end{figure}

\begin{figure}
\vspace{10px}
\centering
    \begin{subfigure}[b]{0.15\textwidth}
        \includegraphics[trim={2580px 0 0 1550px},clip,width=0.488\textwidth]{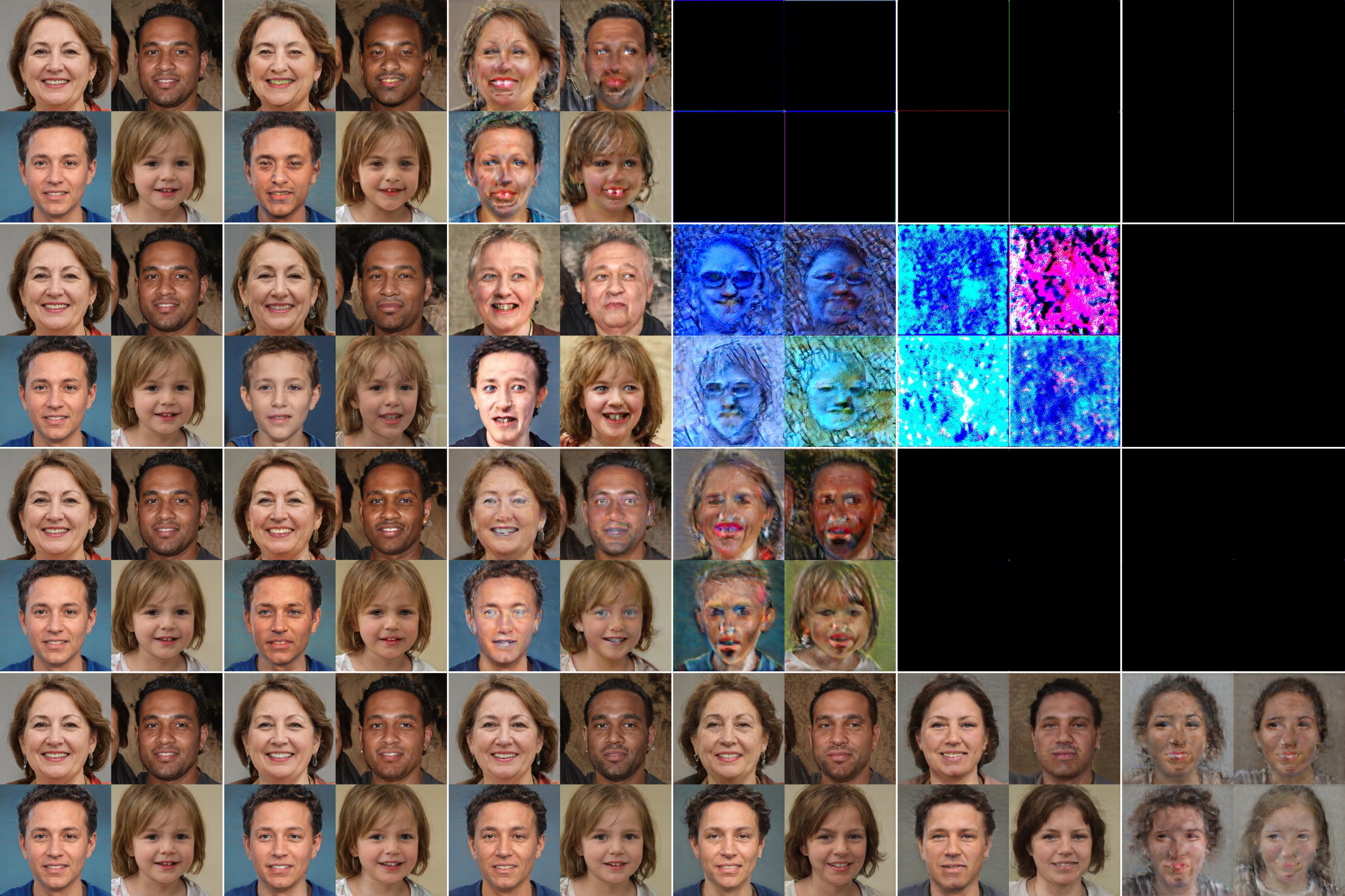}
        \hfill
        \includegraphics[trim={0 0 2580px 1550px},clip,width=0.488\textwidth]{images/fake_images_2_to_6_bits/ffhq140k-paper256-ada-lq-mcq-ocs-aciq.pdf}
        \begin{picture}(0,0)
            \put(7,50){2 bits}
            \put(43,50){32 bits}
        \end{picture}     
        \vspace{-13px}    
        \caption{ADA.}
        \label{fig:fake_images_32_to_2_bits_ffhq_ada}
    \end{subfigure}
    \hfill
        \begin{subfigure}[b]{0.15\textwidth}
        \includegraphics[trim={2580px 0 0 1550px},clip,width=0.488\textwidth]{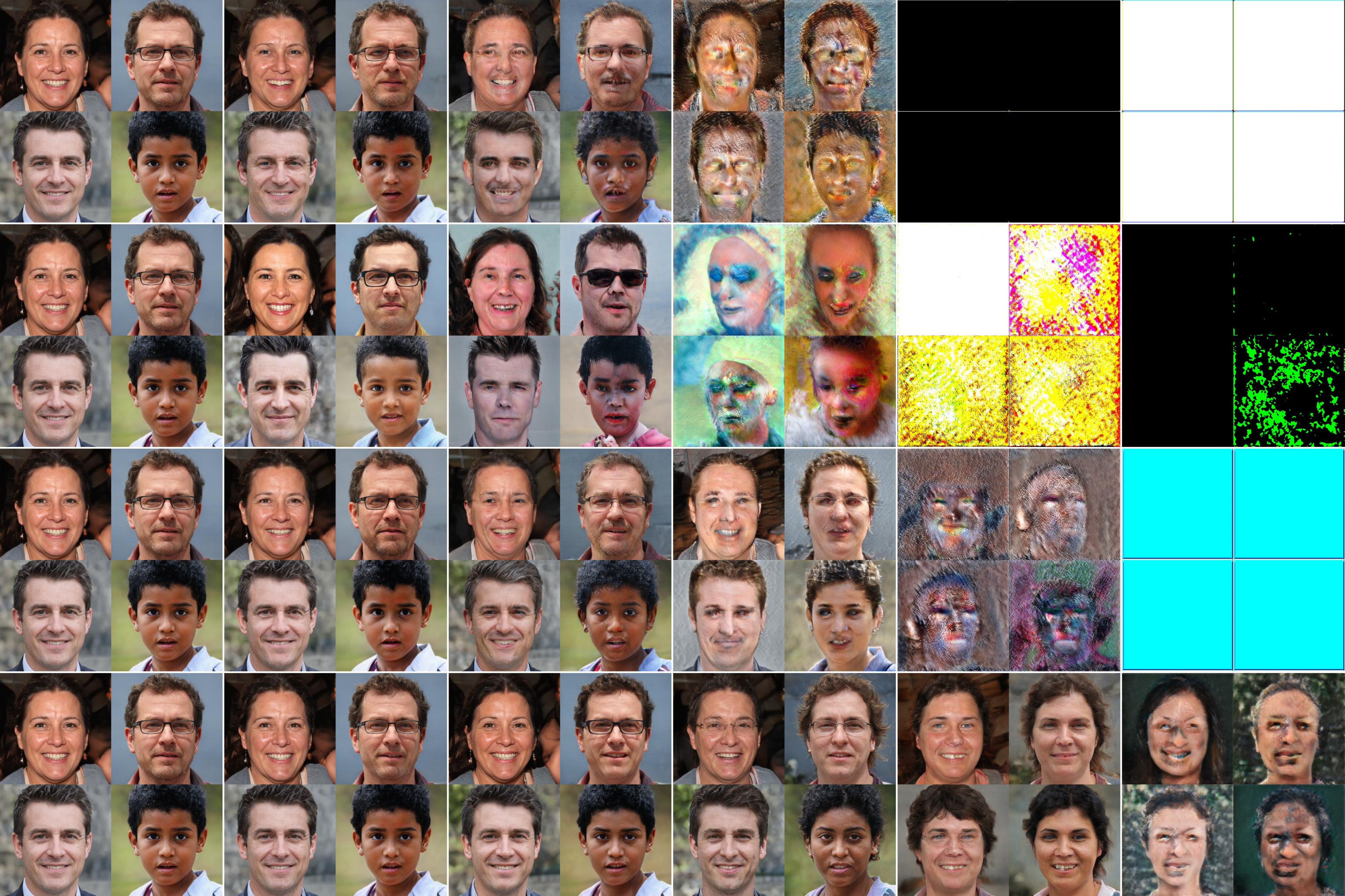}
        \hfill
        \includegraphics[trim={0 0 2580px 1550px},clip,width=0.488\textwidth]{images/fake_images_2_to_6_bits/ffhq140k-paper256-noaug-lq-mcq-ocs-aciq.pdf}
        \begin{picture}(0,0)
            \put(7,50){2 bits}
            \put(43,50){32 bits}
        \end{picture}     
        \vspace{-13px}    
        \caption{StyleGAN2.}
        \label{fig:fake_images_32_to_2_bits_ffhq_stylegan2}
    \end{subfigure}
    \hfill
        \begin{subfigure}[b]{0.15\textwidth}
        \includegraphics[trim={2580px 0 0 1550px},clip,width=0.488\textwidth]{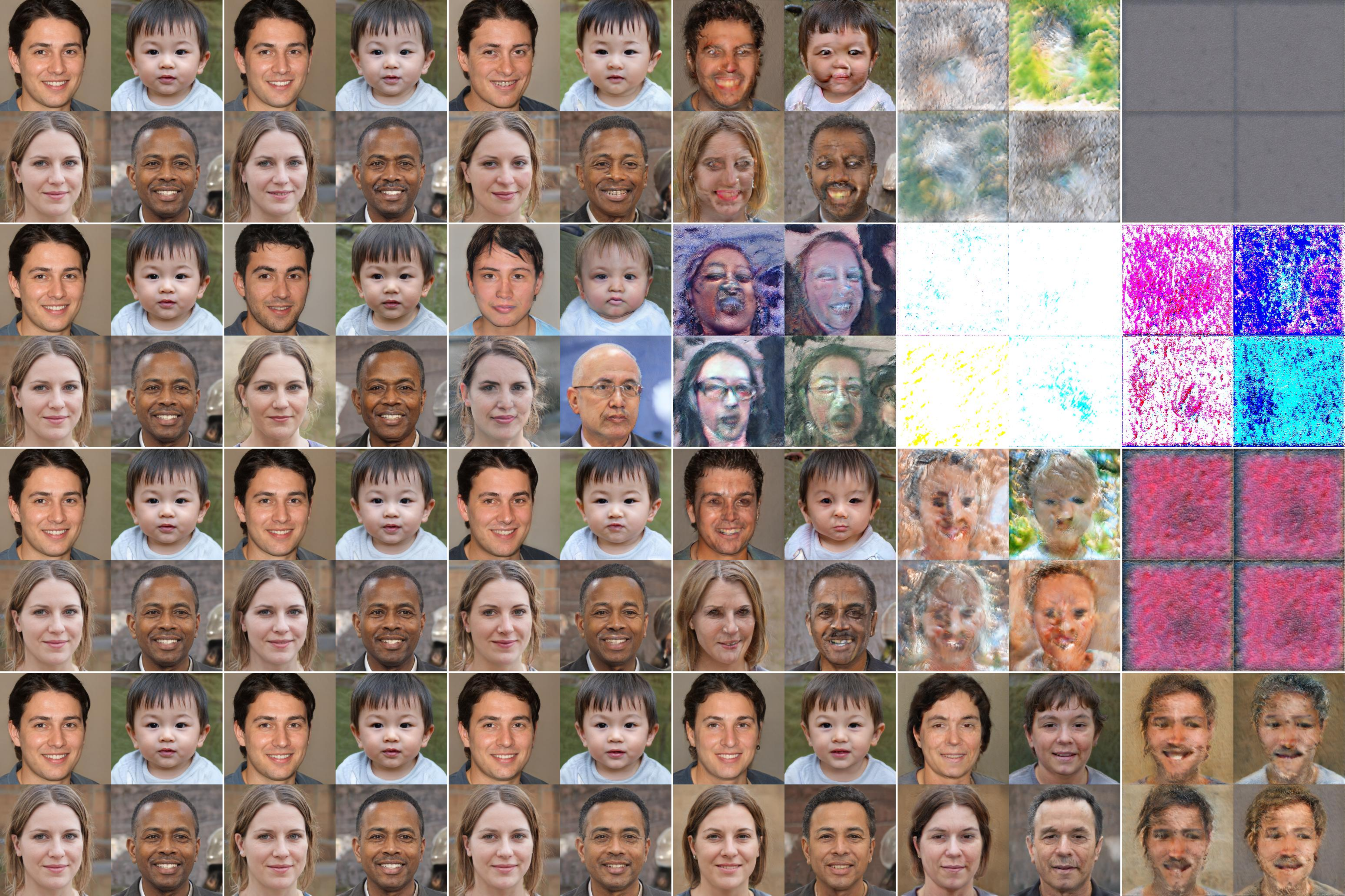}
        \hfill
        \includegraphics[trim={0 0 2580px 1550px},clip,width=0.488\textwidth]{images/fake_images_2_to_6_bits/ffhq140k-paper256-noaug-zcr-lq-mcq-ocs-aciq.pdf}
        \begin{picture}(0,0)
            \put(7,50){2 bits}
            \put(43,50){32 bits}
        \end{picture}     
        \vspace{-13px}    
        \caption{zCR-GAN.}
        \label{fig:fake_images_32_to_2_bits_ffhq_zcr}
    \end{subfigure}
    \hfill
    \vspace{10px}
        \begin{subfigure}[b]{0.15\textwidth}
        \includegraphics[trim={2580px 0 0 1550px},clip,width=0.488\textwidth]{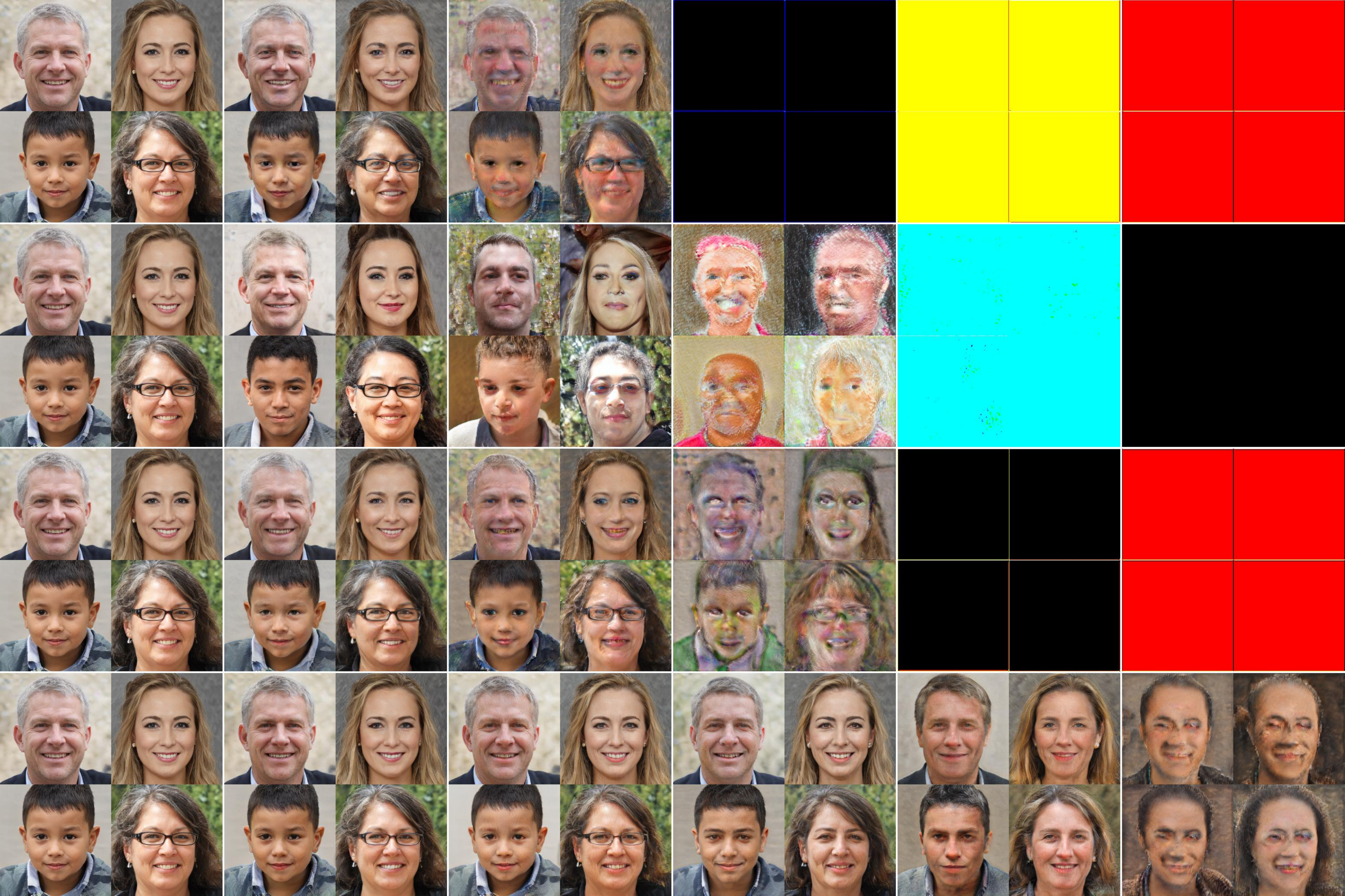}
        \hfill
        \includegraphics[trim={0 0 2580px 1550px},clip,width=0.488\textwidth]{images/fake_images_2_to_6_bits/ffhq140k-paper256-noaug-spectralnorm-lq-mcq-ocs-aciq.pdf}
        \begin{picture}(0,0)
            \put(7,50){2 bits}
            \put(43,50){32 bits}
        \end{picture}     
        \vspace{-13px}    
        \caption{SN-GAN.}
        \label{fig:fake_images_32_to_2_bits_ffhq_sngan}
    \end{subfigure}
    \hfill
        \begin{subfigure}[b]{0.15\textwidth}
        \includegraphics[trim={2580px 0 0 1550px},clip,width=0.488\textwidth]{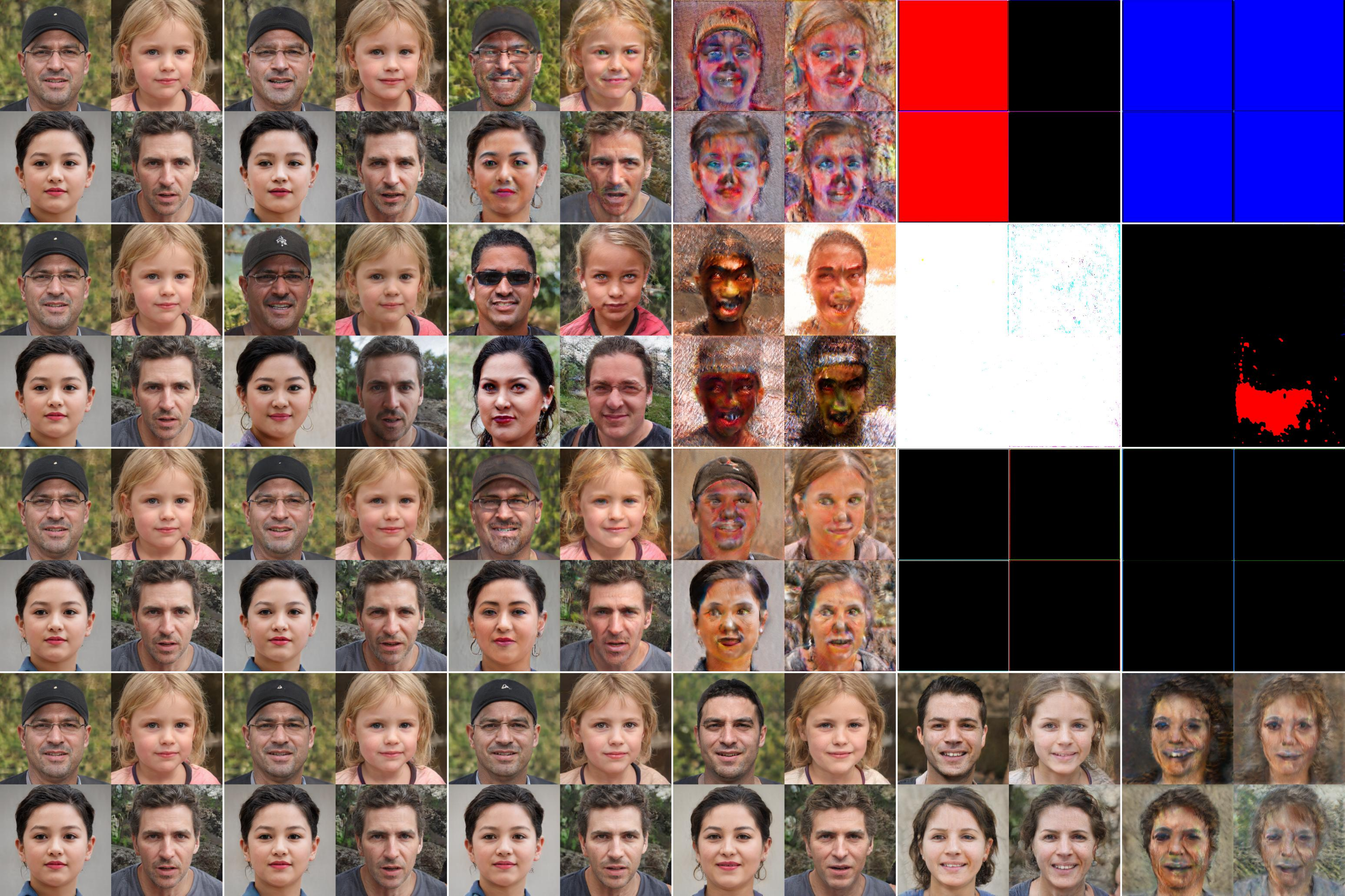}
        \hfill
        \includegraphics[trim={0 0 2580px 1550px},clip,width=0.488\textwidth]{images/fake_images_2_to_6_bits/ffhq140k-paper256-noaug-pagan-lq-mcq-ocs-aciq.pdf}
        \begin{picture}(0,0)
            \put(7,50){2 bits}
            \put(43,50){32 bits}
        \end{picture}     
        \vspace{-13px}    
        \caption{PA-GAN.}
        \label{fig:fake_images_32_to_2_bits_ffhq_pagan}
    \end{subfigure}
    \hfill  
    \begin{subfigure}[b]{0.15\textwidth}
        \includegraphics[trim={2580px 0 0 1550px},clip,width=0.488\textwidth]{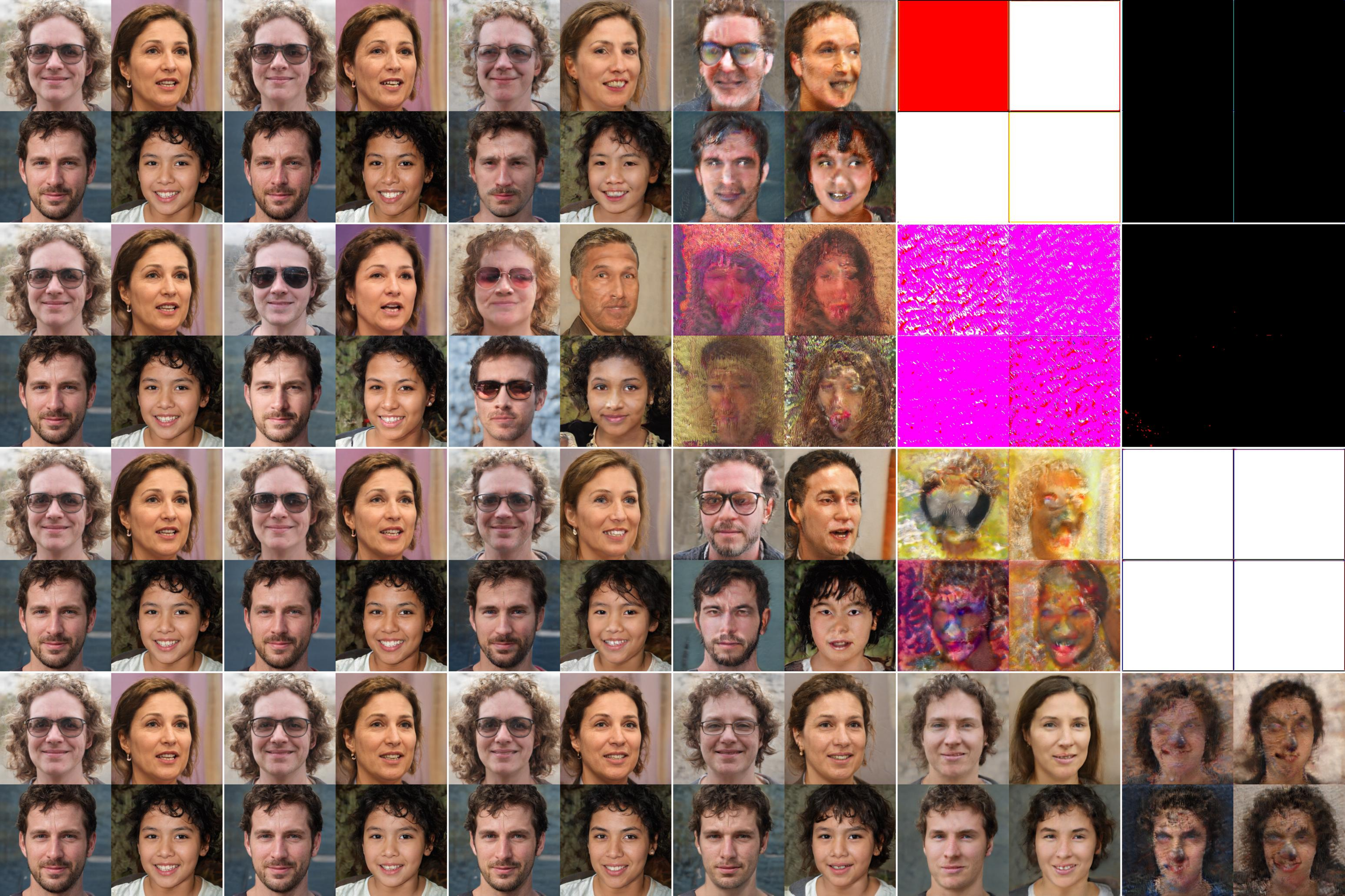}
        \hfill
        \includegraphics[trim={0 0 2580px 1550px},clip,width=0.488\textwidth]{images/fake_images_2_to_6_bits/ffhq140k-paper256-noaug-auxrot-lq-mcq-ocs-aciq.pdf}
        \begin{picture}(0,0)
            \put(7,50){2 bits}
            \put(43,50){32 bits}
        \end{picture}     
        \vspace{-13px}    
        \caption{SS-GAN.}
        \label{fig:fake_images_32_to_2_bits_ffhq_ssgan}
    \end{subfigure}
    \hfill
    \caption{FFHQ 256x256 (C + Q).}
    \label{fig:fake_images_32_to_2_bits_ffhq}
\end{figure}

To complement our previous discussions fueled by visualizing the mean images produced by the models, we will now take a look at some of their generated samples (see Figures~\ref{fig:fake_images_32_to_2_bits_affhq} and \ref{fig:fake_images_32_to_2_bits_ffhq}). All images were generated with the same 4 seeds (transformations at higher bits are in the appendix). At 2 bits, we notice a general lack of diversity in the generated set, however, all models are still able to produce recognizable samples retaining some of their original quality.

\begin{figure}
    \vspace{47px}
    \begin{picture}(0,0)
        \put(0,0){\includegraphics[trim={0 0 0 780px},clip,width=0.47\textwidth]{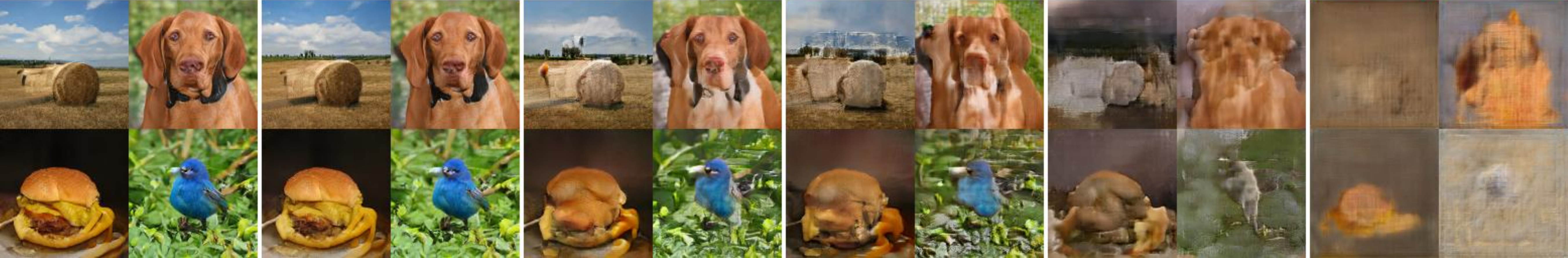}}
        \put(6,40){32 bits}
        \put(47,40){6 bits}
        \put(85,40){5 bits}
        \put(124,40){4 bits}
        \put(163,40){3 bits}
        \put(203,40){2 bits}
    \end{picture}      
    \caption{BigGAN on ImageNet 128x128 (C + Q).}
    \label{fig:fake_images_2_to_6_bits_imagenet_biggan}
\end{figure}
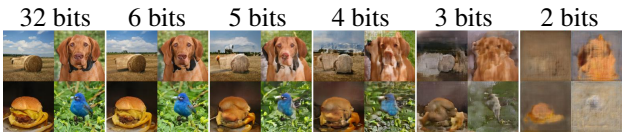

Figure~\ref{fig:fake_images_2_to_6_bits_imagenet_biggan} shows generated images from BigGAN on ImageNet. As we increase the compression levels, objects tend to get increasingly distorted while starting to share similar textures. (Generated images of all models using the other compression techniques can be found in the appendix.)

\subsection{Evaluation metrics}

\begin{table}
\centering
\resizebox{0.47\textwidth}{!}{
\begin{tabular}{|c|c||c|c||c|c||c|c||c|}
\hline
& & \multicolumn{2}{c||}{LSH~$\uparrow$} & \multicolumn{2}{c||}{LSH + KNN~$\uparrow$}  & \multicolumn{2}{c||}{KNN~\cite{impar}~$\uparrow$} & \\
\cline{3-8}
Network & b & $P$ & $R$ & $P$ & $R$ & $P$ & $R$ & FID~\cite{fid}~$\downarrow$\\
\hline\hline
& 32 & 0.949 & 0.934 & 0.830 & 0.766 & 0.689 & 0.493 & 2.8\\
\cline{2-9}
StyleGAN2 & 4 & \bf0.958 & 0.913 & \bf0.881 & 0.743 & \bf0.747 & 0.298 & 12.0\\
(FFHQ-1024) & 3 & 0.945 & 0.847 & \bf0.869 & 0.512 & 0.676 & 0.038 & 50.4\\
& 2 & 0.561 & 0.400 & 0.052 & 0.026 & 0.091 & 0.000 & 157.5\\
\hline\hline                                                 
& 32 & 0.929 & 0.919 & 0.763 & 0.784 & 0.681 & 0.442 & 3.8\\
\cline{2-9}
ADA & 4 & \bf0.941 & 0.906 & \bf0.811 & 0.701 & \bf0.740 & 0.285 & 10.1\\
(FFHQ-256) & 3 & \bf0.969 & 0.815 & \bf0.896 & 0.446 & \bf0.808 & 0.048 & 40.4\\
& 2 & 0.751 & 0.651 & 0.280 & 0.135 & 0.307 & 0.000 & 145.9\\
\hline                              
& 32 & 0.933 & 0.917 & 0.764 & 0.772 & 0.682 & 0.445 & 3.7\\
\cline{2-9}
StyleGAN2 & 4 & \bf0.940 & 0.902 & \bf0.813 & 0.715 & \bf0.802 & 0.220 & 14.4\\
(FFHQ-256) & 3 & \bf0.965 & 0.838 & \bf0.900 & 0.595 & \bf0.891 & 0.043 & 40.2\\               
& 2 & 0.730 & 0.626 & 0.352 & 0.229 & 0.371 & 0.000 & 143.4\\
\hline
& 32 & 0.934 & 0.918 & 0.769 & 0.741 & 0.680 & 0.473 & 3.3\\
\cline{2-9}
zCR-GAN & 4 & \bf0.960 & 0.893 & \bf0.861 & 0.696 & \bf0.766 & 0.300 & 12.6\\
(FFHQ-256) & 3 & \bf0.955 & 0.813 & \bf0.876 & 0.492 & \bf0.828 & 0.043 & 65.2\\                 
& 2 & 0.452 & 0.544 & 0.106 & 0.062 & 0.189 & 0.000 & 157.4\\
\hline
& 32 & 0.932 & 0.916 & 0.772 & 0.693 & 0.717 & 0.383 & 4.5\\
\cline{2-9}                    
SN-GAN & 4 & \bf0.936 & 0.894 & \bf0.819 & \bf0.694 & \bf0.744 & 0.208 & 14.1\\     
(FFHQ-256) & 3 & \bf0.956 & 0.840 & \bf0.866 & 0.442 & \bf0.727 & 0.032 & 49.1\\
& 2 & 0.612 & 0.634 & 0.246 & 0.133 & 0.206 & 0.000 & 140.7\\
\hline
& 32 & 0.932 & 0.918 & 0.764 & 0.693 & 0.683 & 0.449 & 3.8\\
\cline{2-9}
PA-GAN & 4 & \bf0.957 & 0.887 & \bf0.863 & 0.691 & \bf0.789 & 0.260 & 13.5\\
(FFHQ-256) & 3 & \bf0.969 & 0.812 & \bf0.900 & 0.454 & \bf0.797 & 0.041 & 50.2\\
& 2 & 0.425 & 0.376 & 0.071 & 0.027 & 0.214 & 0.000 & 164.8\\
\hline 
& 32 & 0.934 & 0.916 & 0.772 & 0.733 & 0.686 & 0.428 & 4.2\\
\cline{2-9}
SS-GAN & 4 & \bf0.945 & 0.897 & \bf0.827 & \bf0.787 & \bf0.755 & 0.254 & 11.4\\
(FFHQ-256) & 3 & \bf0.967 & 0.804 & \bf0.887 & 0.427 & \bf0.773 & 0.041 & 44.2\\
& 2 & 0.324 & 0.440 & 0.039 & 0.092 & 0.035 & 0.000 & 197.2\\
\hline\hline
& 32 & 0.748 & 0.715 & 0.505 & 0.270 & 0.858 & 0.149 & 10.8\\
\cline{2-9}
BigGAN & 4 & \bf0.816 & 0.588 & \bf0.677 & 0.240 & 0.687 & 0.014 & 44.1\\
(ImageNet) & 3 & \bf0.822 & 0.419 & \bf0.699 & 0.013 & 0.649 & 0.000 & 119.0\\
& 2 & 0.639 & 0.103 & \bf0.505 & 0.000 & \bf0.979 & 0.000 & 191.7\\
\hline
\end{tabular}
}
\caption{FID and precision (P) and recall (R) on FFHQ (1024x1024 and 256x256) and ImageNet with 50k real and generated samples. Average over 3 evaluation runs.}
 \label{tab:results_ffhq_imagenet}
\end{table}

\begin{table}
\centering
\resizebox{0.47\textwidth}{!}{
\begin{tabular}{|c|c||c|c||c|c||c|c||c|}
\hline
& & \multicolumn{2}{c||}{LSH~$\uparrow$} & \multicolumn{2}{c||}{LSH + KNN~$\uparrow$}  & \multicolumn{2}{c||}{KNN~\cite{impar}~$\uparrow$} & KID~\cite{kid}~$\downarrow$\\
\cline{3-8}
Network & b & $P$ & $R$ & $P$ & $R$ & $P$ & $R$ & $\times 10^3$\\
\hline\hline
& 32 & 0.940 & 0.917 & 0.714 & 0.753 & 0.765 & 0.532 & 0.7\\
\cline{2-9}
ADA & 4 & \bf0.946 & 0.901 & \bf0.752 & 0.697 & \bf0.841 & 0.356 & 4.6\\
(AFHQ Cat) & 3 & \bf0.957 & 0.829 & \bf0.773 & 0.566 & \bf0.819 & 0.140 & 21.8\\
& 2 & 0.847 & 0.534 & 0.459 & 0.156 & 0.295 & 0.005 & 55.3\\
\hline
& 32 & 0.776 & 0.779 & 0.484 & 0.509 & 0.743 & 0.605 & 1.1\\
\cline{2-9}
ADA & 4 & 0.775 & 0.762 & \bf0.495 & 0.495 & \bf0.820 & 0.436 & 3.4\\
(AFHQ Dog) & 3 & 0.769 & 0.656 & 0.468 & 0.375 & \bf0.809 & 0.185 & 15.2\\
& 2 & 0.577 & 0.317 & 0.153 & 0.028 & 0.216 & 0.010 & 66.1\\
\hline
& 32 & 0.963 & 0.921 & 0.872 & 0.739 & 0.758 & 0.286 & 0.4\\
\cline{2-9}
ADA & 4 & \bf0.965 & 0.906 & \bf0.880 & 0.705 & \bf0.797 & 0.156 & 1.5\\
(AFHQ Wild) & 3 & \bf0.977 & 0.851 & 0.869 & 0.567 & 0.648 & 0.034 & 8.4\\
& 2 & 0.495 & 0.645 & 0.166 & 0.059 & 0.058 & 0.00 & 72.1\\
\hline
\end{tabular}
}
\caption{KID and precision (P) and recall (R) on AFHQ with $\approx$5k real and generated samples. Average over 10 runs.}
 \label{tab:results_afhq}
\end{table}

In terms of single-valued metrics, we evaluate our compressed models with FID~\cite{fid} and KID~\cite{kid} which are commonly used to evaluate GANs. In particular, we use KID to evaluate our AFHQ models, since it been shown to be less biased than FID at evaluating small datasets~\cite{sg2ada}. We evaluate FID for the models trained on the rest of the datasets.

Separate notions of quality and diversity allow us to determine if they are affected differently with compression. Hence, we test all models using our metrics, \textit{i.e.} LSH and LSH + KNN, as well as the original KNN-based precision and recall metric proposed by Kynkäänniemi \etal~\cite{impar}.

Results of all models compressed to 4 to 2 bits are shown in Tables~\ref{tab:results_ffhq_imagenet} and ~\ref{tab:results_afhq}. Most 4 to 3-bit compressed models are able to produce samples with similar quality as the 32-bit baseline, showing higher precision scores across the board. On the other hand, recall rarely improves, indicating a loss of sample diversity. Similarly, FID and KID always deteriorate with compression and do not reflect the quality retainment in the generated samples from compressed models.

\begin{figure*}
\centering
\begin{subfigure}[b]{0.162\textwidth}
    \includegraphics[width=\textwidth]{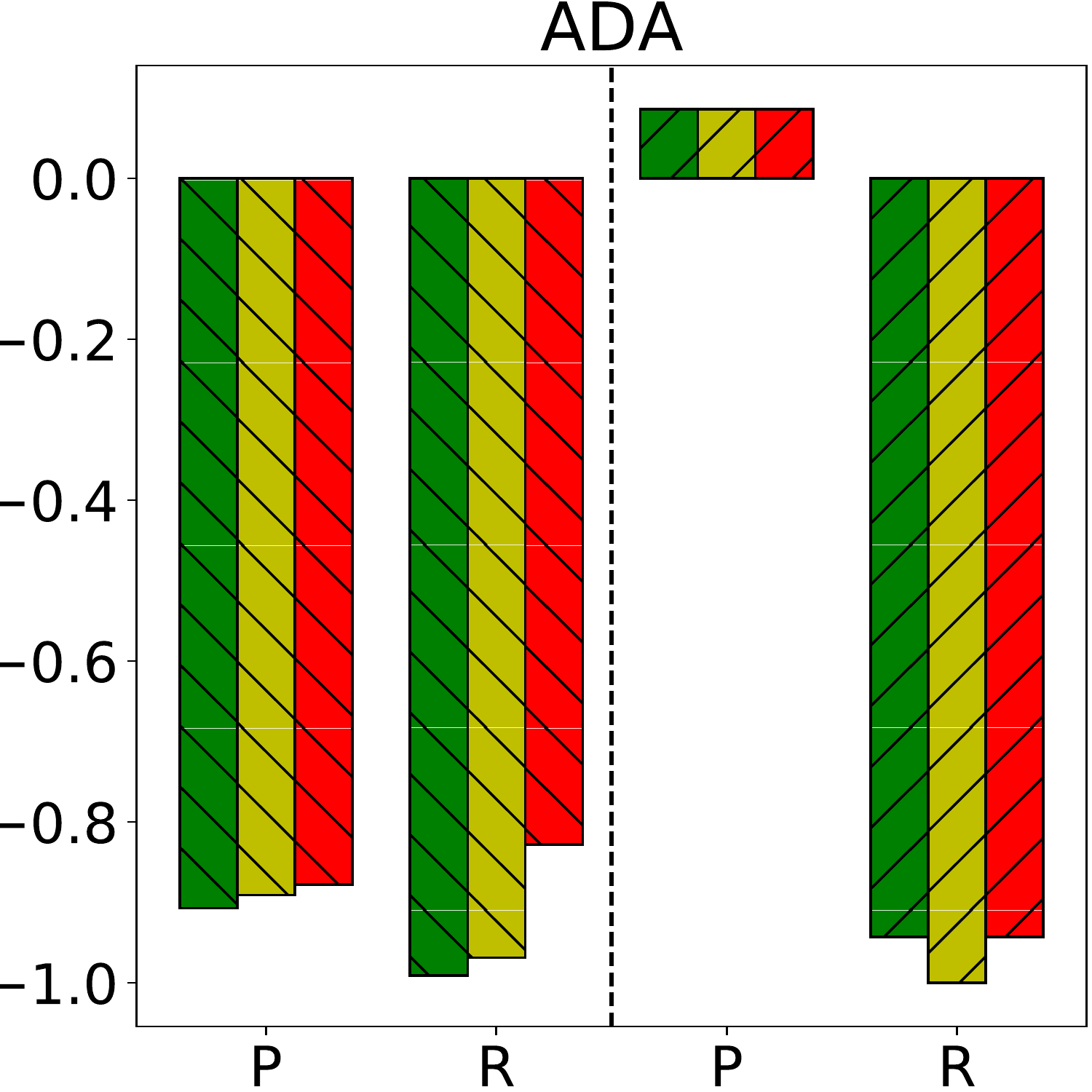}
\end{subfigure}
\hfill
\begin{subfigure}[b]{0.162\textwidth}
    \includegraphics[width=\textwidth]{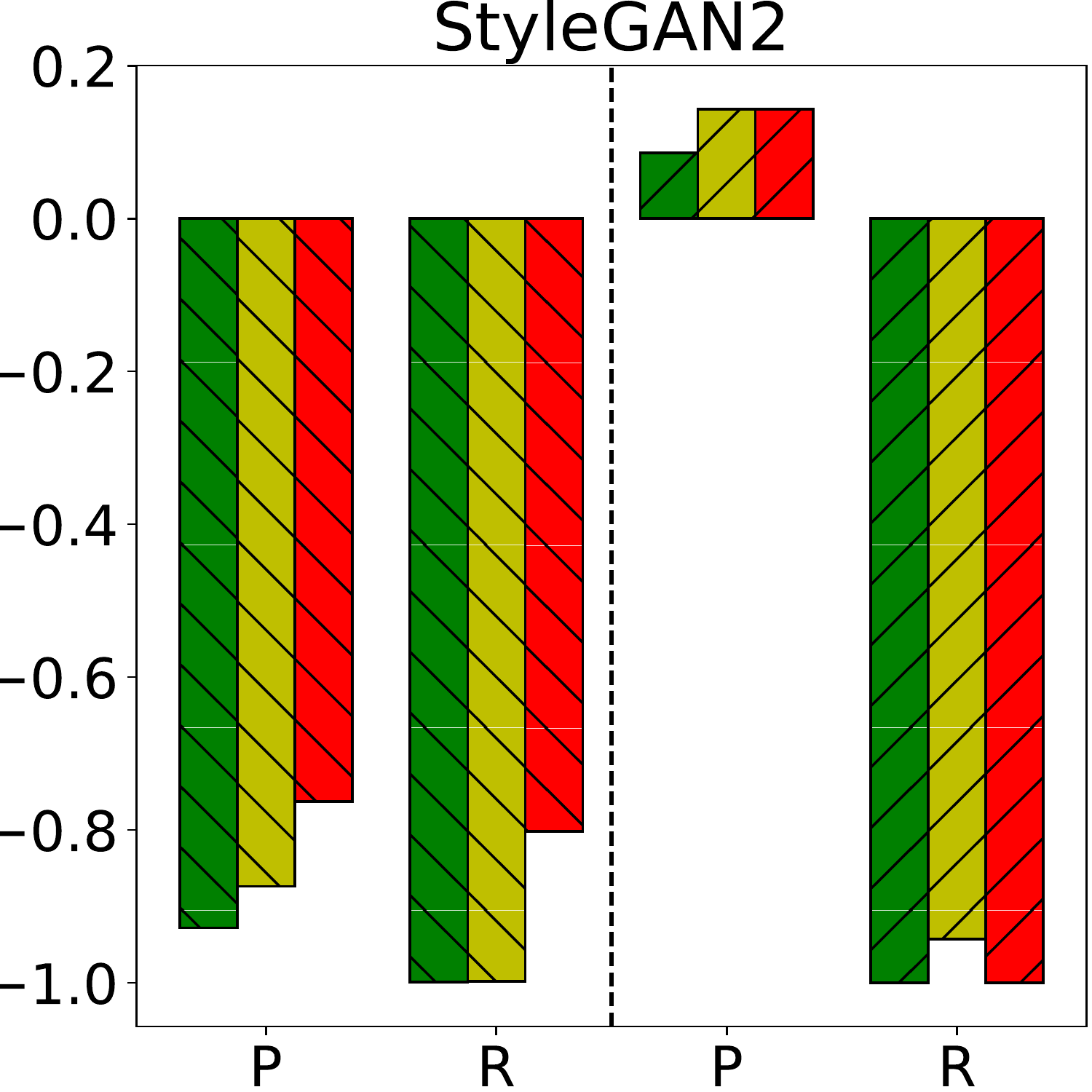}
\end{subfigure}
\hfill
\begin{subfigure}[b]{0.162\textwidth}
    \includegraphics[width=\textwidth]{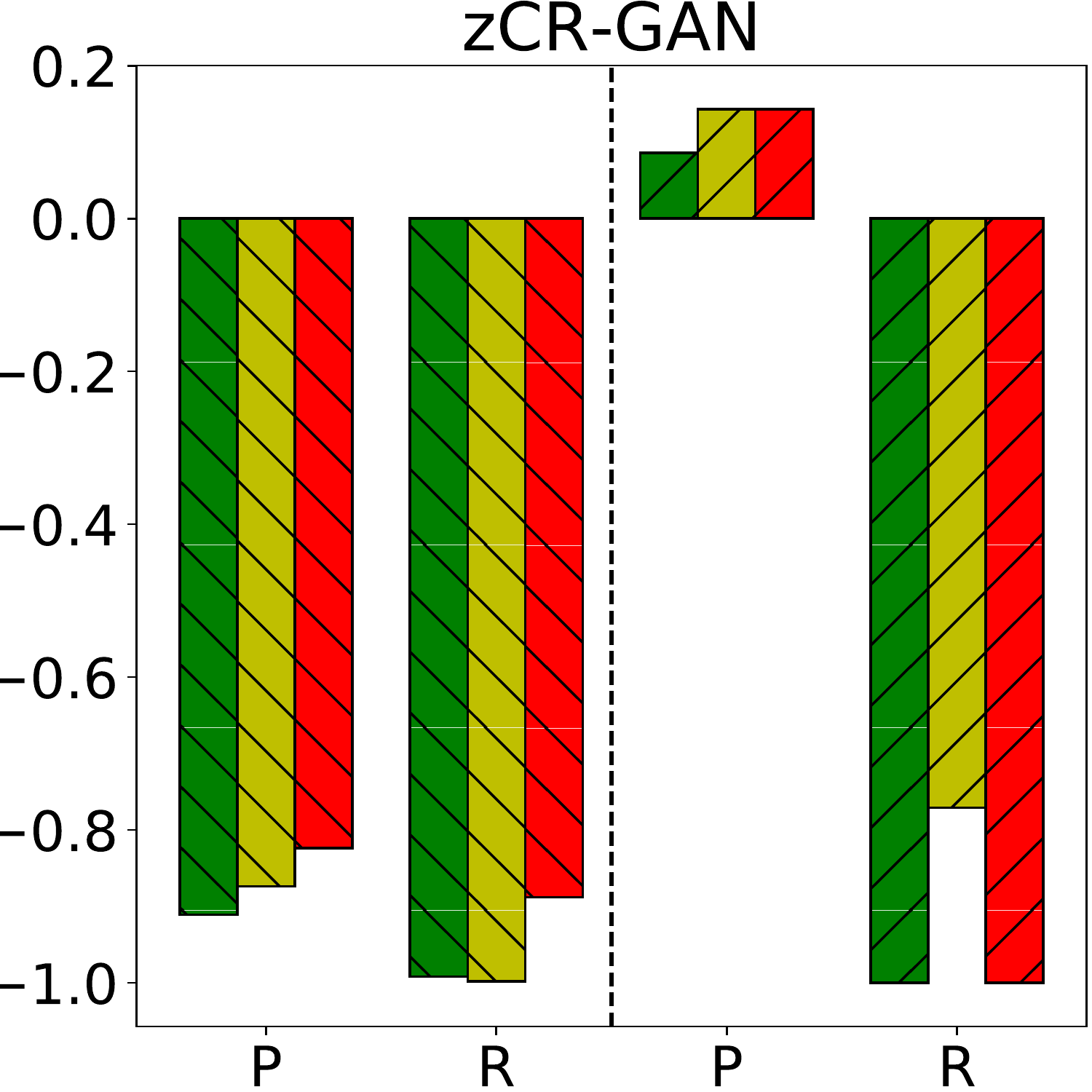}
\end{subfigure}
\hfill
\begin{subfigure}[b]{0.162\textwidth}
    \includegraphics[width=\textwidth]{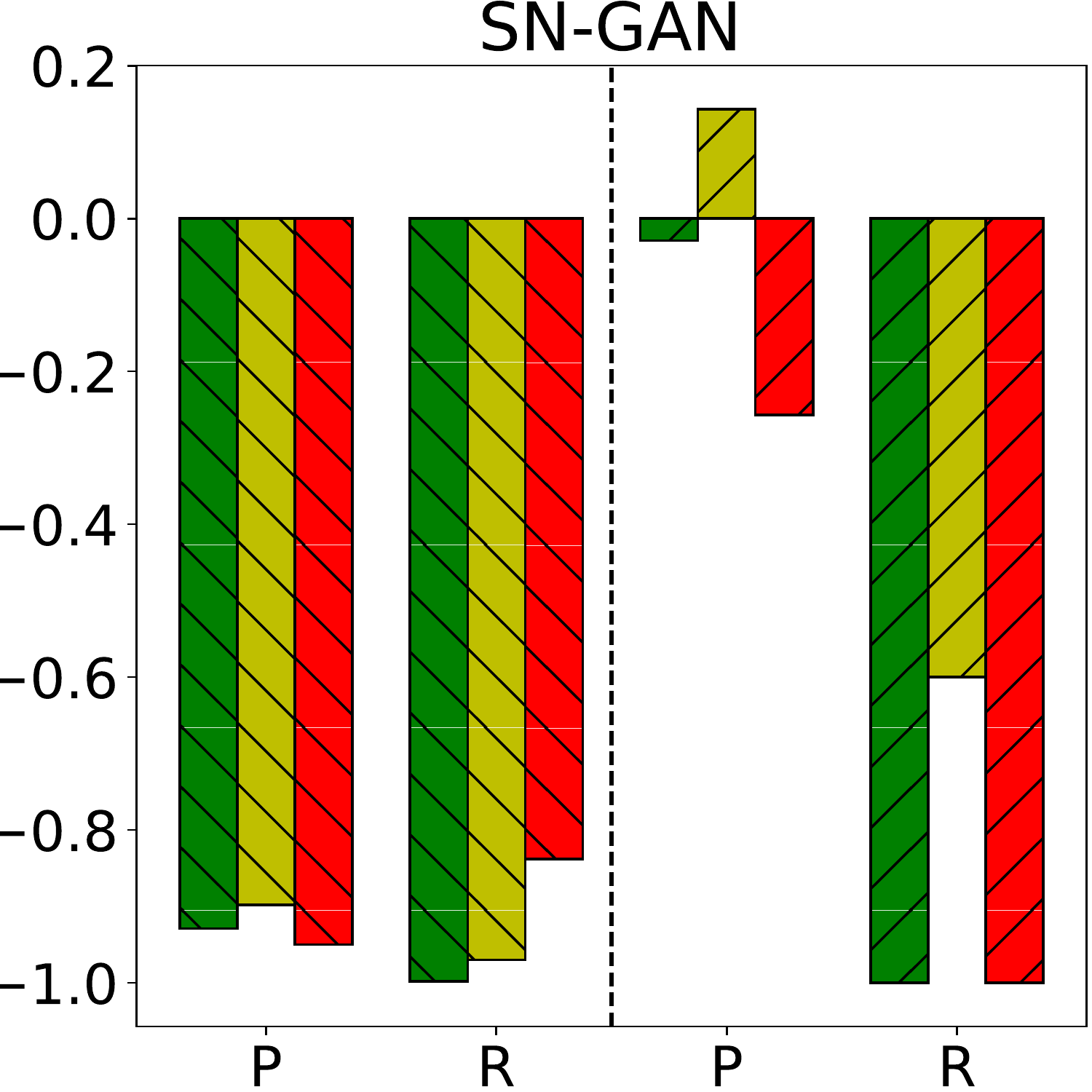}
\end{subfigure}
\hfill
\begin{subfigure}[b]{0.162\textwidth}
    \includegraphics[width=\textwidth]{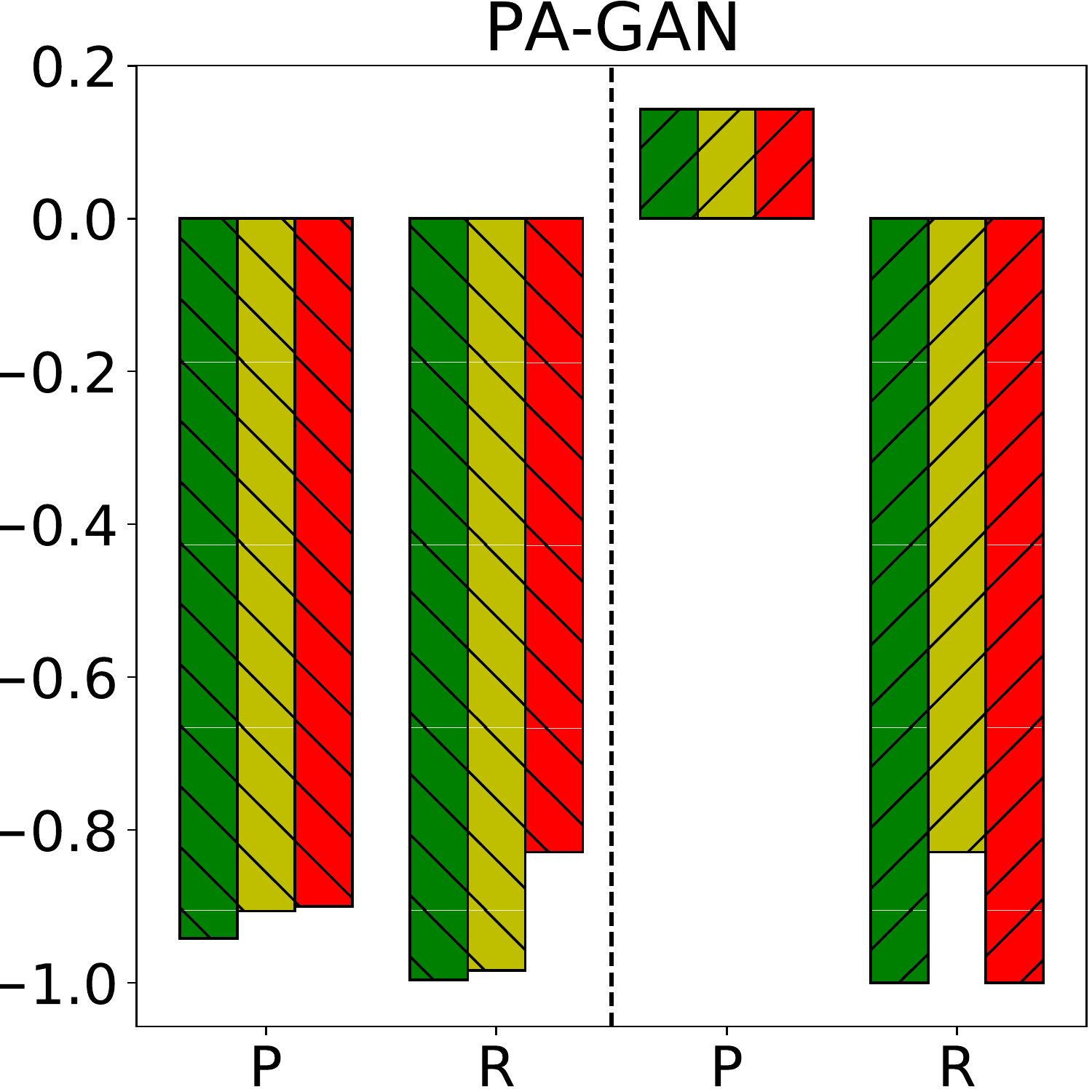}
\end{subfigure}
\hfill
\begin{subfigure}[b]{0.162\textwidth}
    \includegraphics[width=\textwidth]{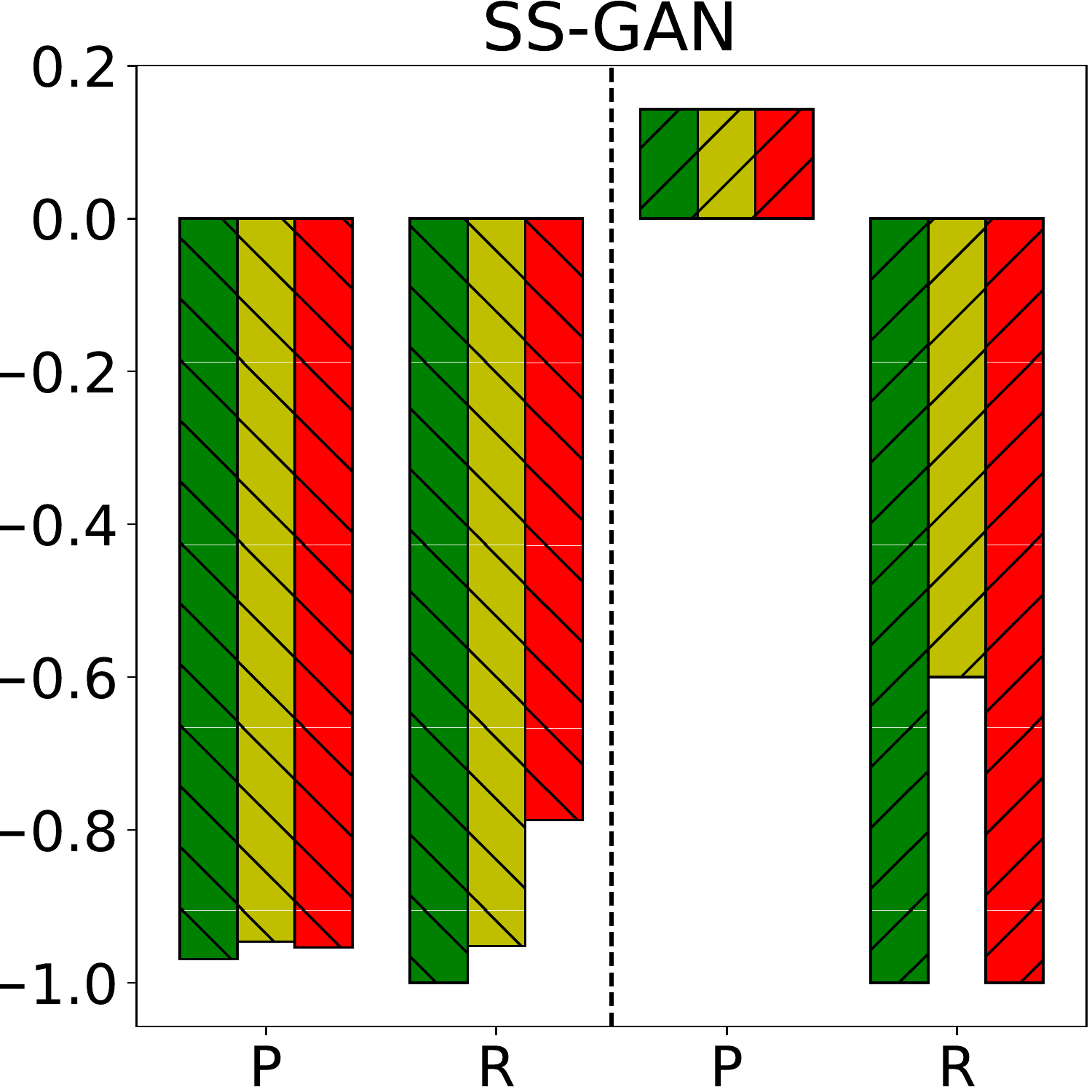}
\end{subfigure}
\hfill
\begin{subfigure}[b]{1.0\textwidth}
    \centering
    \includegraphics[width=0.47\textwidth]{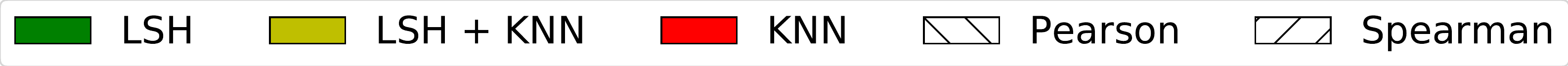}
\end{subfigure}
\vspace{-15px}
\caption{Pearson and Spearman correlations between FID and the different precision and recall methods on FFHQ 256x256.}
\label{fig:correlation_ffhq}
\end{figure*}

\begin{figure}
\centering
\begin{subfigure}[b]{0.47\textwidth}
    \includegraphics[width=\textwidth]{images/correlations_2_to_6_bits/legend.pdf}
\end{subfigure}
\hfill
\begin{subfigure}[b]{0.155\textwidth}
    \includegraphics[width=\textwidth]{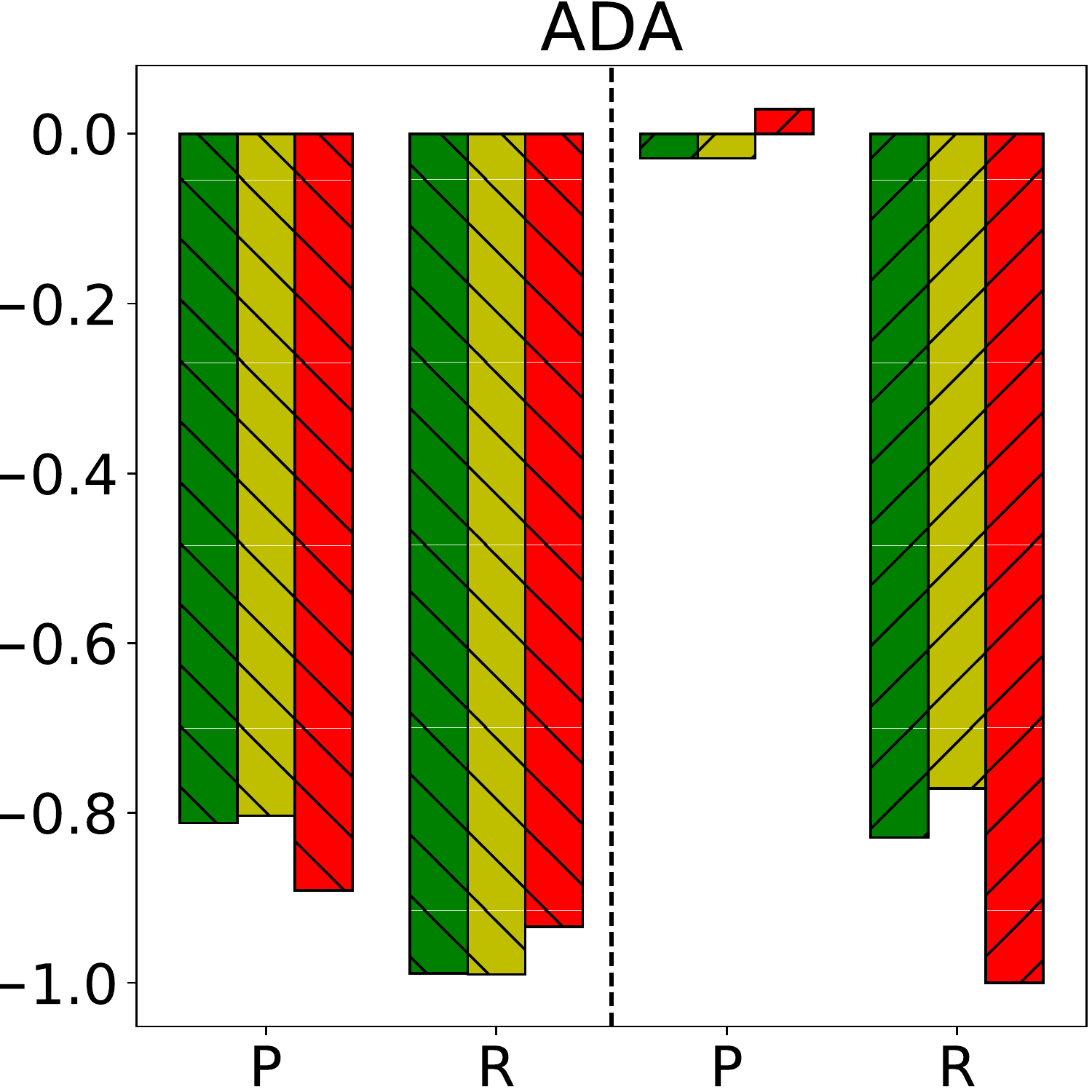}
    \caption{AFHQ Cat.}
\end{subfigure}
\begin{subfigure}[b]{0.155\textwidth}
    \includegraphics[width=\textwidth]{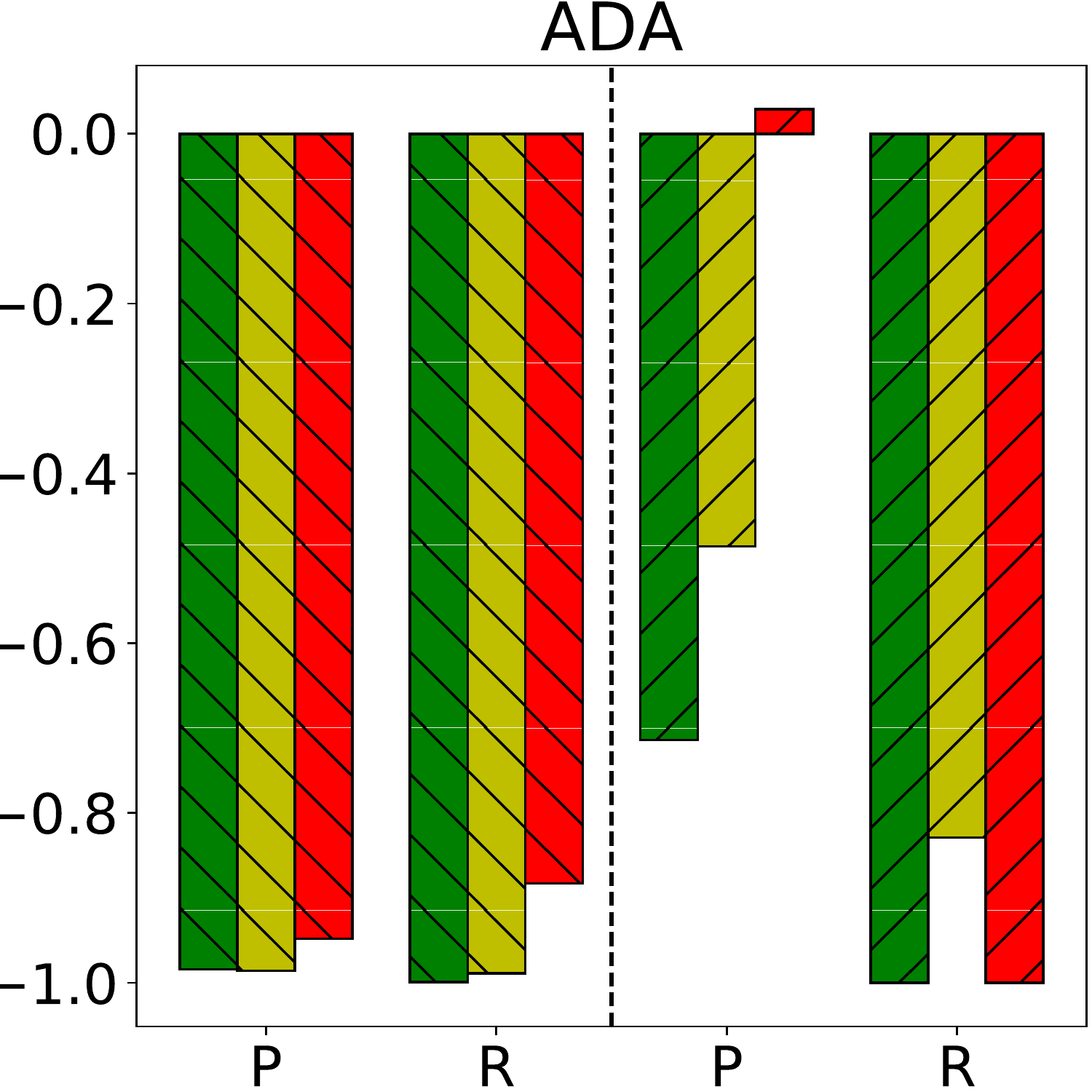}
    \caption{AFHQ Dog.}
\end{subfigure}
\hfill
\begin{subfigure}[b]{0.155\textwidth}
    \includegraphics[width=\textwidth]{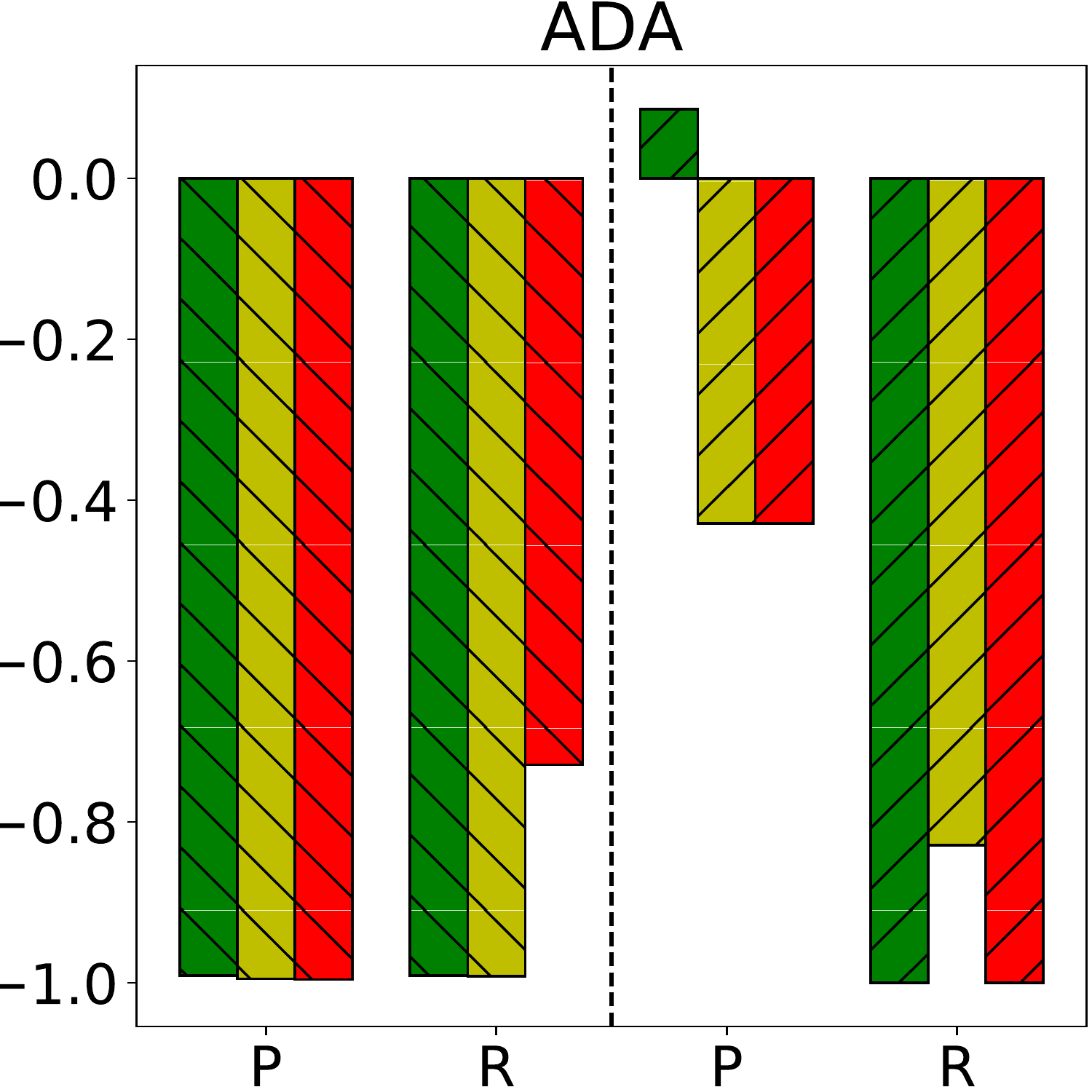}
    \caption{AFHQ Wild.}
\end{subfigure}
\vspace{-15px}
\caption{Pearson and Spearman correlations between KID and the different precision and recall methods on AFHQ.}
\label{fig:correlation_afhq}
\end{figure}

\subsection{Correlations with precision and recall}

The degradation of the FID and KID scores with increased compression levels incentivized us to study their individual correlations with precision and recall. To increase the number of values when calculating the Pearson and Spearman correlations, we use the scores of compression from 6 to 2 bits for each metric. Whereas Pearson directly uses the score values to evaluate their correlation in terms of linear relationships, Spearman uses the ranks of the scores, assessing their correlation in terms of monotonic relationships. Correlation values closer to $-1$ are ideal both for Pearson and Spearman since an inverse trend between FID/KID scores and precision/recall scores is expected, \textit{i.e.} as one increases, the other should decrease, and vice-versa.

Figures~\ref{fig:correlation_ffhq} and~\ref{fig:correlation_afhq} show the correlation results for FID and KID, respectively. We observe that the absolute Pearson correlations are high both for precision and recall. However, absolute Spearman correlations are high for recall but show close to no correlation for precision, in most cases. This suggests that FID and KID scores reflect more sample diversity than sample quality, being biased against high quality but low diversity models, \textit{i.e.} mode collapsed.


\subsection{Pareto frontiers}

If we look at quality or diversity dominant GANs, do they become more balanced after compression? To try to answer this question we took a look at how compressed StyleGAN2 compares against other StyleGAN2 configuration variants originally proposed by Karras \etal~\cite{stylegan2}. Their discussed variants consist in modifications to the original StyleGAN~\cite{stylegan} by applying weight demodulation (config-b), lazy regularization (config-c), path length regularization (config-d), new generator and discriminator architectures (config-e), and larger networks, \textit{i.e.} StyleGAN2. Compared to the other variants, StyleGAN2 improves recall at the cost of sacrificing precision when evaluating with KNN.

\begin{figure}
    \centering
    \begin{subfigure}[b]{0.35\textwidth}
        \includegraphics[width=\textwidth]{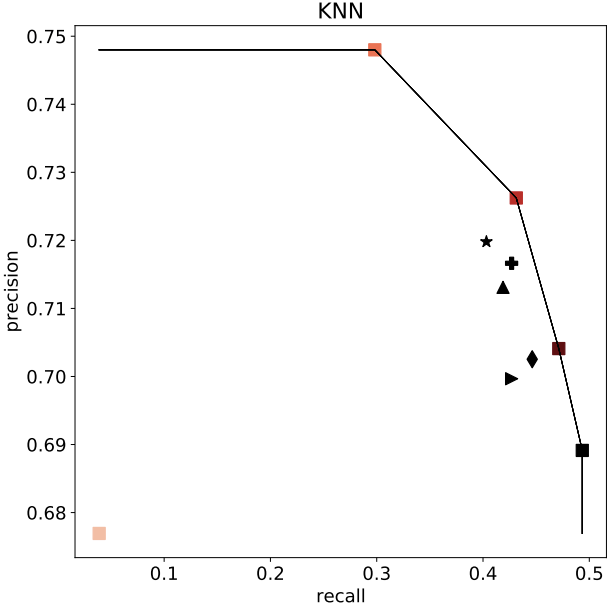}
    \end{subfigure}
    \begin{subfigure}[b]{0.478\textwidth}
        \includegraphics[width=\textwidth]{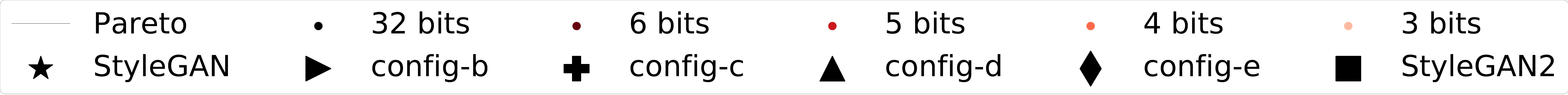}
    \end{subfigure}
    \caption{Pareto frontier of different StyleGAN configurations on FFHQ 1024x1024. Average over 3 evaluation runs.}
    \label{fig:pareto_frontiers_ffhq1024}
\end{figure}

In Figure~\ref{fig:pareto_frontiers_ffhq1024}, we observe that compression in StyleGAN2 improves the precision and recall trade-off relative to the rest of the variants, improving the previous Pareto frontier and achieving higher precision than the other configurations at similar recall levels. This suggests that compression may be used to balance out quality or diversity-dominant GANs.

\section{Conclusion}

With this work, we showed post-training compression in GANs is a viable way of balancing the trade-off between sample quality and sample diversity, on top of reducing the generator's size. In particular, clipping and linear quantization techniques retain the quality of the generated set at the cost of some diversity, which has been an active research topic in current GANs research. Hence, depending on the use-case, compression may act as a simple, yet effective measure to balance existing GANs post-training, \textit{i.e.} without requiring much time and computation.
In the future, we plan to further optimize GANs inference by also compressing activations using the studied compression techniques.


{\small
\bibliographystyle{ieee_fullname}
\bibliography{egbib}

\begin{thebibliography}{10}\itemsep=-1pt

\bibitem{aguinaldo2019compressing}
Angeline Aguinaldo, Ping-Yeh Chiang, Alex Gain, Ameya Patil, Kolten Pearson,
  and Soheil Feizi.
\newblock Compressing gans using knowledge distillation.
\newblock {\em arXiv preprint arXiv:1902.00159}, 2019.

\bibitem{aciq}
Ron Banner, Yury Nahshan, and Daniel Soudry.
\newblock Post training 4-bit quantization of convolutional networks for
  rapid-deployment.
\newblock In H. Wallach, H. Larochelle, A. Beygelzimer, F. d\textquotesingle
  Alch\'{e}-Buc, E. Fox, and R. Garnett, editors, {\em Advances in Neural
  Information Processing Systems}, volume~32, pages 7950--7958. Curran
  Associates, Inc., 2019.

\bibitem{kdtrees}
Jon~Louis Bentley.
\newblock Multidimensional binary search trees used for associative searching.
\newblock {\em Communications of the ACM}, 18(9):509--517, 1975.

\bibitem{kid}
Mikołaj Bińkowski, Dougal~J. Sutherland, Michael Arbel, and Arthur Gretton.
\newblock Demystifying {MMD} {GAN}s.
\newblock In {\em International Conference on Learning Representations}, 2018.

\bibitem{biggan}
Andrew Brock, Jeff Donahue, and Karen Simonyan.
\newblock Large scale {GAN} training for high fidelity natural image synthesis.
\newblock In {\em International Conference on Learning Representations}, 2019.

\bibitem{chen2020distilling}
Hanting Chen, Yunhe Wang, Han Shu, Changyuan Wen, Chunjing Xu, Boxin Shi, Chao
  Xu, and Chang Xu.
\newblock Distilling portable generative adversarial networks for image
  translation.
\newblock In {\em Proceedings of the AAAI Conference on Artificial
  Intelligence}, volume~34, pages 3585--3592, 2020.

\bibitem{ssgan}
Ting Chen, Xiaohua Zhai, Marvin Ritter, Mario Lucic, and Neil Houlsby.
\newblock Self-supervised gans via auxiliary rotation loss.
\newblock In {\em Proceedings of the IEEE/CVF Conference on Computer Vision and
  Pattern Recognition}, pages 12154--12163, 2019.

\bibitem{choi2020starganv2}
Yunjey Choi, Youngjung Uh, Jaejun Yoo, and Jung-Woo Ha.
\newblock Stargan v2: Diverse image synthesis for multiple domains.
\newblock In {\em Proceedings of the IEEE Conference on Computer Vision and
  Pattern Recognition}, 2020.

\bibitem{imagenet}
Jia Deng, Wei Dong, Richard Socher, Li-Jia Li, Kai Li, and Li Fei-Fei.
\newblock Imagenet: A large-scale hierarchical image database.
\newblock In {\em 2009 IEEE conference on computer vision and pattern
  recognition}, pages 248--255. Ieee, 2009.

\bibitem{nas}
Thomas Elsken, Jan~Hendrik Metzen, Frank Hutter, et~al.
\newblock Neural architecture search: A survey.
\newblock {\em J. Mach. Learn. Res.}, 20(55):1--21, 2019.

\bibitem{gans}
Ian Goodfellow, Jean Pouget-Abadie, Mehdi Mirza, Bing Xu, David Warde-Farley,
  Sherjil Ozair, Aaron Courville, and Yoshua Bengio.
\newblock Generative adversarial nets.
\newblock In Z. Ghahramani, M. Welling, C. Cortes, N. Lawrence, and K.~Q.
  Weinberger, editors, {\em Advances in Neural Information Processing Systems},
  volume~27. Curran Associates, Inc., 2014.

\bibitem{fid}
Martin Heusel, Hubert Ramsauer, Thomas Unterthiner, Bernhard Nessler, and Sepp
  Hochreiter.
\newblock Gans trained by a two time-scale update rule converge to a local nash
  equilibrium.
\newblock In I. Guyon, U.~V. Luxburg, S. Bengio, H. Wallach, R. Fergus, S.
  Vishwanathan, and R. Garnett, editors, {\em Advances in Neural Information
  Processing Systems}, volume~30, pages 6626--6637. Curran Associates, Inc.,
  2017.

\bibitem{hinton2015distilling}
Geoffrey Hinton, Oriol Vinyals, and Jeff Dean.
\newblock Distilling the knowledge in a neural network.
\newblock {\em arXiv preprint arXiv:1503.02531}, 2015.

\bibitem{sg2ada}
Tero Karras, Miika Aittala, Janne Hellsten, Samuli Laine, Jaakko Lehtinen, and
  Timo Aila.
\newblock Training generative adversarial networks with limited data.
\newblock In H. Larochelle, M. Ranzato, R. Hadsell, M.~F. Balcan, and H. Lin,
  editors, {\em Advances in Neural Information Processing Systems}, volume~33,
  pages 12104--12114. Curran Associates, Inc., 2020.

\bibitem{stylegan}
T. {Karras}, S. {Laine}, and T. {Aila}.
\newblock A style-based generator architecture for generative adversarial
  networks.
\newblock In {\em 2019 IEEE/CVF Conference on Computer Vision and Pattern
  Recognition (CVPR)}, pages 4396--4405, 2019.

\bibitem{stylegan2}
Tero Karras, Samuli Laine, Miika Aittala, Janne Hellsten, Jaakko Lehtinen, and
  Timo Aila.
\newblock Analyzing and improving the image quality of {StyleGAN}.
\newblock In {\em Proc. CVPR}, 2020.

\bibitem{lit}
Animesh Koratana, Daniel Kang, Peter Bailis, and Matei Zaharia.
\newblock {LIT}: Learned intermediate representation training for model
  compression.
\newblock In Kamalika Chaudhuri and Ruslan Salakhutdinov, editors, {\em
  Proceedings of the 36th International Conference on Machine Learning},
  volume~97 of {\em Proceedings of Machine Learning Research}, pages
  3509--3518. PMLR, 09--15 Jun 2019.

\bibitem{impar}
Tuomas Kynk\"{a}\"{a}nniemi, Tero Karras, Samuli Laine, Jaakko Lehtinen, and
  Timo Aila.
\newblock Improved precision and recall metric for assessing generative models.
\newblock In H. Wallach, H. Larochelle, A. Beygelzimer, F. d\textquotesingle
  Alch\'{e}-Buc, E. Fox, and R. Garnett, editors, {\em Advances in Neural
  Information Processing Systems}, volume~32, pages 3927--3936. Curran
  Associates, Inc., 2019.

\bibitem{li2020gan}
Muyang Li, Ji Lin, Yaoyao Ding, Zhijian Liu, Jun-Yan Zhu, and Song Han.
\newblock Gan compression: Efficient architectures for interactive conditional
  gans.
\newblock In {\em Proceedings of the IEEE/CVF Conference on Computer Vision and
  Pattern Recognition}, 2020.

\bibitem{liu2020binarizing}
Jinglan Liu, Jiaxin Zhang, Yukun Ding, Xiaowei Xu, Meng Jiang, and Yiyu Shi.
\newblock Binarizing weights wisely for edge intelligence: Guide for partial
  binarization of deconvolution-based generators.
\newblock {\em IEEE Transactions on Computer-Aided Design of Integrated
  Circuits and Systems}, 2020.

\bibitem{sngan}
Takeru Miyato, Toshiki Kataoka, Masanori Koyama, and Yuichi Yoshida.
\newblock Spectral normalization for generative adversarial networks.
\newblock In {\em International Conference on Learning Representations}, 2018.

\bibitem{mark-evaluate}
Gon{\c{c}}alo Mordido and Christoph Meinel.
\newblock Mark-evaluate: Assessing language generation using population
  estimation methods.
\newblock In {\em Proceedings of the 28th International Conference on
  Computational Linguistics}, pages 1963--1977, Barcelona, Spain (Online), Dec.
  2020. International Committee on Computational Linguistics.

\bibitem{fti}
Goncalo Mordido, Julian Niedermeier, and Christoph Meinel.
\newblock Assessing image and text generation with topological analysis and
  fuzzy logic.
\newblock In {\em Proceedings of the IEEE/CVF Winter Conference on Applications
  of Computer Vision (WACV)}, pages 2013--2022, January 2021.

\bibitem{mcq}
Gon{\c{c}}alo Mordido, Matthijs Van~Keirsbilck, and Alexander Keller.
\newblock Instant quantization of neural networks using monte carlo methods.
\newblock {\em arXiv preprint arXiv:1905.12253}, 2019.

\bibitem{mcgq}
Goncalo Mordido, Matthijs Van~Keirsbilck, and Alexander Keller.
\newblock Monte carlo gradient quantization.
\newblock In {\em Proceedings of the IEEE/CVF Conference on Computer Vision and
  Pattern Recognition Workshops}, pages 718--719, 2020.

\bibitem{mordido2018dropout}
Gon{\c{c}}alo Mordido, Haojin Yang, and Christoph Meinel.
\newblock Dropout-gan: Learning from a dynamic ensemble of discriminators.
\newblock {\em arXiv preprint arXiv:1807.11346}, 2018.

\bibitem{mordido2020microbatchgan}
Gon{\c{c}}alo Mordido, Haojin Yang, and Christoph Meinel.
\newblock microbatchgan: Stimulating diversity with multi-adversarial
  discrimination.
\newblock In {\em The IEEE Winter Conference on Applications of Computer
  Vision}, pages 3061--3070, 2020.

\bibitem{prd}
Mehdi S.~M. Sajjadi, Olivier Bachem, Mario Lucic, Olivier Bousquet, and Sylvain
  Gelly.
\newblock Assessing generative models via precision and recall.
\newblock In S. Bengio, H. Wallach, H. Larochelle, K. Grauman, N. Cesa-Bianchi,
  and R. Garnett, editors, {\em Advances in Neural Information Processing
  Systems}, volume~31, pages 5228--5237. Curran Associates, Inc., 2018.

\bibitem{is}
Tim Salimans, Ian Goodfellow, Wojciech Zaremba, Vicki Cheung, Alec Radford, Xi
  Chen, and Xi Chen.
\newblock Improved techniques for training gans.
\newblock In D. Lee, M. Sugiyama, U. Luxburg, I. Guyon, and R. Garnett,
  editors, {\em Advances in Neural Information Processing Systems}, volume~29,
  pages 2234--2242. Curran Associates, Inc., 2016.

\bibitem{sauder-etal-2020-best}
Jonathan Sauder, Ting Hu, Xiaoyin Che, Goncalo Mordido, Haojin Yang, and
  Christoph Meinel.
\newblock Best student forcing: A simple training mechanism in adversarial
  language generation.
\newblock In {\em Proceedings of the 12th Language Resources and Evaluation
  Conference}, pages 4680--4688, Marseille, France, May 2020. European Language
  Resources Association.

\bibitem{shu2019co}
Han Shu, Yunhe Wang, Xu Jia, Kai Han, Hanting Chen, Chunjing Xu, Qi Tian, and
  Chang Xu.
\newblock Co-evolutionary compression for unpaired image translation.
\newblock In {\em Proceedings of the IEEE/CVF International Conference on
  Computer Vision}, pages 3235--3244, 2019.

\bibitem{vgg}
Karen Simonyan and Andrew Zisserman.
\newblock Very deep convolutional networks for large-scale image recognition.
\newblock {\em arXiv preprint arXiv:1409.1556}, 2014.

\bibitem{slaney2008locality}
Malcolm Slaney and Michael Casey.
\newblock Locality-sensitive hashing for finding nearest neighbors [lecture
  notes].
\newblock {\em IEEE Signal processing magazine}, 25(2):128--131, 2008.

\bibitem{wang2020ganslimming}
Haotao Wang, Shupeng Gui, Haichuan Yang, Ji Liu, and Zhangyang Wang.
\newblock Gan slimming: All-in-one gan compression by a unified optimization
  framework.
\newblock In {\em European Conference on Computer Vision}, 2020.

\bibitem{qgan}
Peiqi Wang, Dongsheng Wang, Yu Ji, Xinfeng Xie, Haoxuan Song, XuXin Liu,
  Yongqiang Lyu, and Yuan Xie.
\newblock Qgan: Quantized generative adversarial networks.
\newblock {\em arXiv preprint arXiv:1901.08263}, 2019.

\bibitem{yu2020self}
Chong Yu and Jeff Pool.
\newblock Self-supervised gan compression.
\newblock {\em arXiv preprint arXiv:2007.01491}, 2020.

\bibitem{pagan}
Dan Zhang and Anna Khoreva.
\newblock Progressive augmentation of gans.
\newblock In H. Wallach, H. Larochelle, A. Beygelzimer, F. d\textquotesingle
  Alch\'{e}-Buc, E. Fox, and R. Garnett, editors, {\em Advances in Neural
  Information Processing Systems}, volume~32. Curran Associates, Inc., 2019.

\bibitem{ocs}
Ritchie Zhao, Yuwei Hu, Jordan Dotzel, Chris De~Sa, and Zhiru Zhang.
\newblock Improving neural network quantization without retraining using
  outlier channel splitting.
\newblock In {\em International conference on machine learning}, pages
  7543--7552. PMLR, 2019.

\bibitem{zcr}
Zhengli Zhao, Sameer Singh, Honglak Lee, Zizhao Zhang, Augustus Odena, and Han
  Zhang.
\newblock Improved consistency regularization for gans.
\newblock {\em arXiv preprint arXiv:2002.04724}, 2020.

\end{thebibliography}
}

\clearpage
\appendix

\section{Algorithm}

Pseudo-code for the proposed metrics (LSH and LSH+KNN) is presented in Algorithm~\ref{alg:methods}.

\begin{algorithm*}
\SetAlgoLined
\SetKwInOut{Input}{Input}
\Input{Set of real and generated images $(X_r, X_g)$, feature network $F$ retrieving $d$-dimensional embeddings.}

  \DontPrintSemicolon
  \SetKwFunction{FPR}{Precision-Recall}
  \SetKwFunction{FH}{hyperplanes}
  \SetKwFunction{FHT}{Hash-Table}
  \SetKwFunction{FME}{LSH}
  \SetKwFunction{FMEKNN}{LSH+KNN}
  \SetKwProg{Fn}{function}{:}{}
  \Fn{\FPR{$X_r$, $X_g$, $F$, $k$}}{
        $\Phi_r \leftarrow F(X_r)$\;
        $\Phi_g \leftarrow F(X_g)$\;
        \tcp{Generate random hyperplanes}
        $H \leftarrow$ $\floor{\log(\vert \Phi_r \vert + \vert \Phi_g \vert)}$\;
        $h \leftarrow$ random.normal($H$, $D$)\;
        $b \leftarrow$ random.uniform($H$)\;
        \tcp{Initialize hash-tables}
        $HT_r \leftarrow$ \FHT{$\Phi_r$, $h$, $b$}\;
        $HT_g \leftarrow$ \FHT{$\Phi_g$, $h$, $b$}\;
        \tcp{Evaluate with LSH}
        precision$_\text{LSH} \leftarrow$ \FME{$HT_r$, $HT_g$, $k$}\;
        recall$_\text{LSH} \leftarrow$ \FME{$HT_g$, $HT_r$, $k$}\;
        \tcp{Evaluate with LSH+KNN}
        precision$_\text{LSH+KNN} \leftarrow$ \FMEKNN{$HT_r$, $HT_g$, $k$}\;
        recall$_\text{LSH+KNN} \leftarrow$ \FMEKNN{$HT_g$, $HT_r$, $k$}\;
        \KwRet $\text{precision}_\text{LSH},\text{recall}_\text{LSH},\text{precision}_\text{LSH+KNN},\text{recall}_\text{LSH+KNN}$\;
    }
    \;
    \Fn{\FHT{$\theta$, $h$, $b$}}{
        \For{$\phi \in \theta$}{
            key $\leftarrow$ (dot($\phi,h) + b \geq$ 0).int()\;
            \uIf{key $\notin$ HT$_\theta$}{
                HT$_\theta$[key] = [$\phi$]\;
            }
            \Else{
                HT$_\theta$[key].append($\phi$)\;
            }
        }
        \KwRet HT$_\theta$\;
    }
    \;
    \Fn{\FME{$HT_a$, $HT_b$}}{
       $n \leftarrow 0$\;
       \For{$\phi \in HT_a.values$}{
            \If{$HT_a.key(\phi) \in HT_b$}{
                $n \leftarrow n+1$\;
            }
       }
       \KwRet $n / \vert HT_a.values\vert$\;
    }
    \;
    \Fn{\FMEKNN{$HT_a$, $HT_b$, $k$}}{
        \For{$\phi \in HT_a.values$}{
            $d \leftarrow \{ \lVert \phi - \phi^\prime \rVert_2 \}$ \textit{\textbf{ for} all} $\phi^\prime \in HT_a[HT_a.key(\phi)]$\; 
            $r_\phi \leftarrow \text{min}_{k+1}(d)$\;
        }
        $n \leftarrow 0$\;
        \For{$\phi \in HT_b.values$}{
            \If{$\lVert \phi - \phi^\prime \rVert_2 \leq r_{\phi^\prime}$ \textbf{ for} any $\phi^\prime \in HT_a[HT_b.key(\phi)]$}{ 
                $n \leftarrow n + 1$\;
            }
        }
        \KwRet $n / \vert HT_b.values\vert$\;

    }
    
 \caption{Precision and recall metrics using locality-sensitive hashing.}
 \label{alg:methods}
\end{algorithm*}

\section{Realism scores}

\begin{figure}[h]
\centering
\begin{subfigure}[b]{0.47\textwidth}
    \centering
    \includegraphics[width=0.49\textwidth]{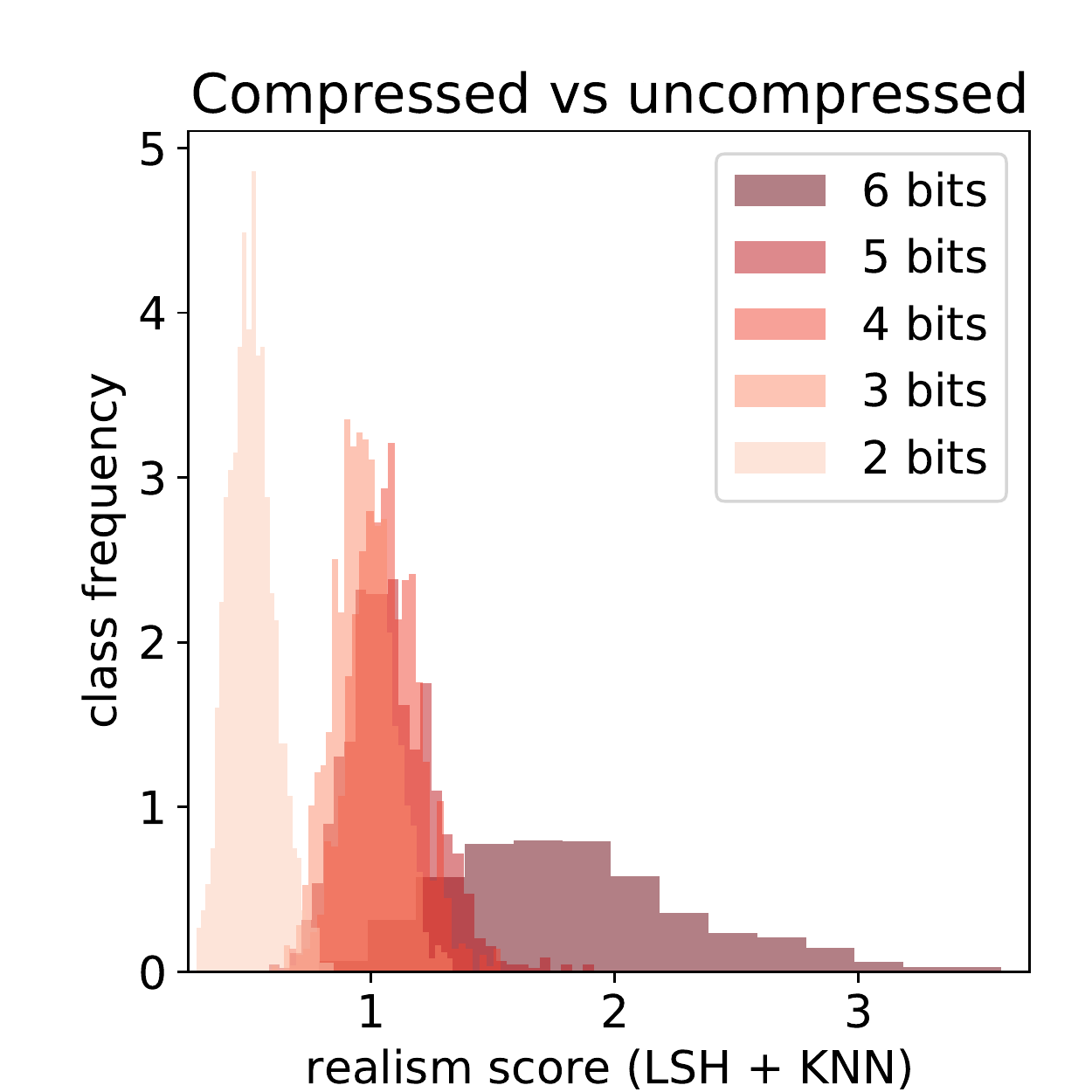}
    \hfill
    \includegraphics[width=0.49\textwidth]{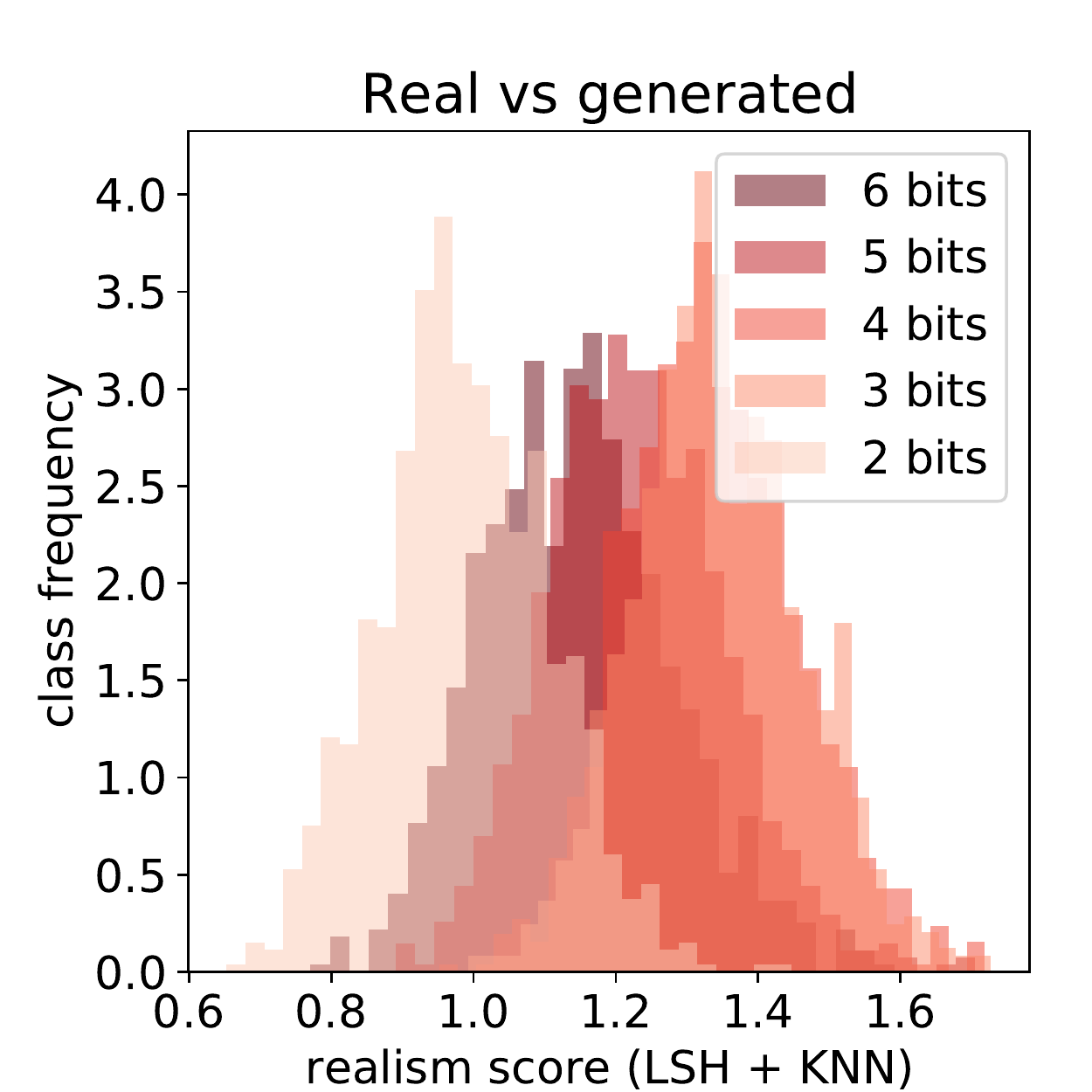}
\end{subfigure}
\caption{BigGAN's mean class realism scores on ImageNet 128x128.}
\label{fig:realism_score_imagenet}
\end{figure}

Figure~\ref{fig:realism_score_imagenet} shows the mean realism scores of each ImageNet class using our LSH+KNN metric. 
We evaluated the images generated by the compressed models against the images generated by the 32-bit BigGAN (left) as well as the images from the real set (right), as usual. When assessing the generated sets of the compressed and uncompressed models, we see a decline in the realism scores, as expected. When comparing the compressed models with the real data, LSH+KNN shows that realism scores increase until the 2-bit compression mark, and decline at 2 bits. This goes in accordance with our findings in the paper. 

\section{Mean and generated images}

\begin{figure*}
    \begin{picture}(100,300)      
        \put(0,0){\includegraphics[width=0.98\textwidth]{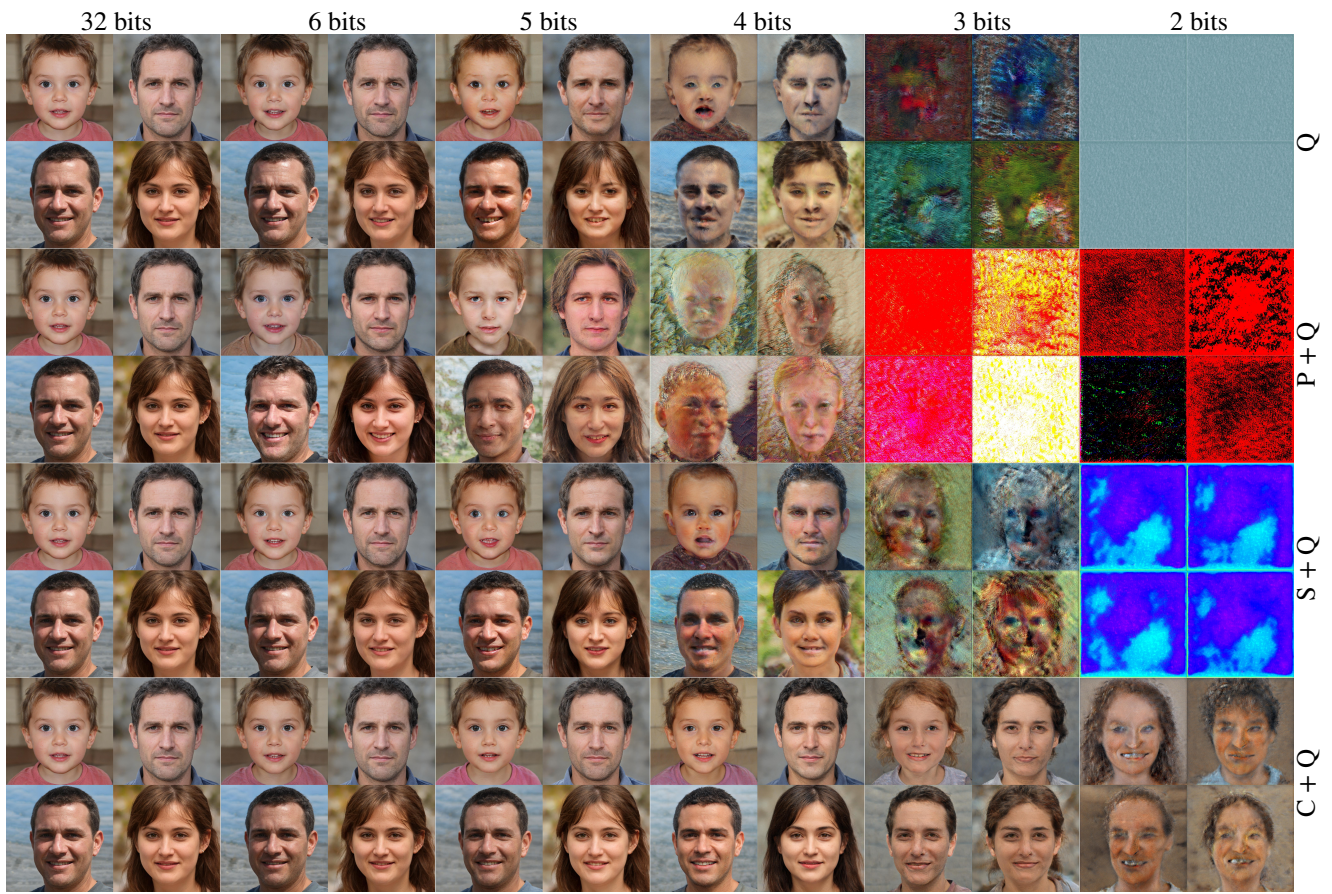}}
        \put(28,326){32 bits}
        \put(114,326){6 bits}
        \put(194,326){5 bits}
        \put(275,326){4 bits}
        \put(358,326){3 bits}
        \put(440,326){2 bits}
        \put(488,280){\rotatebox{90}{Q}}
        \put(488,191){\rotatebox{90}{P + Q}}
        \put(488,110){\rotatebox{90}{S + Q}}
        \put(488,28){\rotatebox{90}{C + Q}}
    \end{picture}     
    \caption{StyleGAN2 on FFHQ 1024x1024.}
    \label{fig:appendix_ffhq1024_stylegan2}
\end{figure*}

\begin{figure*}
    \includegraphics[width=0.97\textwidth]{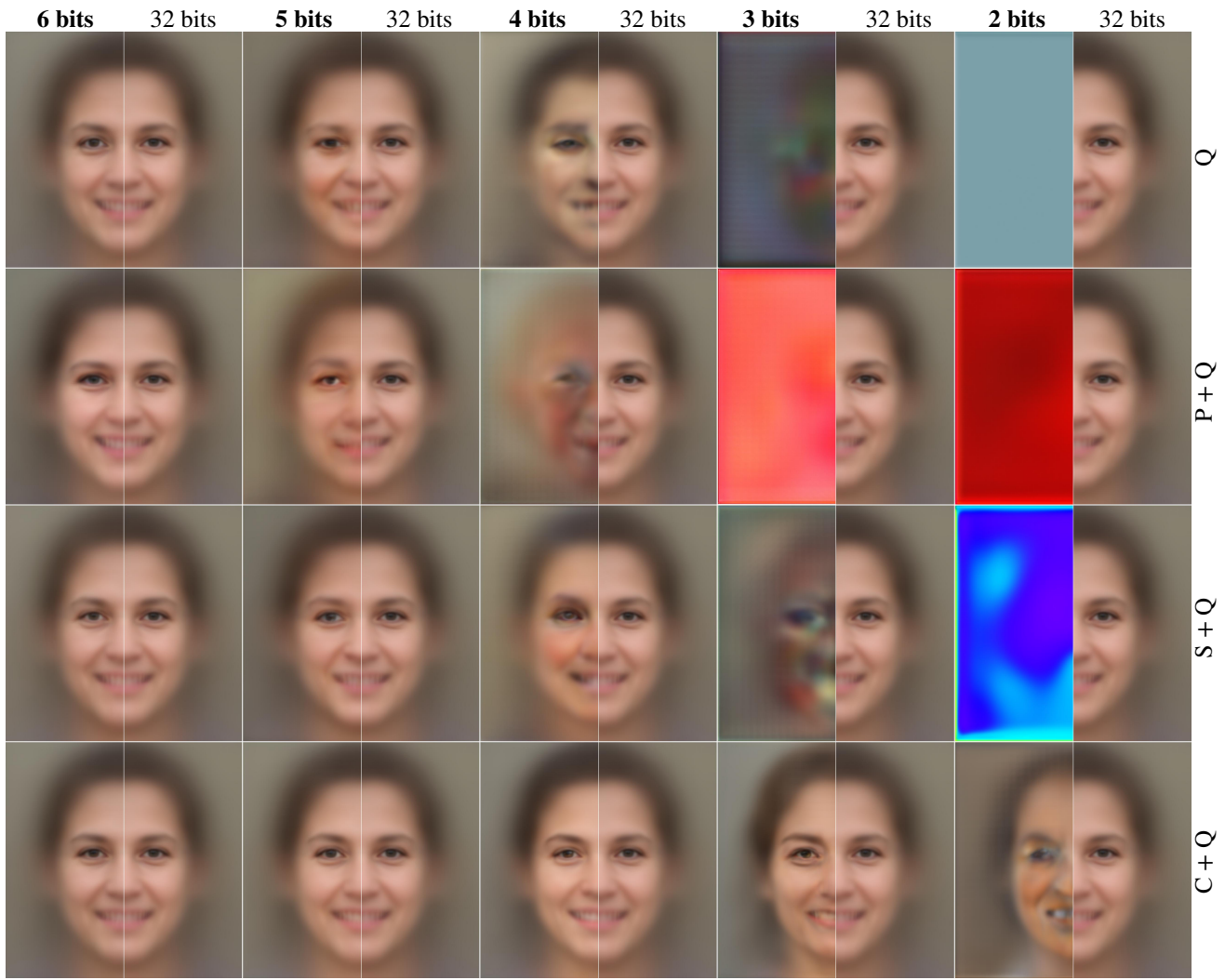}
    \begin{picture}(0,0)
        \put(-472,387){\bf6 bits}
        \put(-426,387){32 bits}
        \put(-375,387){\bf5 bits}
        \put(-330,387){32 bits}
        \put(-280,387){\bf4 bits}
        \put(-235,387){32 bits}
        \put(-185,387){\bf3 bits}
        \put(-135,387){32 bits}
        \put(-85,387){\bf2 bits}
        \put(-40,387){32 bits}
        \put(-1,330){\rotatebox{90}{Q}}
        \put(-1,226){\rotatebox{90}{P + Q}}
        \put(-1,130){\rotatebox{90}{S + Q}}
        \put(-1,34){\rotatebox{90}{C + Q}}
    \end{picture}     
    \caption{StyleGAN2 mean 10k images on FFHQ 1024x1024.}
    \label{fig:appendix_mean_images_2_to_6_bits_ffhq1024_stylegan2}
\end{figure*}

\begin{figure*}
    \begin{picture}(100,300)      
        \put(0,0){\includegraphics[width=0.98\textwidth]{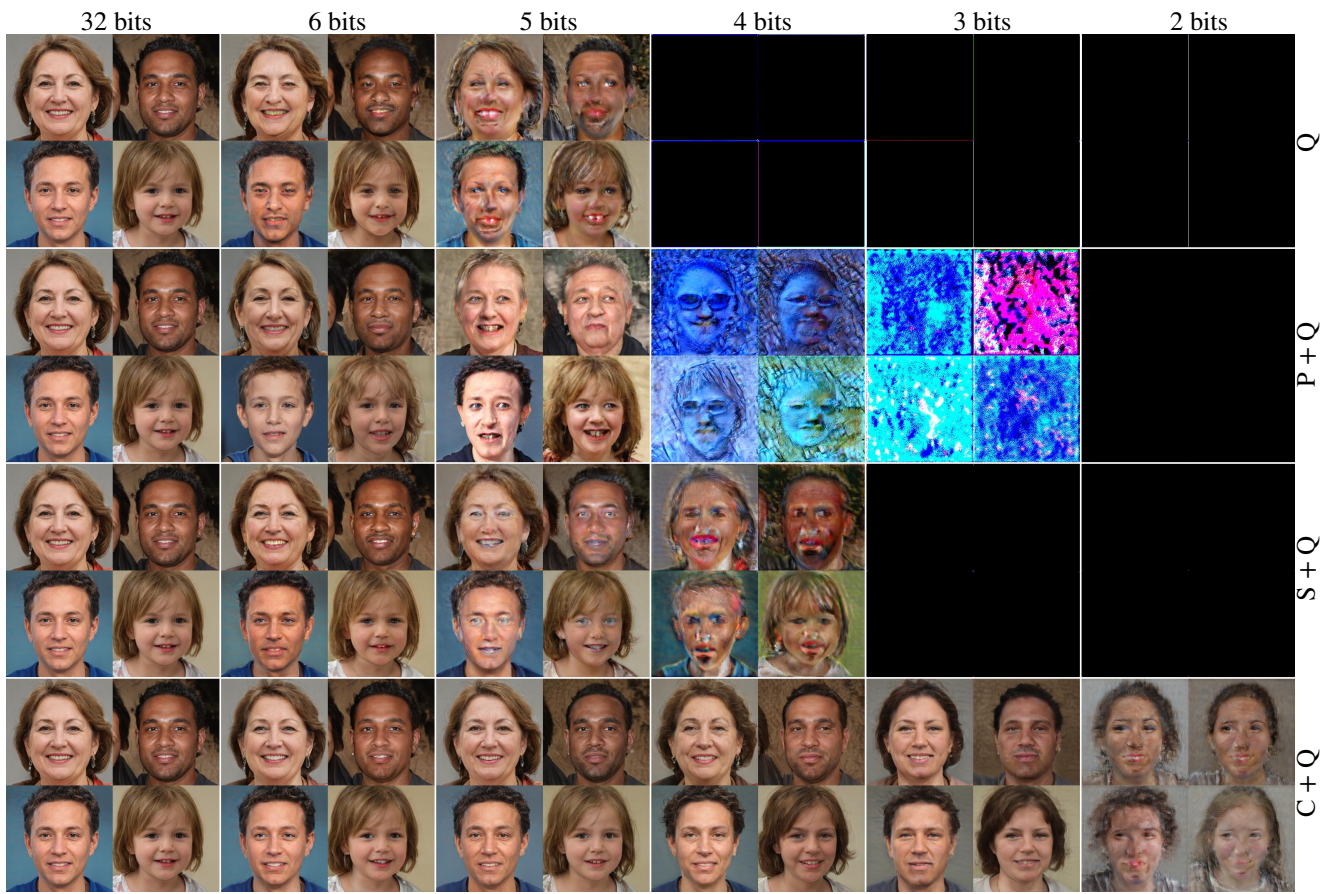}}
        \put(28,326){32 bits}
        \put(114,326){6 bits}
        \put(194,326){5 bits}
        \put(275,326){4 bits}
        \put(358,326){3 bits}
        \put(440,326){2 bits}
        \put(488,280){\rotatebox{90}{Q}}
        \put(488,191){\rotatebox{90}{P + Q}}
        \put(488,110){\rotatebox{90}{S + Q}}
        \put(488,28){\rotatebox{90}{C + Q}}
    \end{picture}     
    \caption{ADA on FFHQ 256x256.}
    \label{fig:appendix_ffhq_ada}
\end{figure*}

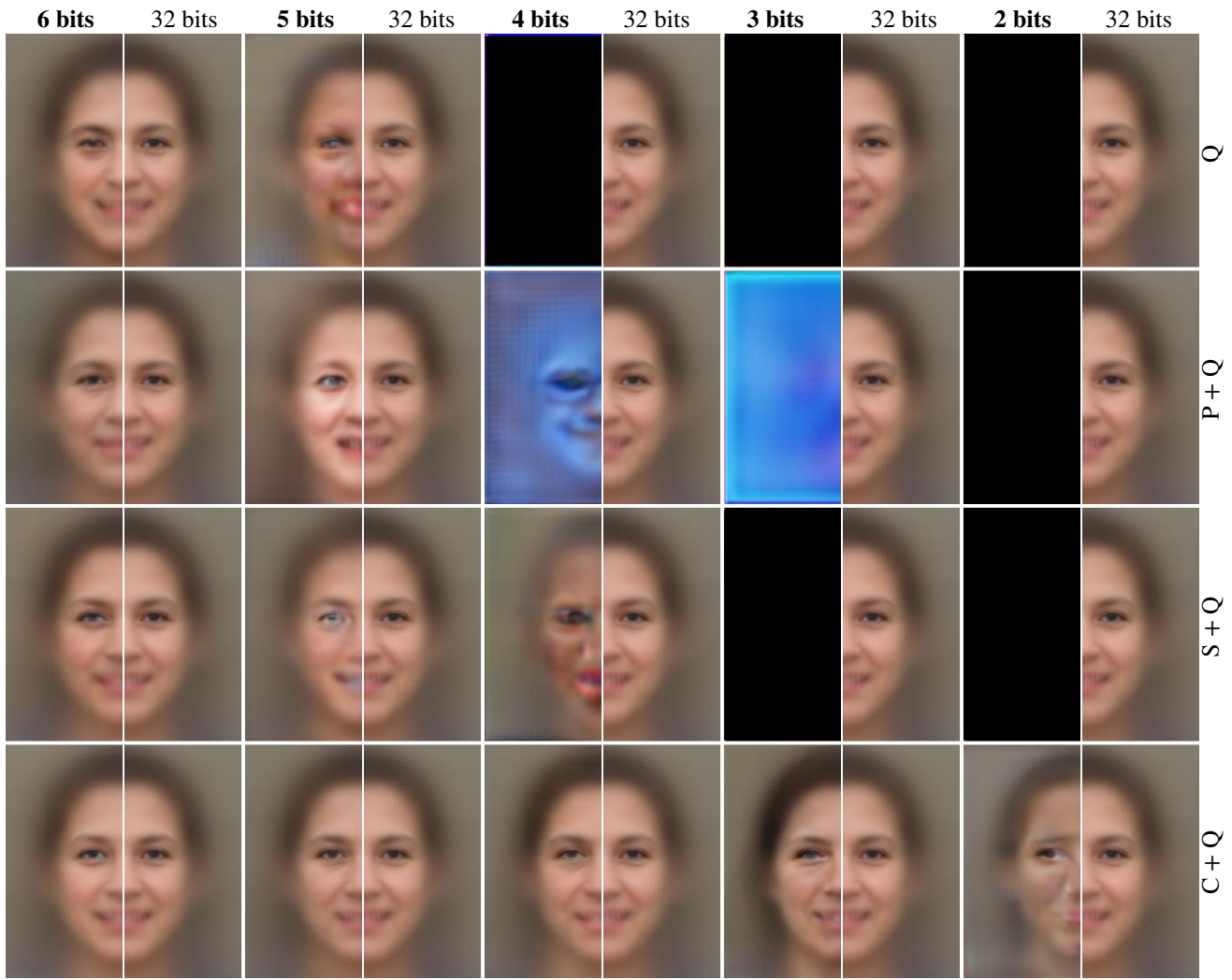
\begin{figure*}
    \includegraphics[width=0.97\textwidth]{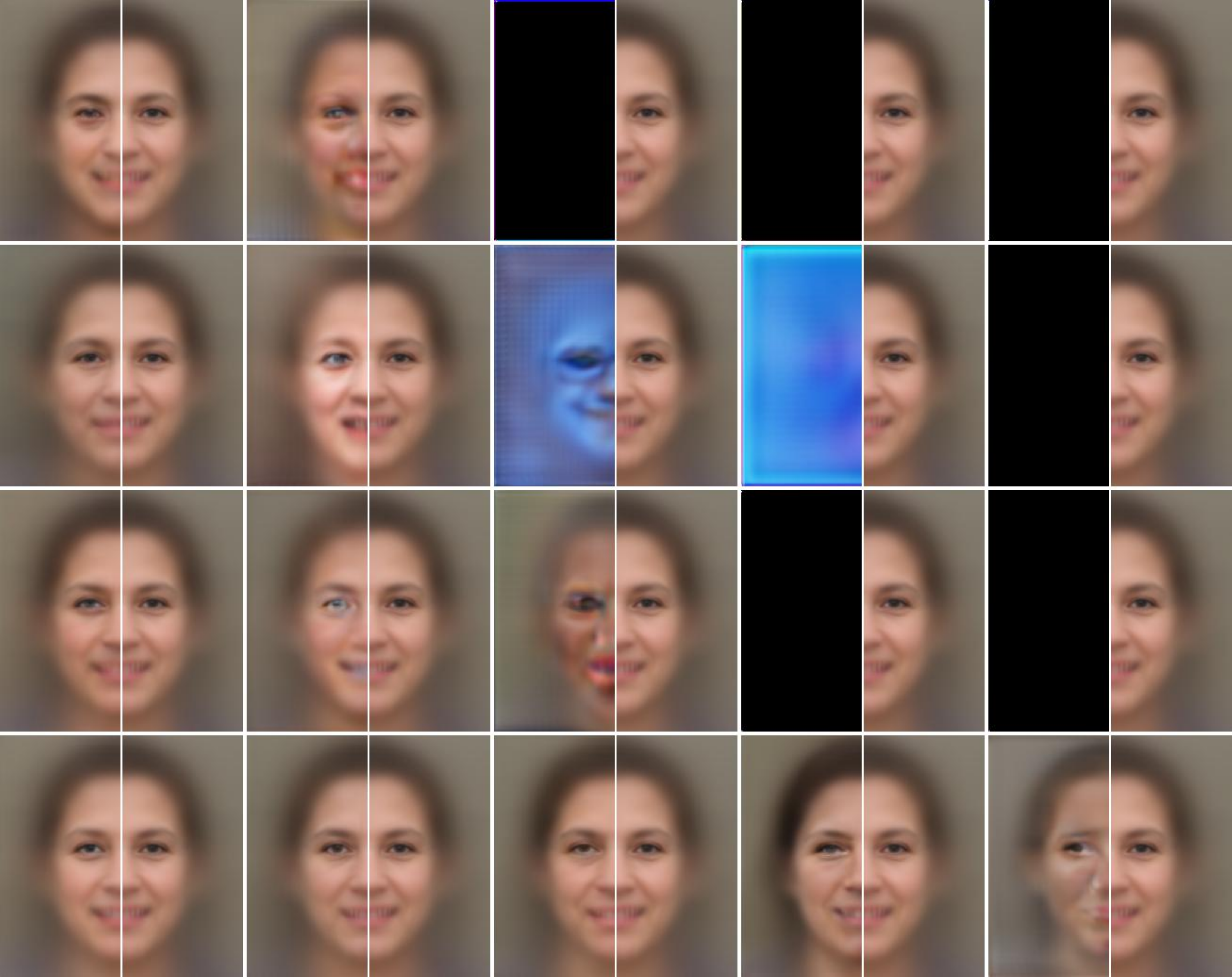}
    \begin{picture}(0,0)
        \put(-472,385){\bf6 bits}
        \put(-426,385){32 bits}
        \put(-375,385){\bf5 bits}
        \put(-330,385){32 bits}
        \put(-280,385){\bf4 bits}
        \put(-235,385){32 bits}
        \put(-185,385){\bf3 bits}
        \put(-135,385){32 bits}
        \put(-85,385){\bf2 bits}
        \put(-40,385){32 bits}
        \put(-1,330){\rotatebox{90}{Q}}
        \put(-1,226){\rotatebox{90}{P + Q}}
        \put(-1,130){\rotatebox{90}{S + Q}}
        \put(-1,34){\rotatebox{90}{C + Q}}
    \end{picture}     
    \caption{ADA mean 10k images on FFHQ 256x256.}
    \label{fig:appendix_mean_images_2_to_6_bits_ffhq_ada}
\end{figure*}

\begin{figure*}
    \begin{picture}(100,300)      
        \put(0,0){\includegraphics[width=0.98\textwidth]{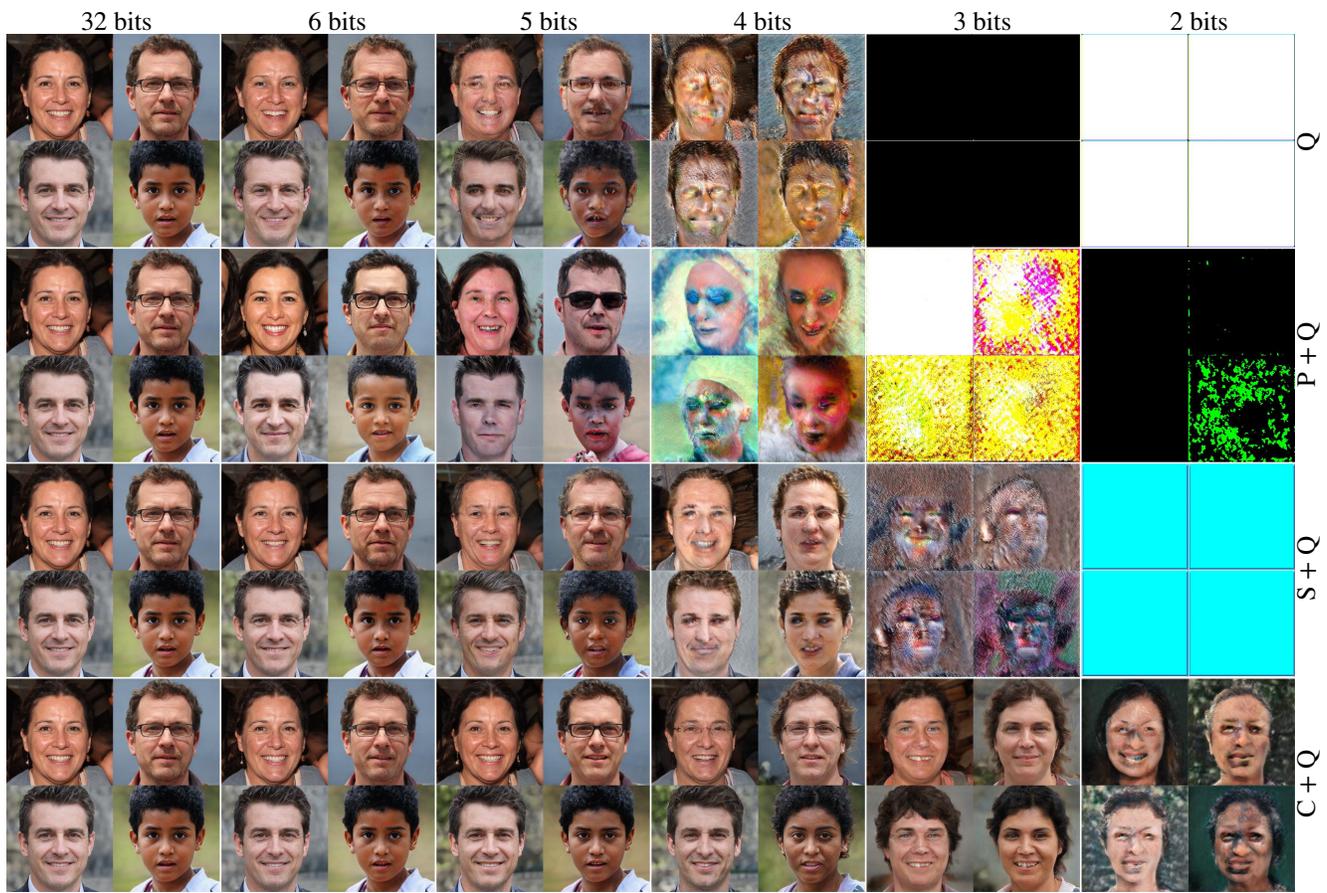}}
        \put(28,326){32 bits}
        \put(114,326){6 bits}
        \put(194,326){5 bits}
        \put(275,326){4 bits}
        \put(358,326){3 bits}
        \put(440,326){2 bits}
        \put(488,280){\rotatebox{90}{Q}}
        \put(488,191){\rotatebox{90}{P + Q}}
        \put(488,110){\rotatebox{90}{S + Q}}
        \put(488,28){\rotatebox{90}{C + Q}}
    \end{picture}     
    \caption{StyleGAN2 on FFHQ 256x256.}
    \label{fig:appendix_ffhq_stylegan2}
\end{figure*}

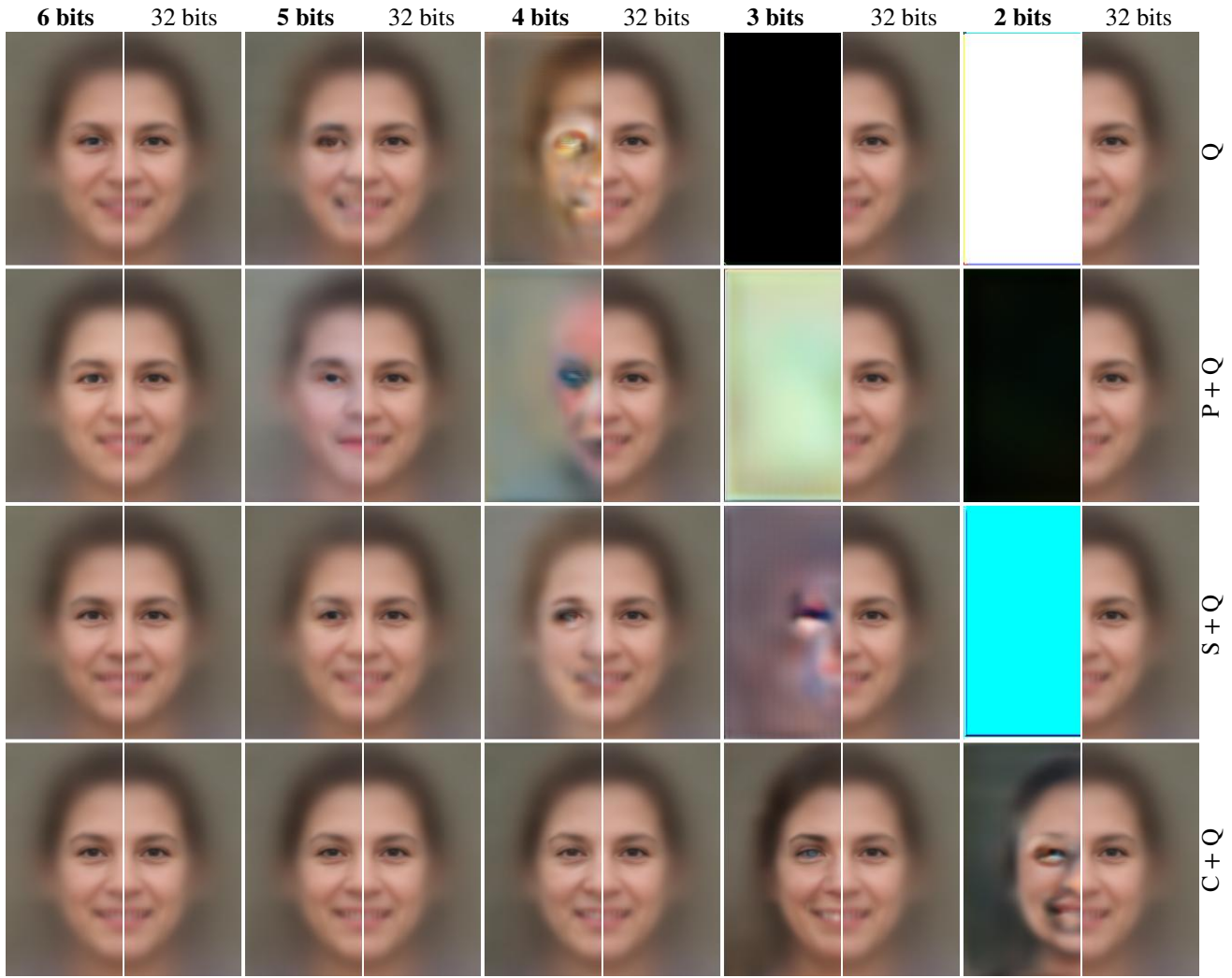
\begin{figure*}
    \includegraphics[width=0.97\textwidth]{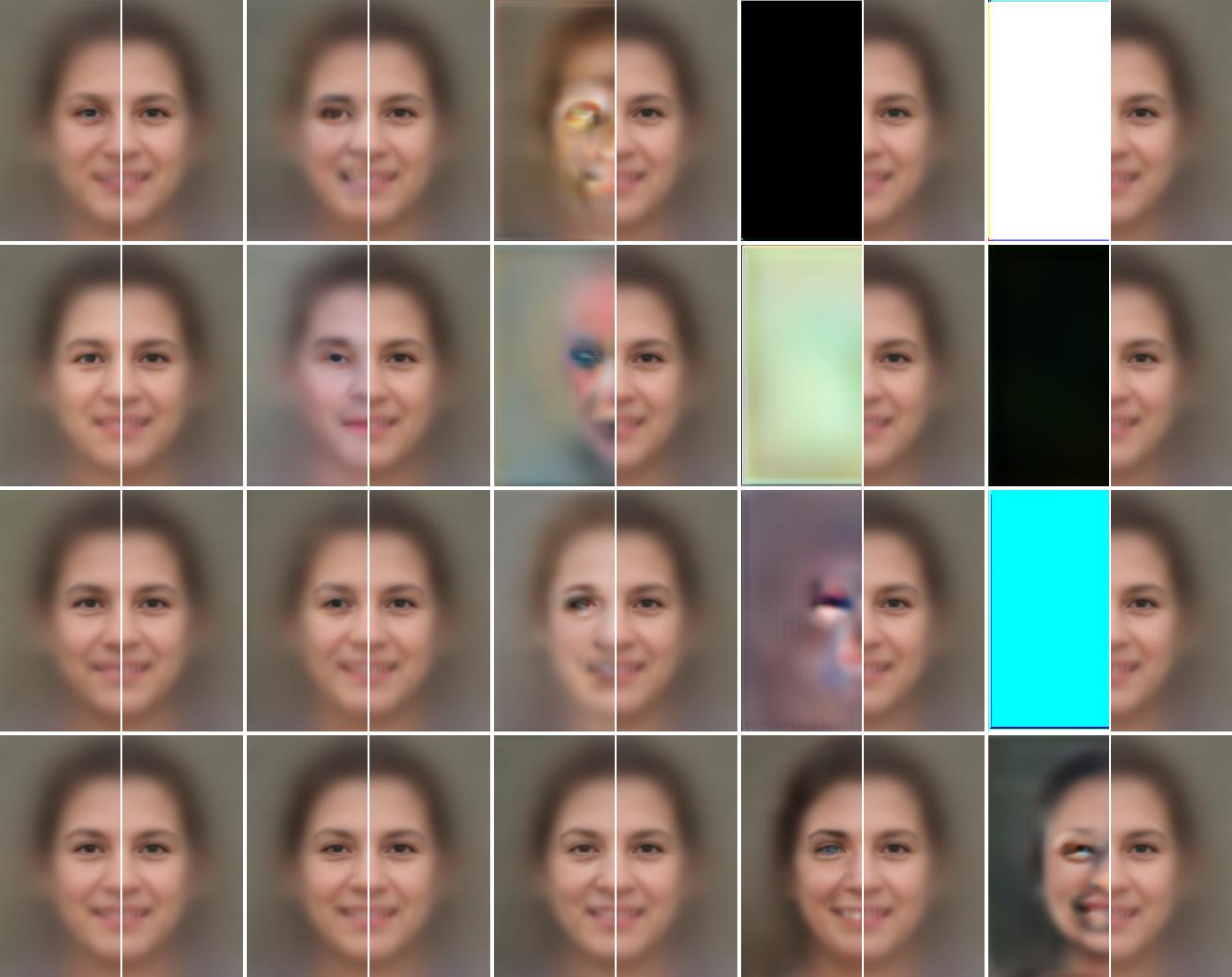}
    \begin{picture}(0,0)
        \put(-472,385){\bf6 bits}
        \put(-426,385){32 bits}
        \put(-375,385){\bf5 bits}
        \put(-330,385){32 bits}
        \put(-280,385){\bf4 bits}
        \put(-235,385){32 bits}
        \put(-185,385){\bf3 bits}
        \put(-135,385){32 bits}
        \put(-85,385){\bf2 bits}
        \put(-40,385){32 bits}
        \put(-1,330){\rotatebox{90}{Q}}
        \put(-1,226){\rotatebox{90}{P + Q}}
        \put(-1,130){\rotatebox{90}{S + Q}}
        \put(-1,34){\rotatebox{90}{C + Q}}
    \end{picture}     
    \caption{StyleGAN2 mean 10k images on FFHQ 256x256.}
    \label{fig:appendix_mean_images_2_to_6_bits_ffhq_stylegan2}
\end{figure*}

\begin{figure*}
    \begin{picture}(100,300)      
        \put(0,0){\includegraphics[width=0.98\textwidth]{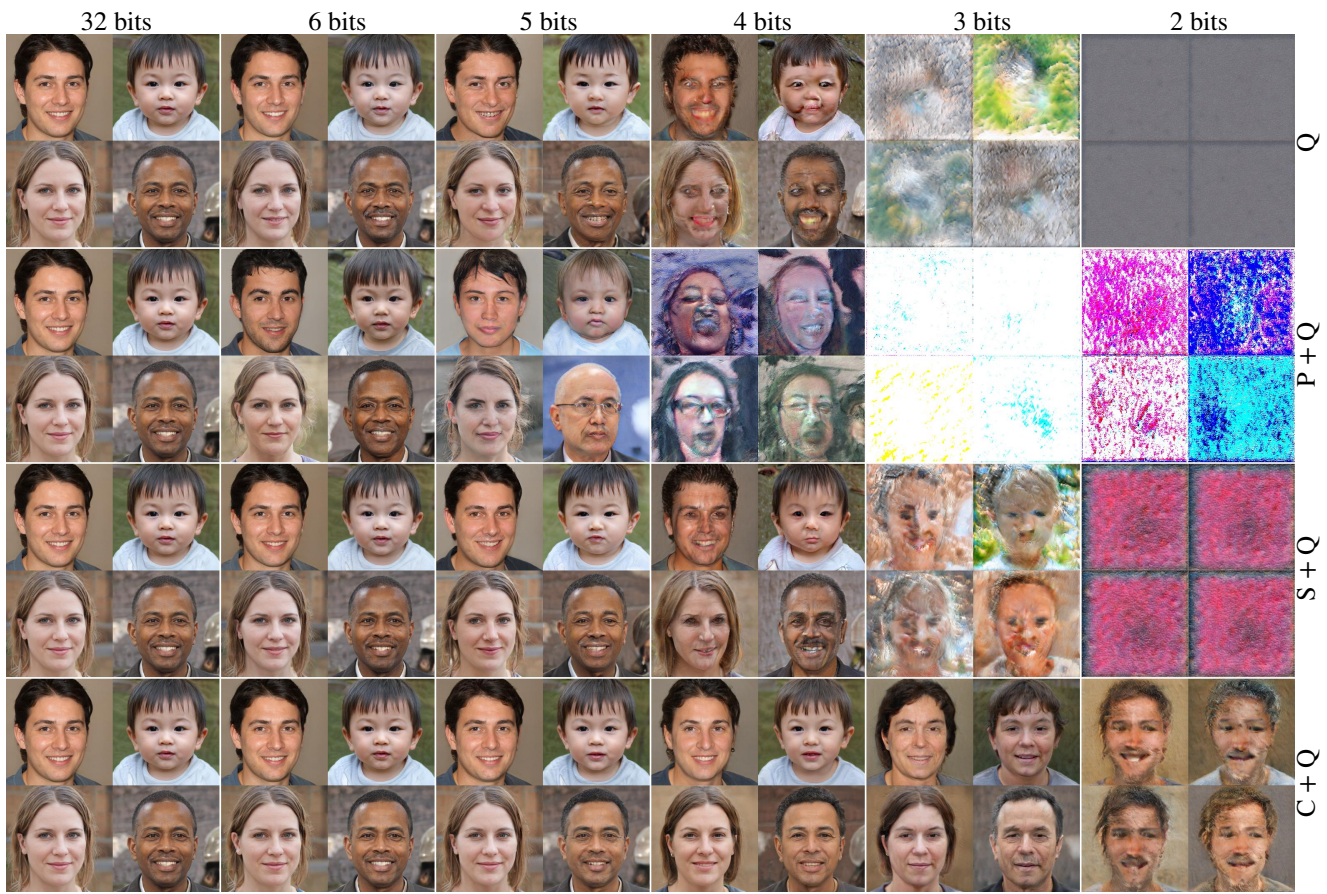}}
        \put(28,326){32 bits}
        \put(114,326){6 bits}
        \put(194,326){5 bits}
        \put(275,326){4 bits}
        \put(358,326){3 bits}
        \put(440,326){2 bits}
        \put(488,280){\rotatebox{90}{Q}}
        \put(488,191){\rotatebox{90}{P + Q}}
        \put(488,110){\rotatebox{90}{S + Q}}
        \put(488,28){\rotatebox{90}{C + Q}}
    \end{picture}     
    \caption{zCR-GAN on FFHQ 256x256.}
    \label{fig:appendix_ffhq_zcr}
\end{figure*}

\begin{figure*}
    \includegraphics[width=0.97\textwidth]{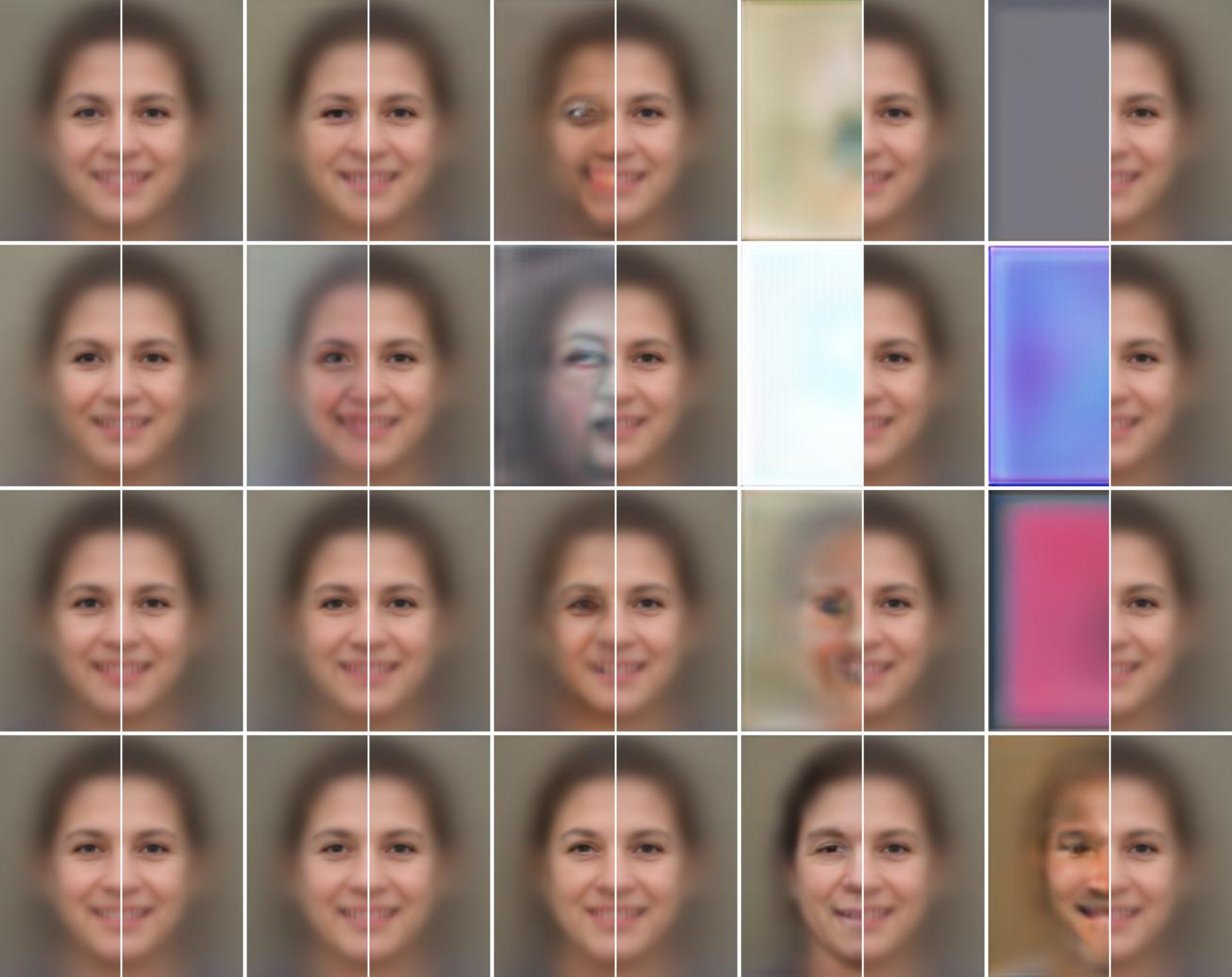}
    \begin{picture}(0,0)
        \put(-472,385){\bf6 bits}
        \put(-426,385){32 bits}
        \put(-375,385){\bf5 bits}
        \put(-330,385){32 bits}
        \put(-280,385){\bf4 bits}
        \put(-235,385){32 bits}
        \put(-185,385){\bf3 bits}
        \put(-135,385){32 bits}
        \put(-85,385){\bf2 bits}
        \put(-40,385){32 bits}
        \put(-1,330){\rotatebox{90}{Q}}
        \put(-1,226){\rotatebox{90}{P + Q}}
        \put(-1,130){\rotatebox{90}{S + Q}}
        \put(-1,34){\rotatebox{90}{C + Q}}
    \end{picture}     
    \caption{zCR-GAN mean 10k images on FFHQ 256x256.}
    \label{fig:appendix_mean_images_2_to_6_bits_ffhq_zcr}
\end{figure*}
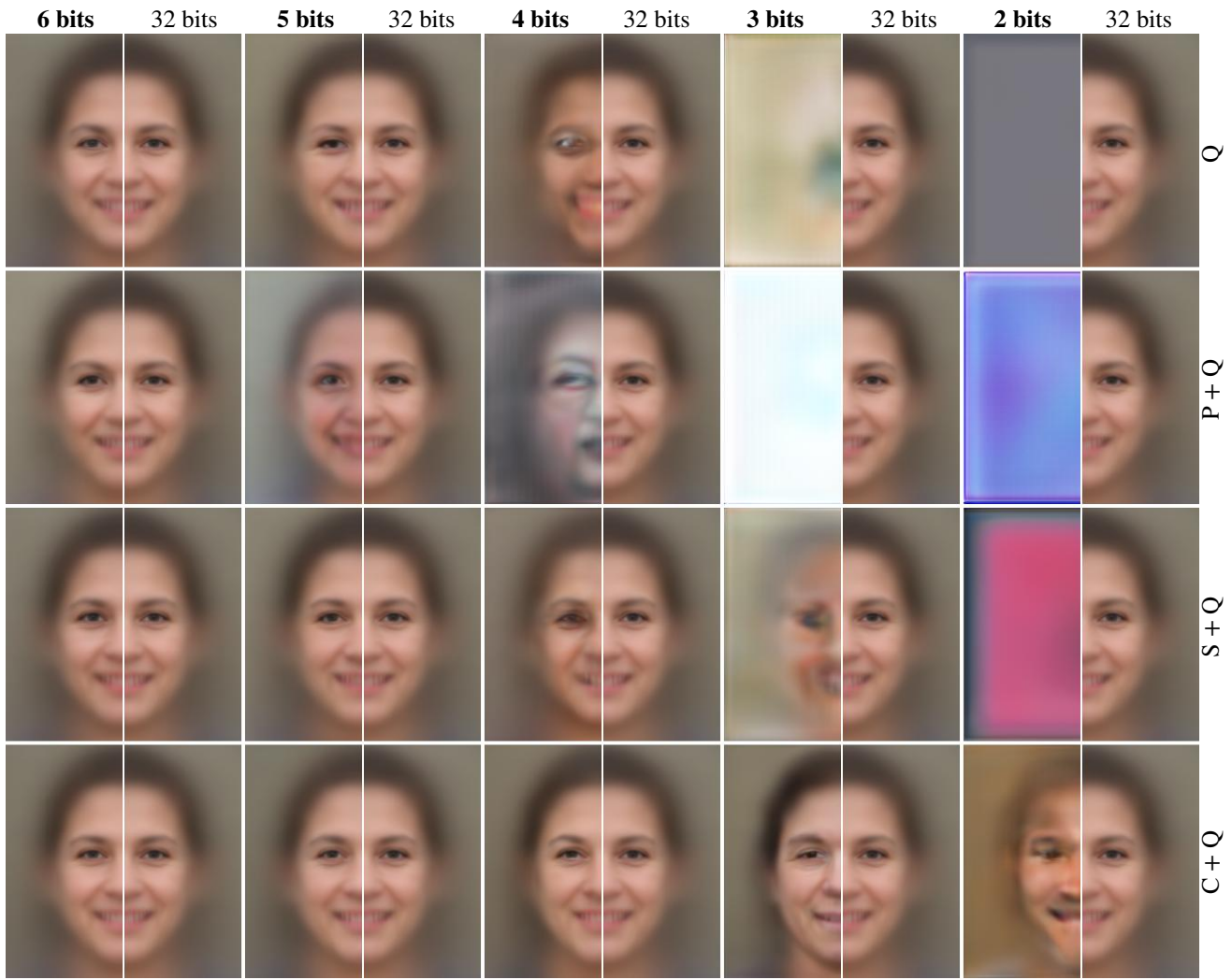

\begin{figure*}
    \begin{picture}(100,300)      
        \put(0,0){\includegraphics[width=0.98\textwidth]{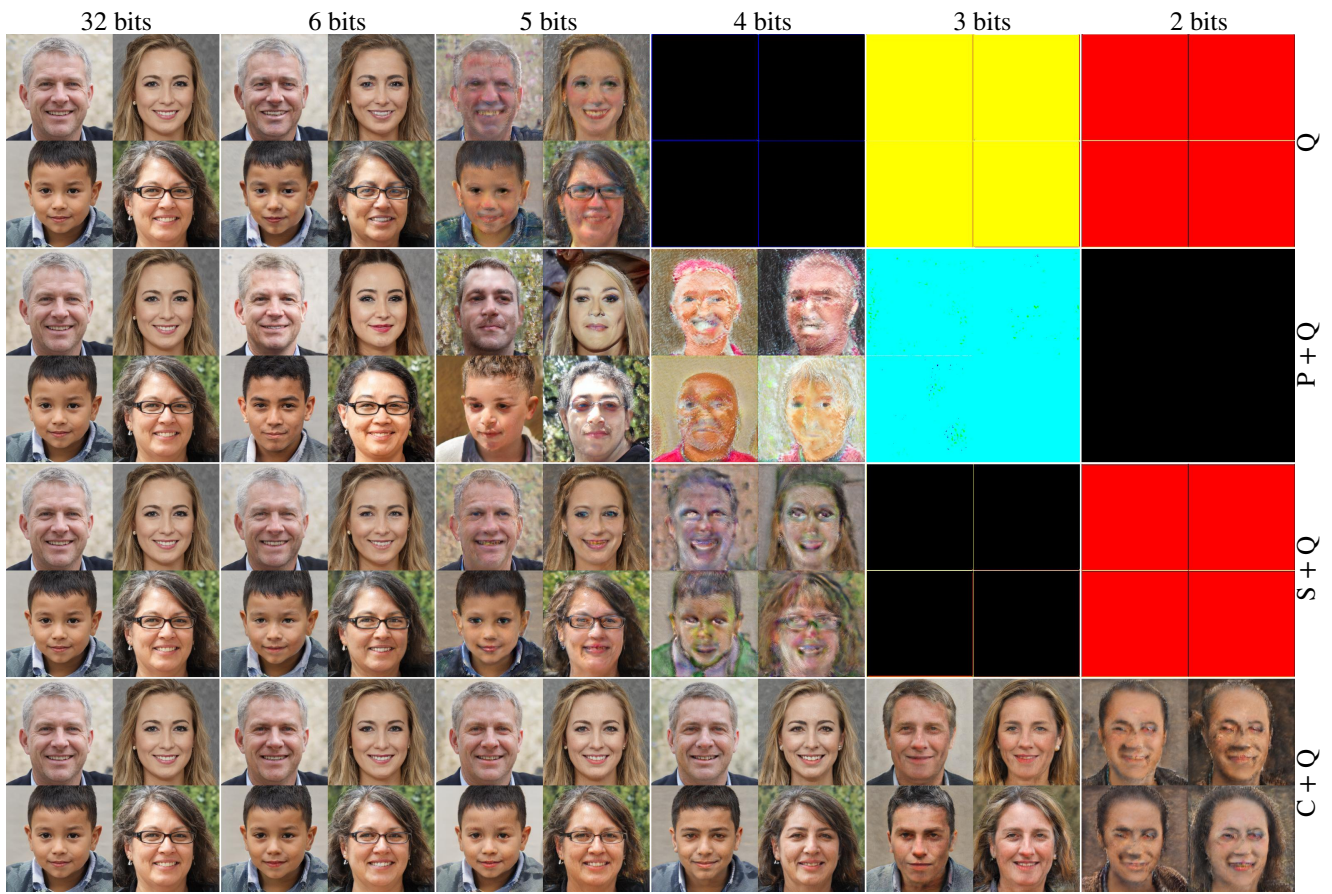}}
        \put(28,326){32 bits}
        \put(114,326){6 bits}
        \put(194,326){5 bits}
        \put(275,326){4 bits}
        \put(358,326){3 bits}
        \put(440,326){2 bits}
        \put(488,280){\rotatebox{90}{Q}}
        \put(488,191){\rotatebox{90}{P + Q}}
        \put(488,110){\rotatebox{90}{S + Q}}
        \put(488,28){\rotatebox{90}{C + Q}}
    \end{picture}     
    \caption{SN-GAN on FFHQ 256x256.}
    \label{fig:appendix_ffhq_sngan}
\end{figure*}

\begin{figure*}
    \includegraphics[width=0.97\textwidth]{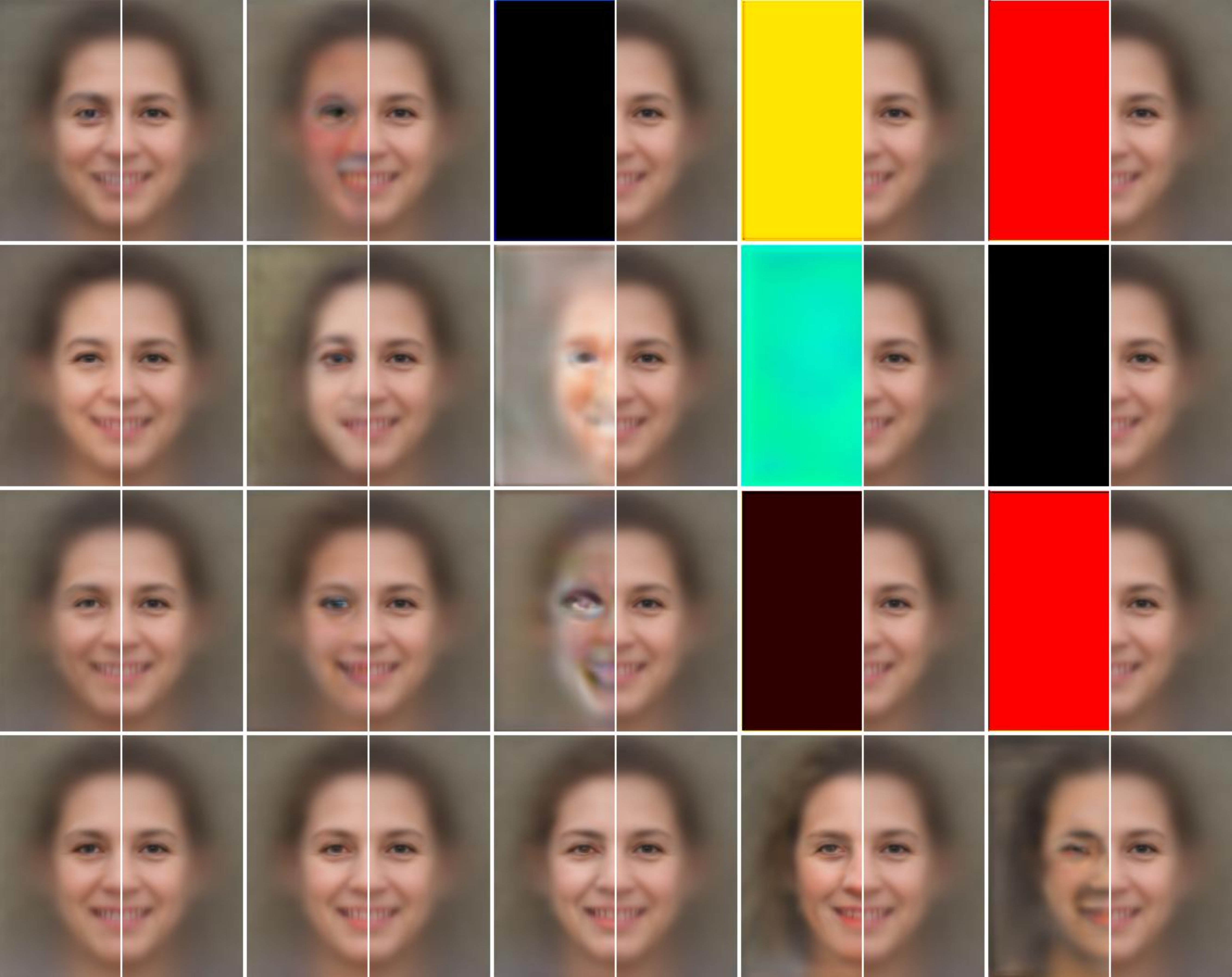}
    \begin{picture}(0,0)
        \put(-472,385){\bf6 bits}
        \put(-426,385){32 bits}
        \put(-375,385){\bf5 bits}
        \put(-330,385){32 bits}
        \put(-280,385){\bf4 bits}
        \put(-235,385){32 bits}
        \put(-185,385){\bf3 bits}
        \put(-135,385){32 bits}
        \put(-85,385){\bf2 bits}
        \put(-40,385){32 bits}
        \put(-1,330){\rotatebox{90}{Q}}
        \put(-1,226){\rotatebox{90}{P + Q}}
        \put(-1,130){\rotatebox{90}{S + Q}}
        \put(-1,34){\rotatebox{90}{C + Q}}
    \end{picture}     
    \caption{SN-GAN mean 10k images on FFHQ 256x256.}
    \label{fig:appendix_mean_images_2_to_6_bits_ffhq_sngan}
\end{figure*}

\begin{figure*}
    \begin{picture}(100,300)      
        \put(0,0){\includegraphics[width=0.98\textwidth]{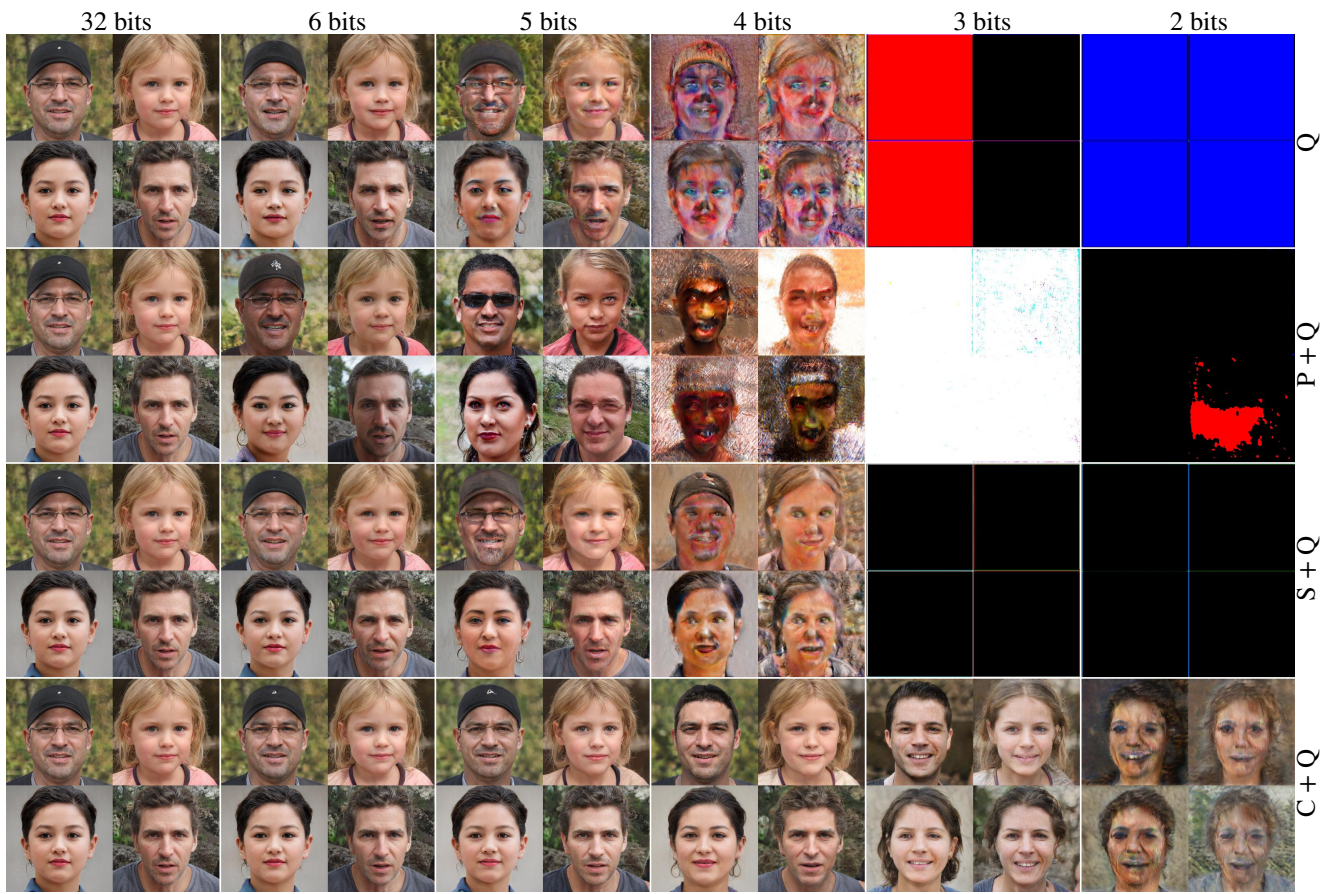}}
        \put(28,326){32 bits}
        \put(114,326){6 bits}
        \put(194,326){5 bits}
        \put(275,326){4 bits}
        \put(358,326){3 bits}
        \put(440,326){2 bits}
        \put(488,280){\rotatebox{90}{Q}}
        \put(488,191){\rotatebox{90}{P + Q}}
        \put(488,110){\rotatebox{90}{S + Q}}
        \put(488,28){\rotatebox{90}{C + Q}}
    \end{picture}     
    \caption{PA-GAN on FFHQ 256x256.}
    \label{fig:appendix_ffhq_pagan}
\end{figure*}

\begin{figure*}
    \includegraphics[width=0.97\textwidth]{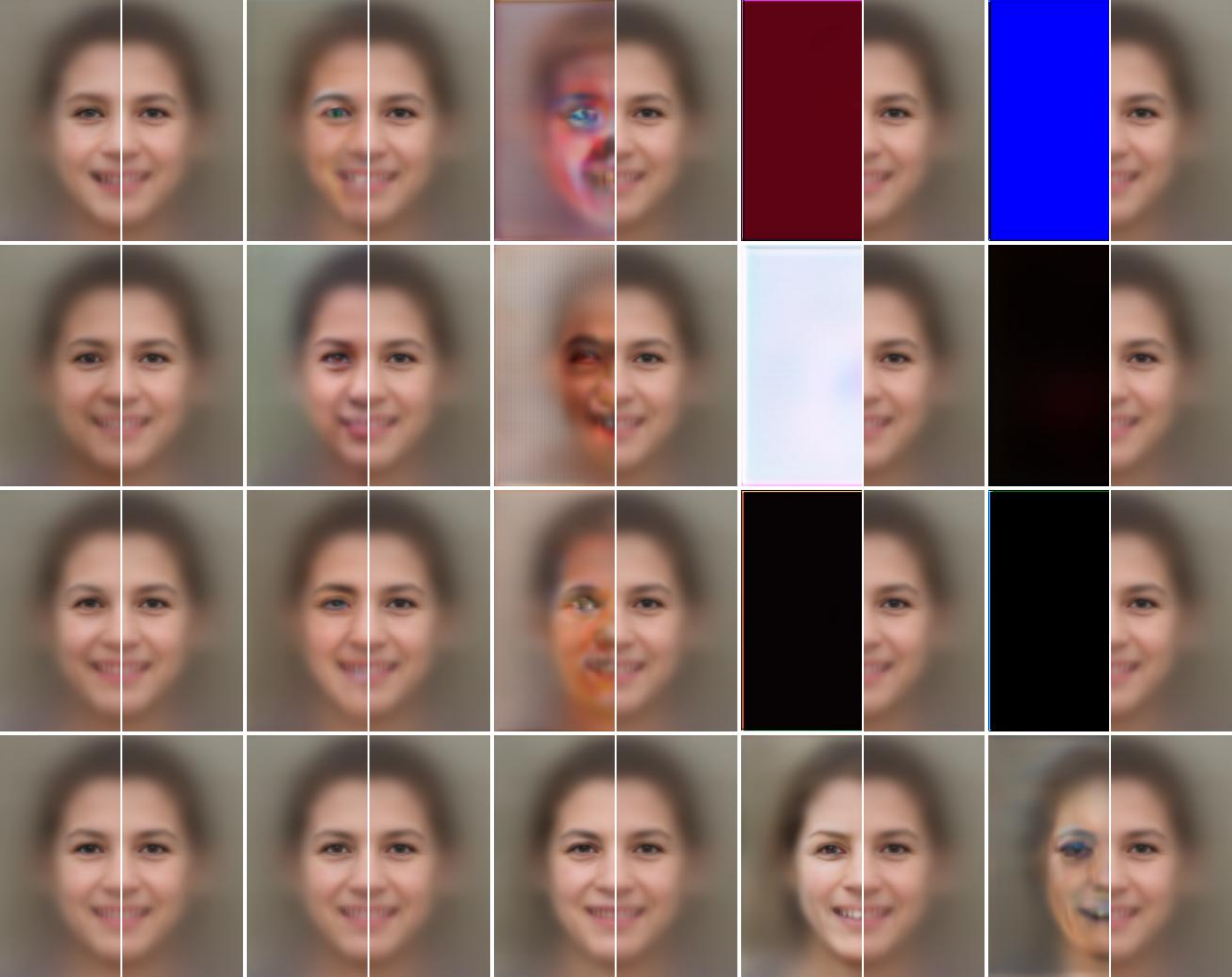}
    \begin{picture}(0,0)
        \put(-472,385){\bf6 bits}
        \put(-426,385){32 bits}
        \put(-375,385){\bf5 bits}
        \put(-330,385){32 bits}
        \put(-280,385){\bf4 bits}
        \put(-235,385){32 bits}
        \put(-185,385){\bf3 bits}
        \put(-135,385){32 bits}
        \put(-85,385){\bf2 bits}
        \put(-40,385){32 bits}
        \put(-1,330){\rotatebox{90}{Q}}
        \put(-1,226){\rotatebox{90}{P + Q}}
        \put(-1,130){\rotatebox{90}{S + Q}}
        \put(-1,34){\rotatebox{90}{C + Q}}
    \end{picture}     
    \caption{PA-GAN mean 10k images on FFHQ 256x256.}
    \label{fig:appendix_mean_images_2_to_6_bits_ffhq_pagan}
\end{figure*}
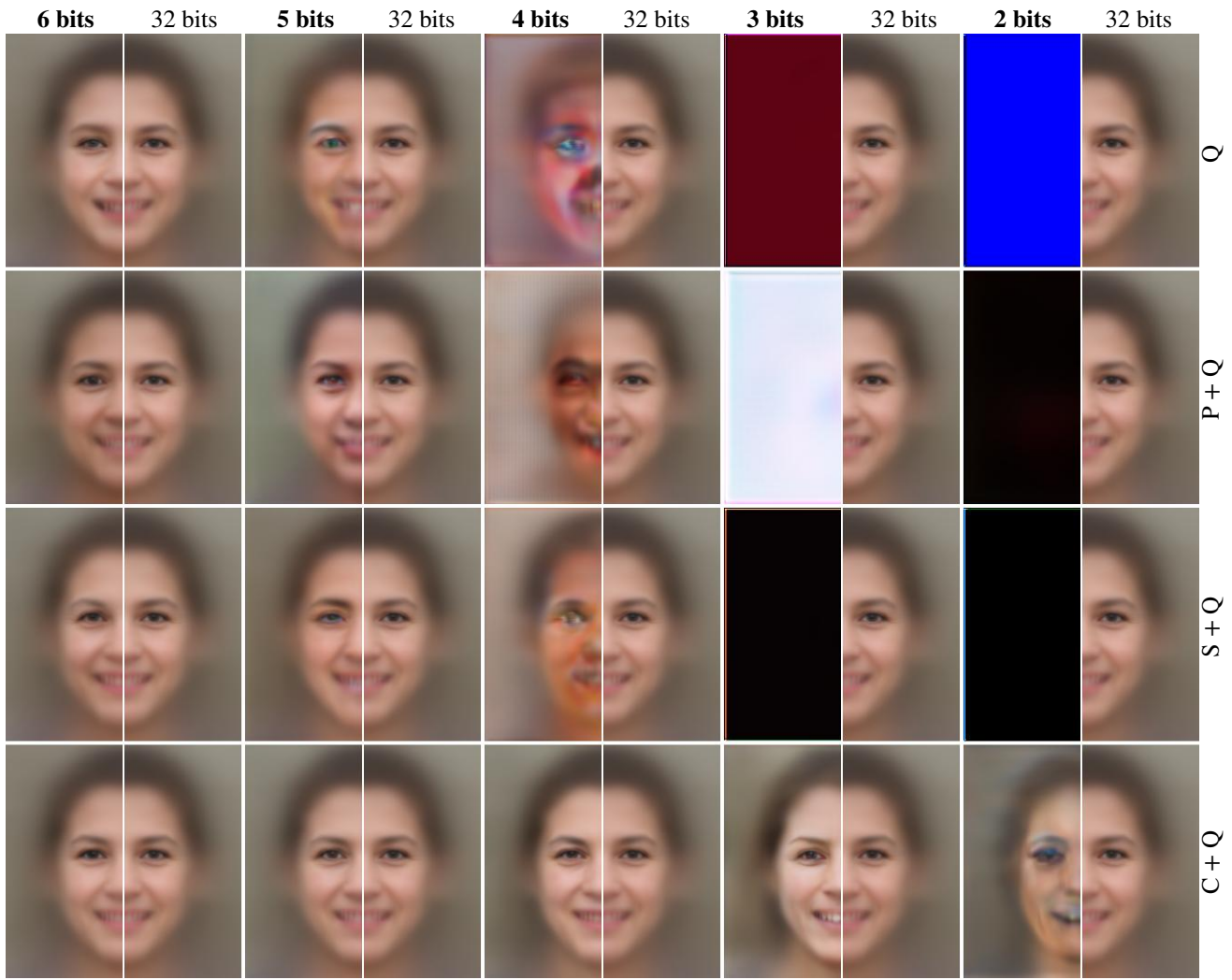

\begin{figure*}
    \begin{picture}(100,300)      
        \put(0,0){\includegraphics[width=0.98\textwidth]{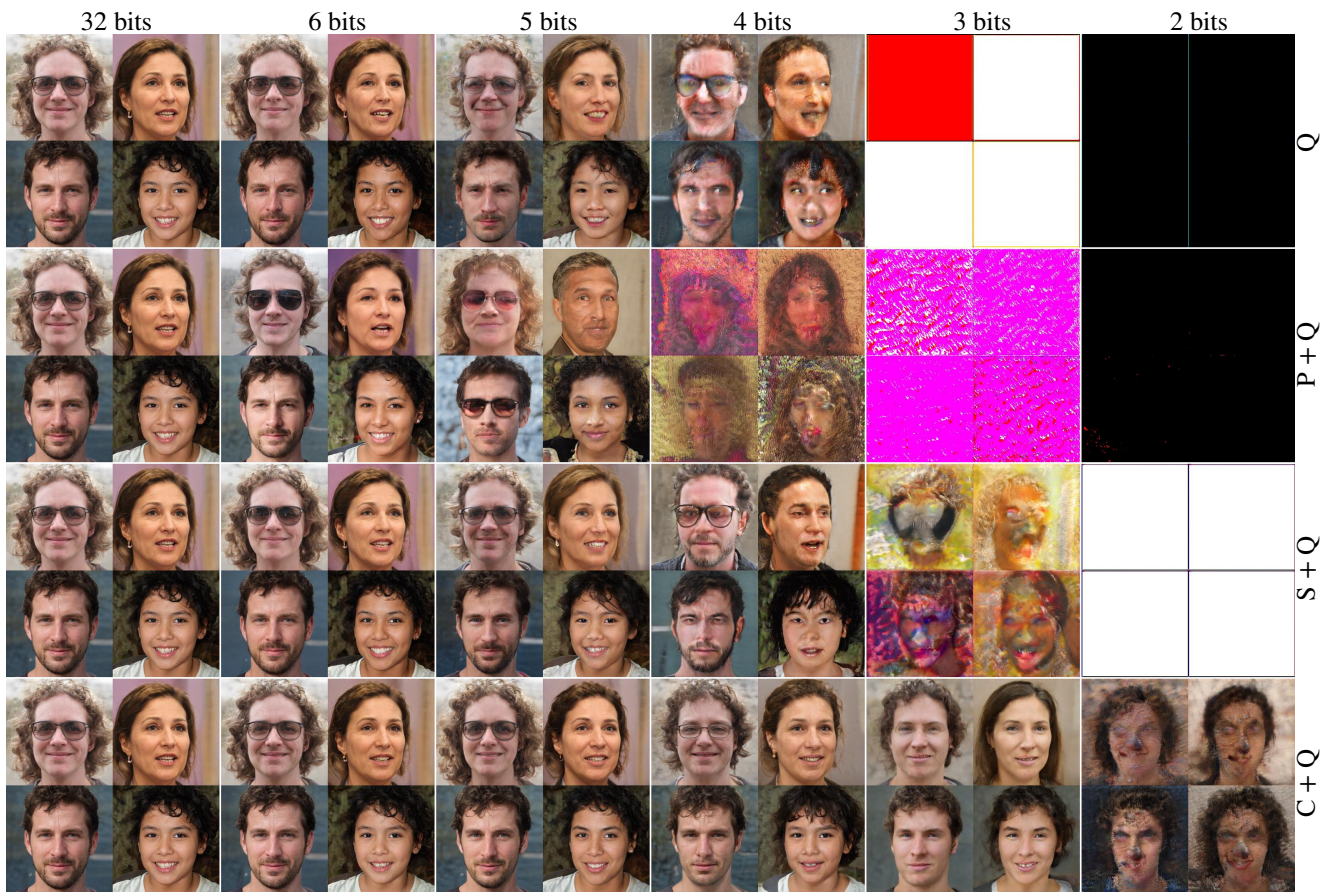}}
        \put(28,326){32 bits}
        \put(114,326){6 bits}
        \put(194,326){5 bits}
        \put(275,326){4 bits}
        \put(358,326){3 bits}
        \put(440,326){2 bits}
        \put(488,280){\rotatebox{90}{Q}}
        \put(488,191){\rotatebox{90}{P + Q}}
        \put(488,110){\rotatebox{90}{S + Q}}
        \put(488,28){\rotatebox{90}{C + Q}}
    \end{picture}     
    \caption{SS-GAN on FFHQ 256x256.}
    \label{fig:appendix_ffhq_ssgan}
\end{figure*}

\begin{figure*}
    \includegraphics[width=0.97\textwidth]{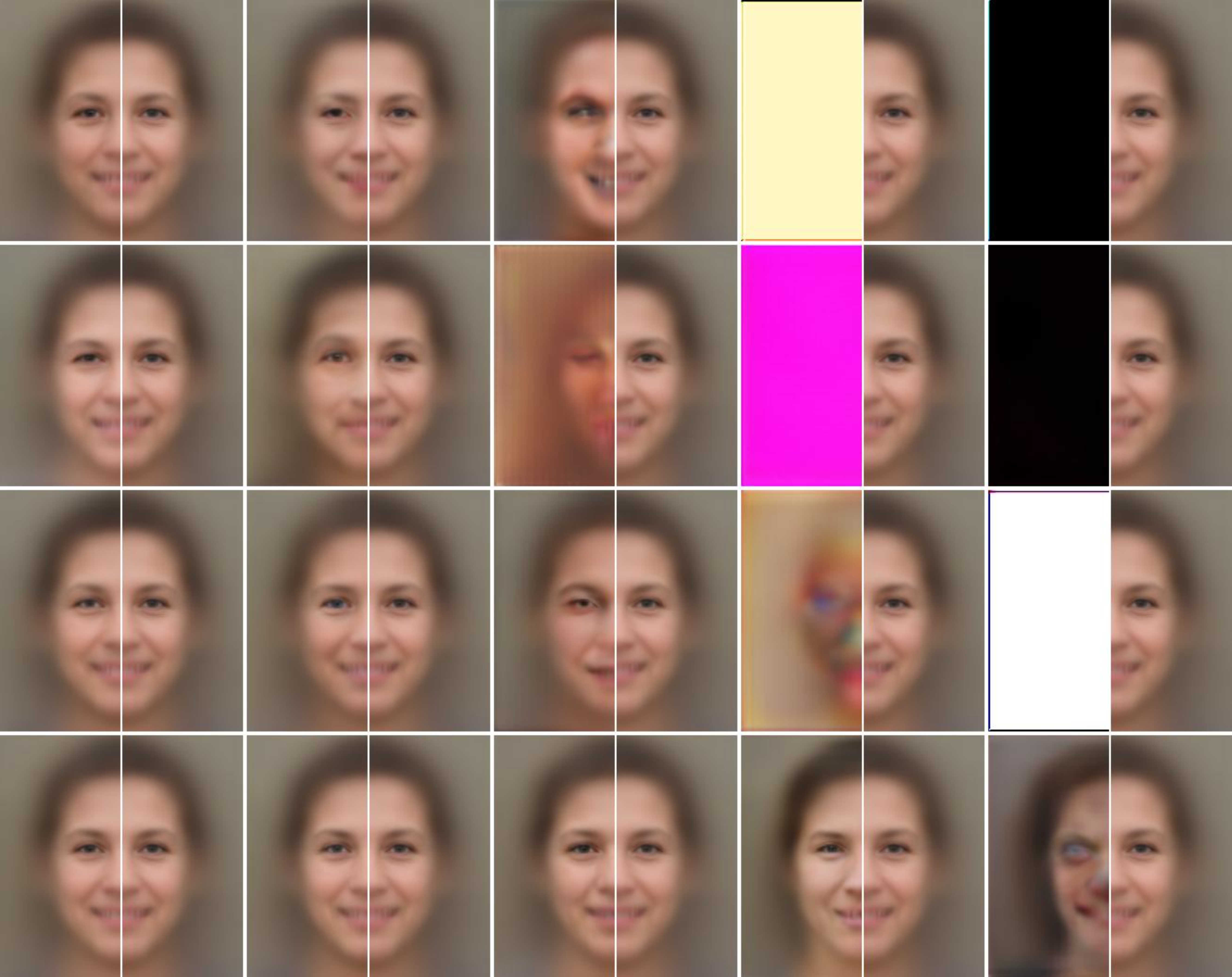}
    \begin{picture}(0,0)
        \put(-472,385){\bf6 bits}
        \put(-426,385){32 bits}
        \put(-375,385){\bf5 bits}
        \put(-330,385){32 bits}
        \put(-280,385){\bf4 bits}
        \put(-235,385){32 bits}
        \put(-185,385){\bf3 bits}
        \put(-135,385){32 bits}
        \put(-85,385){\bf2 bits}
        \put(-40,385){32 bits}
        \put(-1,330){\rotatebox{90}{Q}}
        \put(-1,226){\rotatebox{90}{P + Q}}
        \put(-1,130){\rotatebox{90}{S + Q}}
        \put(-1,34){\rotatebox{90}{C + Q}}
    \end{picture}     
    \caption{SS-GAN mean 10k images on FFHQ 256x256.}
    \label{fig:appendix_mean_images_2_to_6_bits_ffhq_ssgan}
\end{figure*}

\begin{figure*}
    \begin{picture}(100,300)      
        \put(0,0){\includegraphics[width=0.98\textwidth]{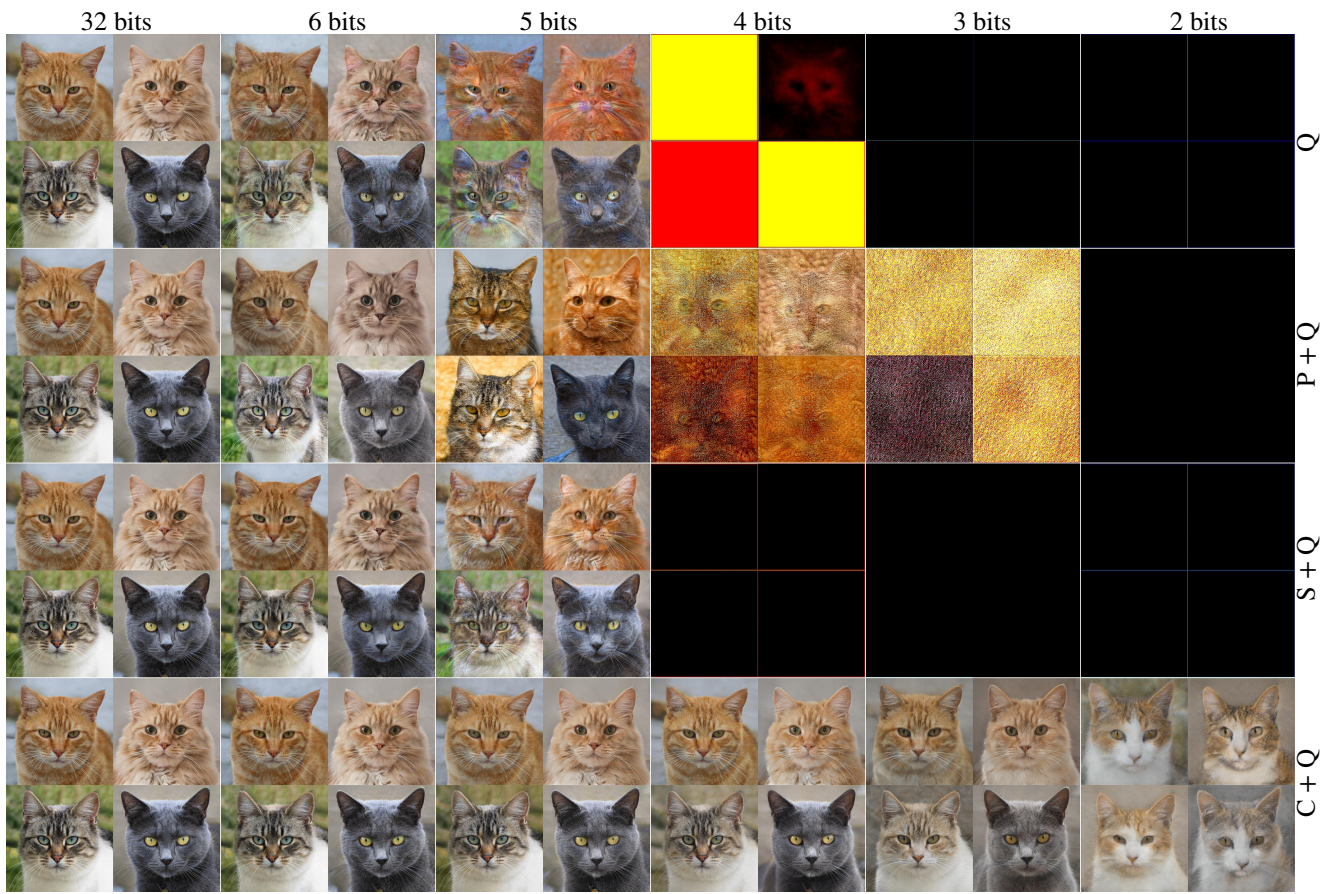}}
        \put(28,326){32 bits}
        \put(114,326){6 bits}
        \put(194,326){5 bits}
        \put(275,326){4 bits}
        \put(358,326){3 bits}
        \put(440,326){2 bits}
        \put(488,280){\rotatebox{90}{Q}}
        \put(488,191){\rotatebox{90}{P + Q}}
        \put(488,110){\rotatebox{90}{S + Q}}
        \put(488,28){\rotatebox{90}{C + Q}}
    \end{picture}     
    \caption{ADA on AFHQ Cat 512x512.}
    \label{fig:appendix_afhqcat_ada}
\end{figure*}

\begin{figure*}
    \includegraphics[width=0.97\textwidth]{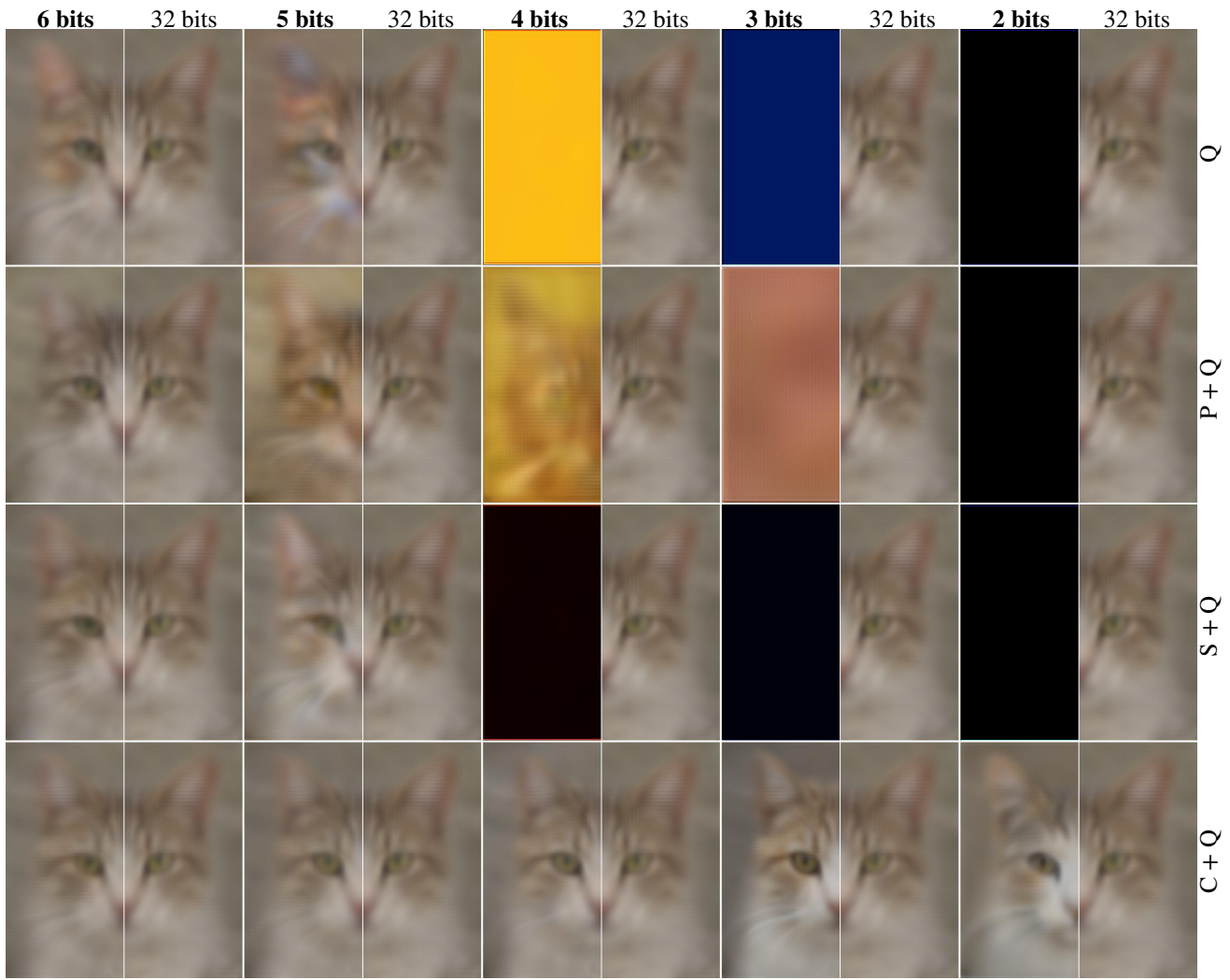}
    \begin{picture}(0,0)
        \put(-472,385){\bf6 bits}
        \put(-426,385){32 bits}
        \put(-375,385){\bf5 bits}
        \put(-330,385){32 bits}
        \put(-280,385){\bf4 bits}
        \put(-235,385){32 bits}
        \put(-185,385){\bf3 bits}
        \put(-135,385){32 bits}
        \put(-85,385){\bf2 bits}
        \put(-40,385){32 bits}
        \put(-1,330){\rotatebox{90}{Q}}
        \put(-1,226){\rotatebox{90}{P + Q}}
        \put(-1,130){\rotatebox{90}{S + Q}}
        \put(-1,34){\rotatebox{90}{C + Q}}
    \end{picture}     
    \caption{ADA mean 10k images on AFHQ Cat 512x512.}
    \label{fig:appendix_mean_images_2_to_6_bits_afhqcat_ada}
\end{figure*}

\begin{figure*}
    \begin{picture}(100,300)      
        \put(0,0){\includegraphics[width=0.98\textwidth]{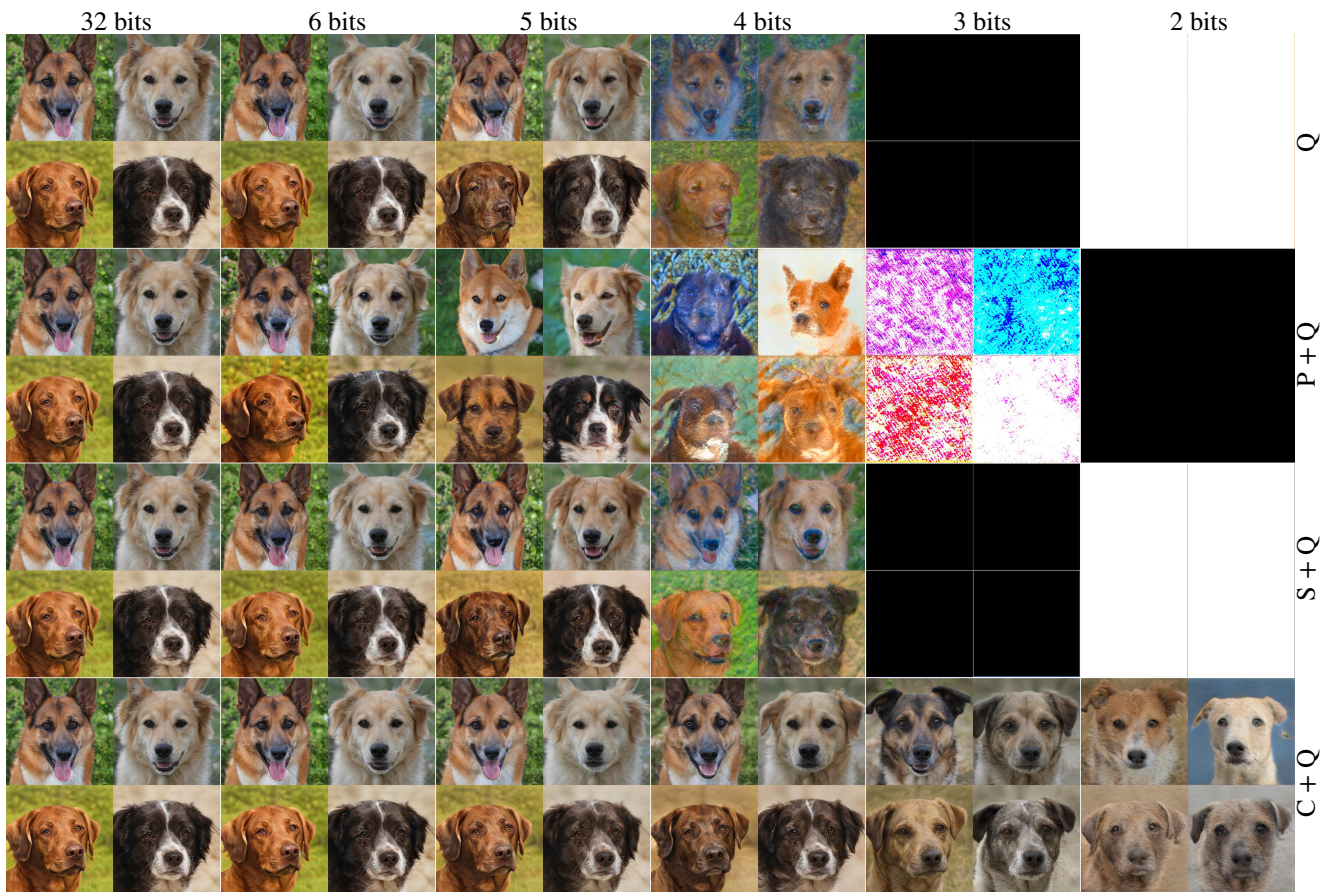}}
        \put(28,326){32 bits}
        \put(114,326){6 bits}
        \put(194,326){5 bits}
        \put(275,326){4 bits}
        \put(358,326){3 bits}
        \put(440,326){2 bits}
        \put(488,280){\rotatebox{90}{Q}}
        \put(488,191){\rotatebox{90}{P + Q}}
        \put(488,110){\rotatebox{90}{S + Q}}
        \put(488,28){\rotatebox{90}{C + Q}}
    \end{picture}     
    \caption{ADA on AFHQ Dog 512x512.}
    \label{fig:appendix_afhqdog_ada}
\end{figure*}

\begin{figure*}
    \includegraphics[width=0.97\textwidth]{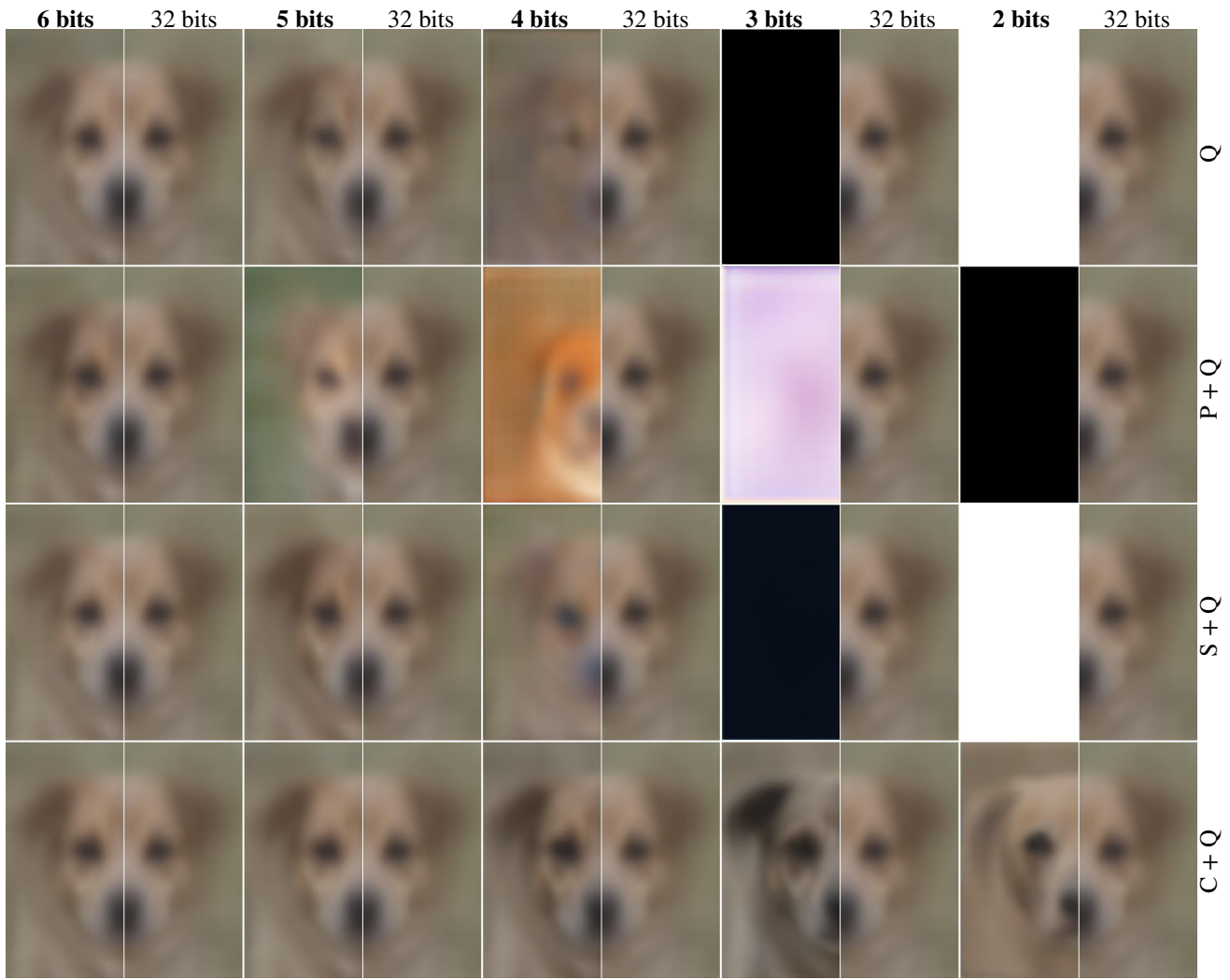}
    \begin{picture}(0,0)
        \put(-472,385){\bf6 bits}
        \put(-426,385){32 bits}
        \put(-375,385){\bf5 bits}
        \put(-330,385){32 bits}
        \put(-280,385){\bf4 bits}
        \put(-235,385){32 bits}
        \put(-185,385){\bf3 bits}
        \put(-135,385){32 bits}
        \put(-85,385){\bf2 bits}
        \put(-40,385){32 bits}
        \put(-1,330){\rotatebox{90}{Q}}
        \put(-1,226){\rotatebox{90}{P + Q}}
        \put(-1,130){\rotatebox{90}{S + Q}}
        \put(-1,34){\rotatebox{90}{C + Q}}
    \end{picture}     
    \caption{ADA mean 10k images on AFHQ Dog 512x512.}
    \label{fig:appendix_mean_images_2_to_6_bits_afhqdog_ada}
\end{figure*}

\begin{figure*}
    \begin{picture}(100,300)      
        \put(0,0){\includegraphics[width=0.98\textwidth]{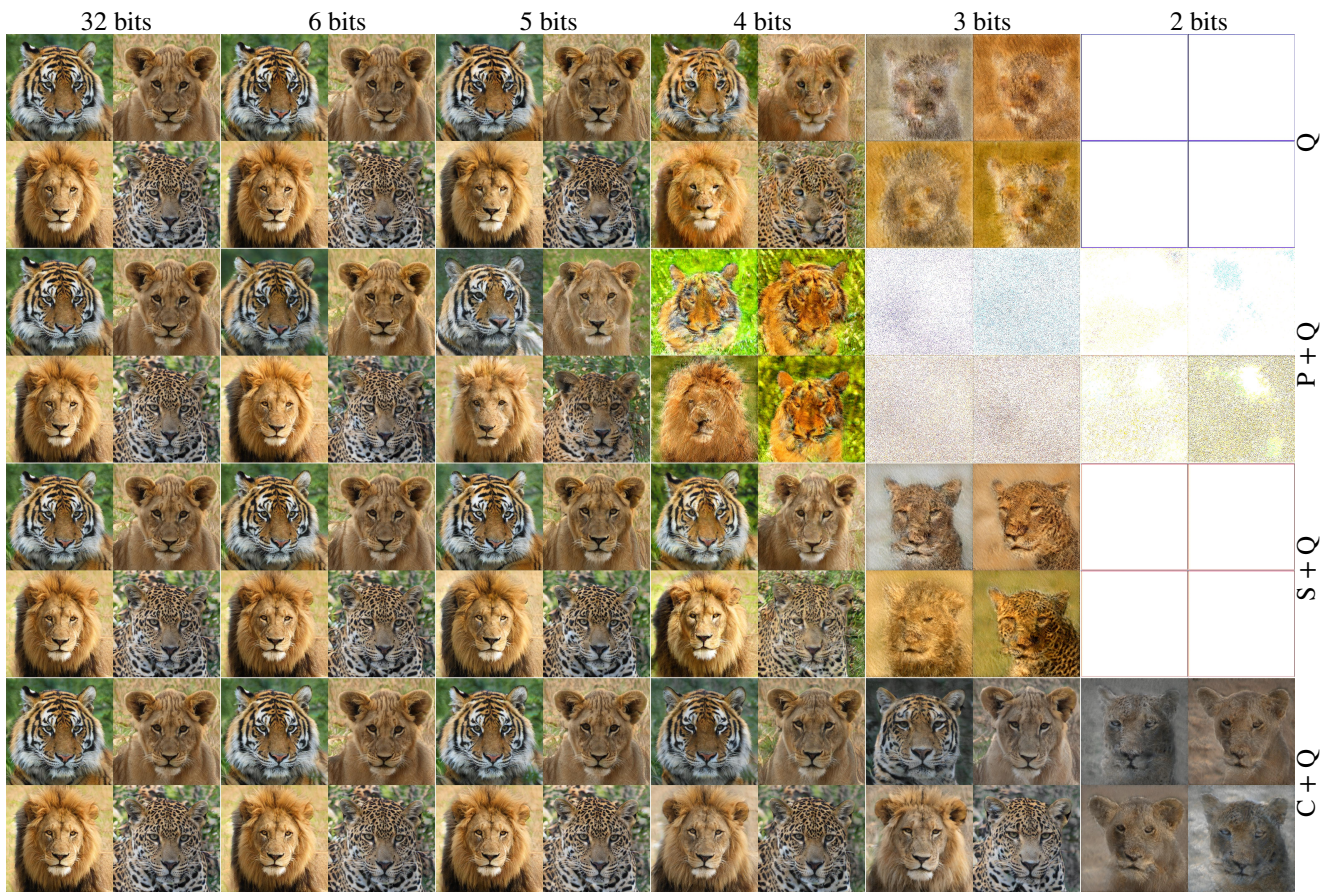}}
        \put(28,326){32 bits}
        \put(114,326){6 bits}
        \put(194,326){5 bits}
        \put(275,326){4 bits}
        \put(358,326){3 bits}
        \put(440,326){2 bits}
        \put(488,280){\rotatebox{90}{Q}}
        \put(488,191){\rotatebox{90}{P + Q}}
        \put(488,110){\rotatebox{90}{S + Q}}
        \put(488,28){\rotatebox{90}{C + Q}}
    \end{picture}     
    \caption{ADA on AFHQ Wild 512x512.}
    \label{fig:appendix_afhqwild_ada}
\end{figure*}

\begin{figure*}
    \includegraphics[width=0.97\textwidth]{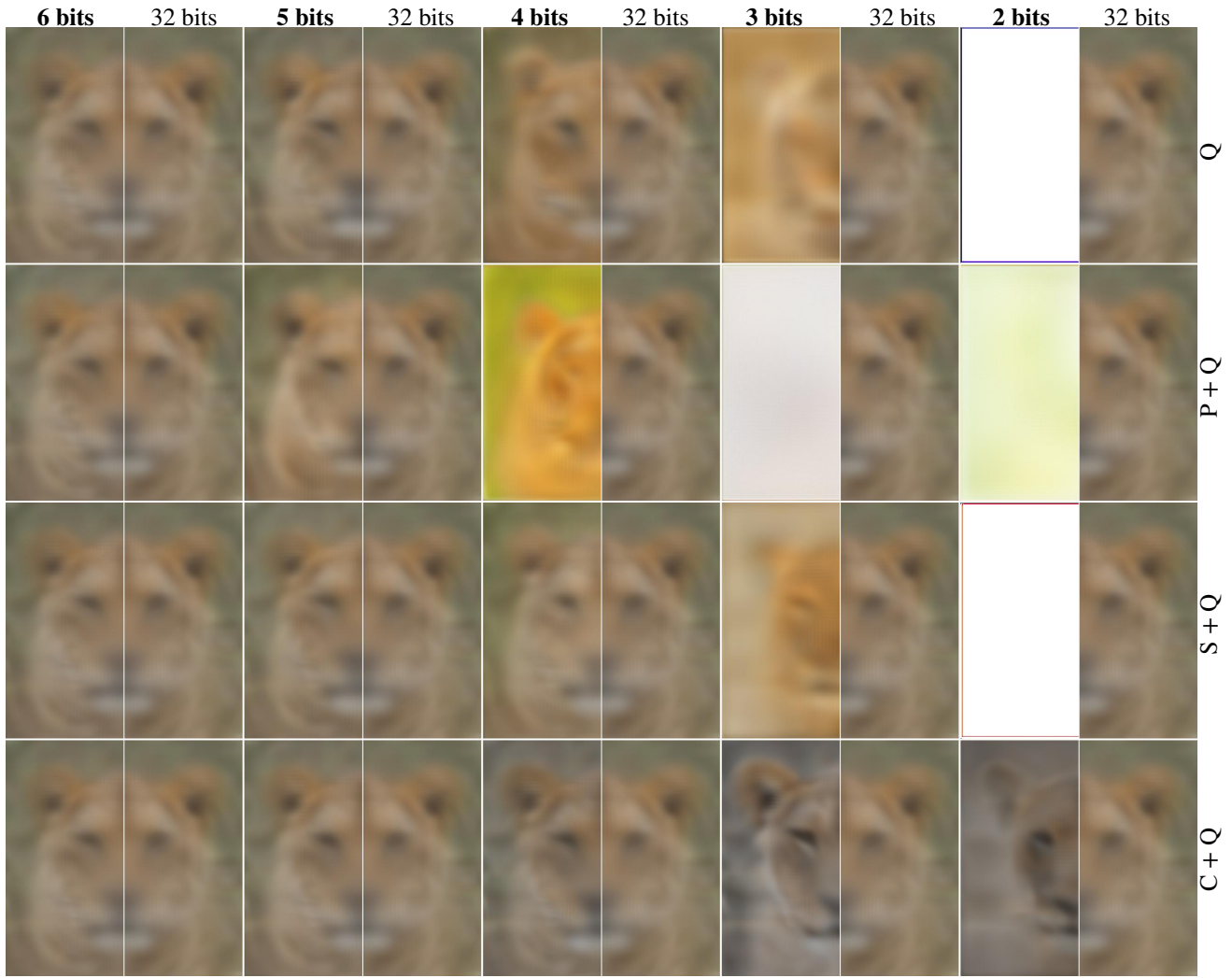}
    \begin{picture}(0,0)
        \put(-472,385){\bf6 bits}
        \put(-426,385){32 bits}
        \put(-375,385){\bf5 bits}
        \put(-330,385){32 bits}
        \put(-280,385){\bf4 bits}
        \put(-235,385){32 bits}
        \put(-185,385){\bf3 bits}
        \put(-135,385){32 bits}
        \put(-85,385){\bf2 bits}
        \put(-40,385){32 bits}
        \put(-1,330){\rotatebox{90}{Q}}
        \put(-1,226){\rotatebox{90}{P + Q}}
        \put(-1,130){\rotatebox{90}{S + Q}}
        \put(-1,34){\rotatebox{90}{C + Q}}
    \end{picture}     
    \caption{ADA mean 10k images on AFHQ Wild 512x512.}
    \label{fig:appendix_mean_images_2_to_6_bits_afhqwild_ada}
\end{figure*}

\begin{figure*}
    \begin{picture}(100,300)      
        \put(0,0){\includegraphics[width=0.98\textwidth]{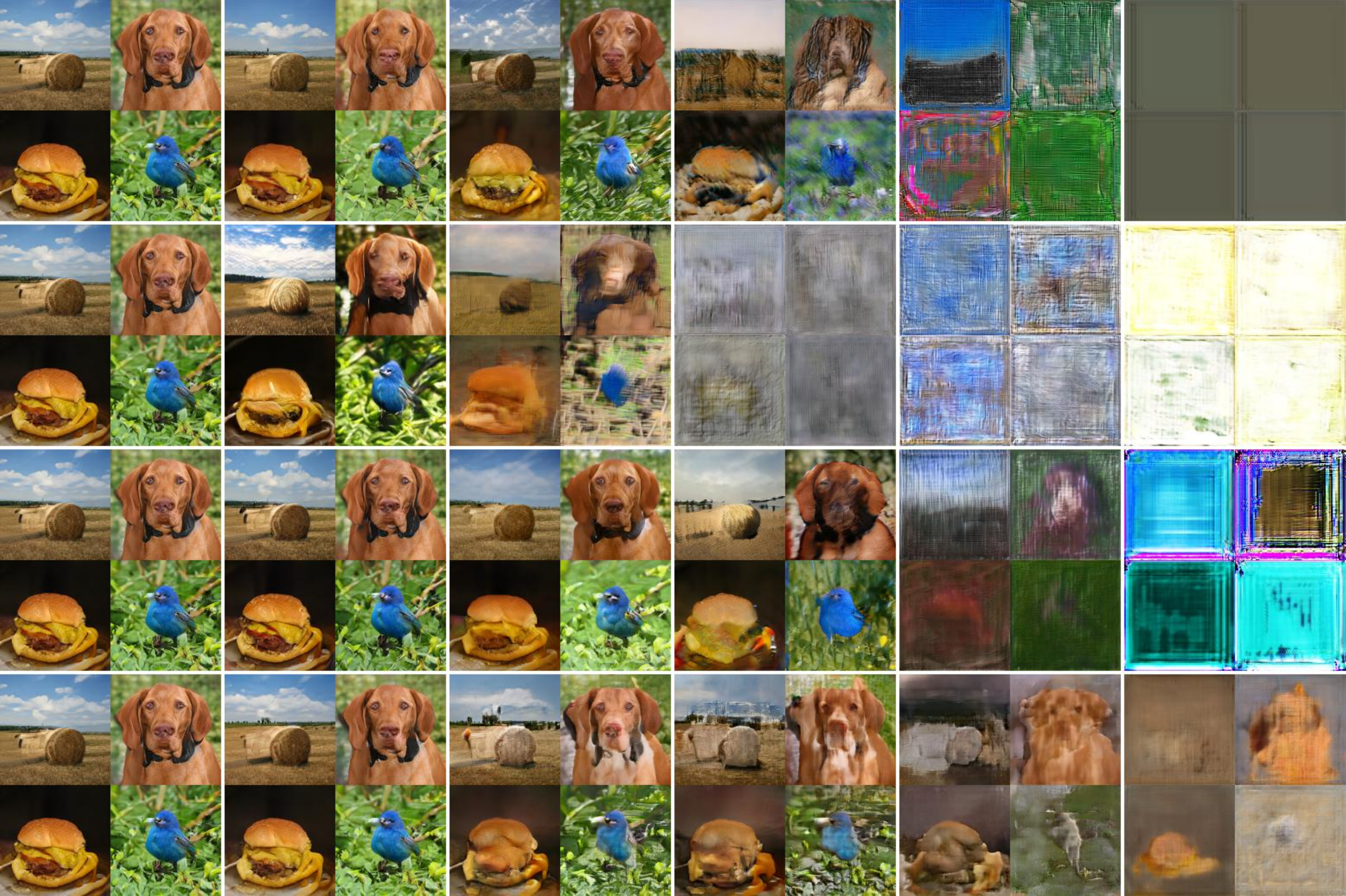}}
        \put(28,326){32 bits}
        \put(114,326){6 bits}
        \put(194,326){5 bits}
        \put(275,326){4 bits}
        \put(358,326){3 bits}
        \put(440,326){2 bits}
        \put(488,280){\rotatebox{90}{Q}}
        \put(488,191){\rotatebox{90}{P + Q}}
        \put(488,110){\rotatebox{90}{S + Q}}
        \put(488,28){\rotatebox{90}{C + Q}}
    \end{picture}     
    \caption{BigGAN on ImageNet 128x128.}
    \label{fig:appendix_imagenet_biggan}
\end{figure*}

Additional generated images of the studied networks and datasets are presented in Figures \ref{fig:appendix_ffhq1024_stylegan2}, \ref{fig:appendix_ffhq_ada}, \ref{fig:appendix_ffhq_stylegan2}, \ref{fig:appendix_ffhq_zcr}, \ref{fig:appendix_ffhq_sngan}, \ref{fig:appendix_ffhq_pagan}, \ref{fig:appendix_ffhq_ssgan}, \ref{fig:appendix_afhqcat_ada}, \ref{fig:appendix_afhqdog_ada}, \ref{fig:appendix_afhqwild_ada}, and \ref{fig:appendix_imagenet_biggan}. The respective mean 10k images are presented in Figures \ref{fig:appendix_mean_images_2_to_6_bits_ffhq1024_stylegan2}, \ref{fig:appendix_mean_images_2_to_6_bits_ffhq_ada}, \ref{fig:appendix_mean_images_2_to_6_bits_ffhq_stylegan2}, \ref{fig:appendix_mean_images_2_to_6_bits_ffhq_zcr}, \ref{fig:appendix_mean_images_2_to_6_bits_ffhq_sngan}, \ref{fig:appendix_mean_images_2_to_6_bits_ffhq_pagan}, \ref{fig:appendix_mean_images_2_to_6_bits_ffhq_ssgan}, \ref{fig:appendix_mean_images_2_to_6_bits_afhqcat_ada}, \ref{fig:appendix_mean_images_2_to_6_bits_afhqdog_ada}, and \ref{fig:appendix_mean_images_2_to_6_bits_afhqwild_ada}. 

\end{document}